\documentclass[lettersize,journal]{IEEEtran}
\usepackage[pagebackref,breaklinks,colorlinks,urlcolor=black]{hyperref} 
\hypersetup{
    citecolor=green!80!black
}
\usepackage{amsmath,amsfonts}
\usepackage{array}
\usepackage[caption=false,font=normalsize,labelfont=sf,textfont=sf]{subfig}
\usepackage{textcomp}
\usepackage{stfloats}
\usepackage{url}
\usepackage{verbatim}
\usepackage{graphicx}
\usepackage{cite}
\usepackage{amsmath}
\usepackage{amssymb}
\usepackage{mathtools}
\usepackage[ruled,vlined,linesnumbered]{algorithm2e}
\usepackage{multirow}
\usepackage{xcolor,colortbl}
\usepackage{booktabs}       
\usepackage{bbding}

\usepackage[capitalize,noabbrev]{cleveref}
\crefname{section}{Sec.}{Secs.}
\Crefname{section}{Section}{Sections}
\Crefname{table}{Table}{Tables}
\crefname{table}{Tab.}{Tabs.}
\crefname{figure}{Fig.}{Figs.}
\crefname{wrapfigure}{Fig.}{Figs.}
\crefname{equation}{Eq.}{Eqs.}
\crefname{algorithm}{Alg.}{Algs.}
\crefname{appendix}{App.}{Apps.}

\begin{document}
\title{Task-Distributionally Robust \\ Data-Free
Meta-Learning}

\author{Zixuan Hu,
        Yongxian Wei,
        Li Shen,
        Zhenyi Wang,
        Baoyuan Wu,
        Chun Yuan,
        Dacheng Tao,~\IEEEmembership{Fellow,~IEEE}
\thanks{This project is supported by the National Research Foundation, Singapore, under its NRF Professorship Award No. NRF-P2024-001.}
\thanks{Zixuan Hu and Dacheng Tao are with the College of Computing and Data Science, Nanyang Technological University, Singapore 639798 (e-mail:\ ZIXUAN014@e.ntu.edu.sg;\ dacheng.tao@ntu.edu.sg).}
\thanks{Li Shen is with the School of Cyber Science and Technology, Shenzhen Campus of Sun Yat-sen University, Shenzhen 518107, China (e-mail:\ mathshenli@gmail.com).}
\thanks{Yongxian Wei and Chun Yuan are with the Tsinghua Shenzhen International Graduate School, Tsinghua University, Shenzhen 518055, China (e-mail:\ weiyx23@mails.tsinghua.edu.cn;\ yuanc@sz.tsinghua.edu.cn).}
\thanks{Zhenyi Wang is with the Department of Computer Science and AI Institute, University
of Central Florida, FL 32816, USA (e-mail: zhenyi.wang@ucf.edu).}
\thanks{Baoyuan Wu is with School of Data Science, the Chinese University of
Hong Kong, Shenzhen (CUHK-Shenzhen) and Secure Computing Lab of
Big Data, Shenzhen Research Institute of Big Data (SBRID), Shenzhen
518172, China (e-mail: wubaoyuan1987@gmail.com).}
\thanks{Corresponding author: Li Shen.}
\thanks{This work has been accepted for publication in IEEE Transactions on Pattern Analysis and Machine Intelligence (IEEE TPAMI) in September 2025.}
}

\markboth{Journal of \LaTeX\ Class Files,~Vol.~14, No.~8, August~2021}%
{Shell \MakeLowercase{\textit{et al.}}: A Sample Article Using IEEEtran.cls for IEEE Journals}

\IEEEpubid{0000--0000/00\$00.00~\copyright~ IEEE}

\maketitle

\begin{abstract}
    Data-Free Meta-Learning (DFML) aims to enable efficient learning of unseen few-shot tasks, by meta-learning from multiple pre-trained models without accessing their original training data. While existing DFML methods typically generate synthetic data from these models to perform meta-learning, a comprehensive analysis of DFML’s robustness—particularly its failure modes and vulnerability to potential attacks—remains notably absent. Such an analysis is crucial as algorithms often operate in complex and uncertain real-world environments. 
    This paper fills this significant gap by systematically investigating the robustness of DFML, identifying two critical but previously overlooked vulnerabilities: Task-Distribution Shift (TDS) and Task-Distribution Corruption (TDC). TDS refers to the sequential shifts in the evolving task distribution, leading to the catastrophic forgetting of previously learned meta-knowledge. TDC exposes a security flaw of DFML, revealing its susceptibility to attacks when the pre-trained model pool includes untrustworthy models that deceptively claim to be beneficial but are actually harmful.
    To mitigate these vulnerabilities, we propose a trustworthy DFML framework comprising three components: synthetic task reconstruction, meta-learning with task memory interpolation, and automatic model selection. Specifically, utilizing model inversion techniques, we reconstruct synthetic tasks from multiple pre-trained models to perform meta-learning. To prevent forgetting, we introduce a strategy to replay interpolated historical tasks to efficiently recall previous meta-knowledge. Furthermore, our framework seamlessly incorporates an automatic model selection mechanism to automatically filter out untrustworthy models  during the meta-learning process.
     Extensive experiments across various datasets with two types of untrustworthy models confirm the superiority of our method in significantly enhancing the robustness of DFML. Code is available at \url{https://github.com/Egg-Hu/Trustworthy-DFML}.
\end{abstract}


\section{Introduction}
\label{sec:intro}
\IEEEPARstart{M}{eta-learning}, also known as \textit{learning to learn}, aims to enable efficient learning of unseen few-shot tasks by leveraging general \textit{meta-knowledge} from a collection of related tasks.
Traditional data-based meta-learning \cite{finn2017model} typically assumes the availability of task-specific training and testing data, also known as the \textit{support set} and \textit{query set}, for each task.
However, the assumption that such specific data will always be available for every task is often impractical in real-world applications. In many scenarios, accessibility to task-specific data may be restricted due to privacy concerns, proprietary rights, or the sheer volume of the dataset. For example, platforms like GitHub and HuggingFace offer multiple pre-trained models, but they do not typically provide the original training data used to train these models. This limitation poses significant challenges for traditional data-based meta-learning methods and also unveils an intriguing research question: Is it feasible to leverage the vast availability of pre-trained models to facilitate efficient learning of new tasks?

\IEEEpubidadjcol
More recently, Data-Free Meta-Learning (DFML) \cite{wang2022metalearning,hu2023architecture,hu2023learning,wei2024free,wei2024task} has been proposed as an innovative paradigm of meta-learning, seeking to learn the meta-knowledge solely from  multiple pre-trained models without accessing their original training data. This paradigm not only circumvents the barriers imposed by data accessibility but also leverages the vast availability of pre-trained models, potentially transforming how meta-learning systems work in data-restricted environments. Typically, existing DFML methods create multiple tasks with synthetic data (\textit{i.e.}, synthetic tasks) from the pre-trained models to perform meta-learning. While promising, a thorough analysis of DFML’s robustness—particularly its failure modes and vulnerability to attacks—remains notably absent. Such an analysis is crucial as algorithms often operate in complex and uncertain real-world environments.

\begin{figure*}[!t]
  \centering
    \includegraphics[width=0.99\linewidth]{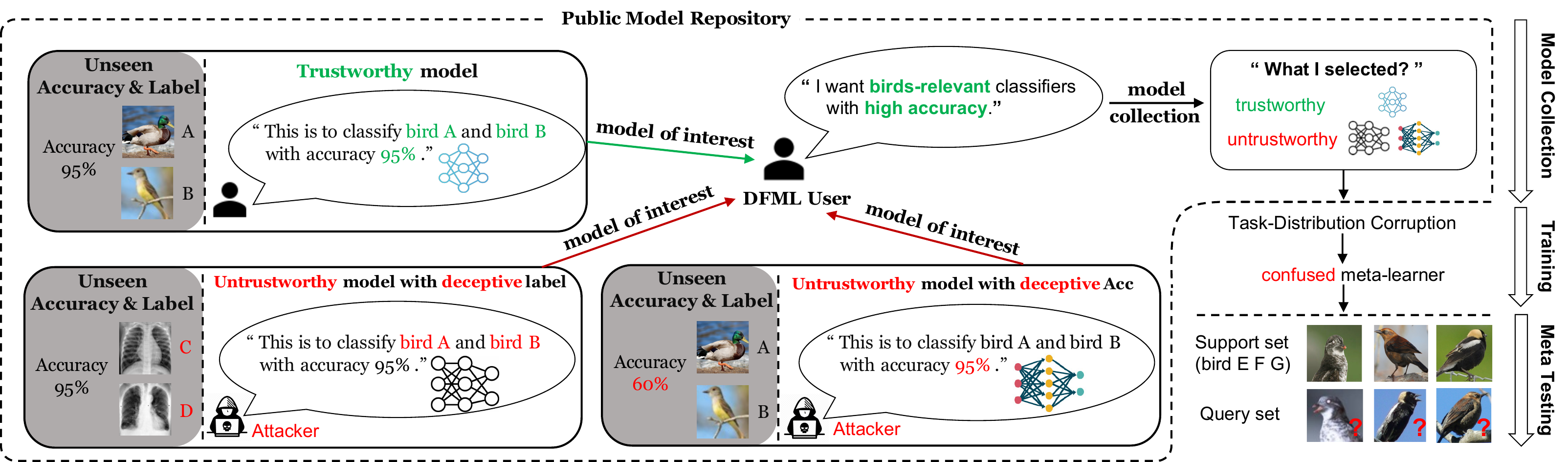}
   \caption{An illustrative example of how Task-Distribution Corruption (TDC) happens. TDC exposes a security flaw of DFML, revealing its susceptibility to attacks when the pre-trained model pool includes untrustworthy models that deceptively claim to be beneficial but are actually harmful. This figure illustrates two types of untrustworthy models: 
(i) models with deceptive labels (\textit{e.g.}, claiming to classify classes A and B, but actually classifying C and D), and (ii) models of deceptive accuracy (\textit{e.g.}, claiming a high accuracy when the true accuracy is low). When employing DFML methods to solve tasks of rare bird species classification, DFML users actively collect high-accuracy models of interest, such as similar bird or animal classifiers of high accuracy, as meta-training resources. However, this subjective selection also exposes a security vulnerability: attackers can manipulate the labels or accuracy associated with the models, thereby misleading users into collecting harmful models.
}
   \label{fig:type}
   \vspace{-0.3cm}
\end{figure*}

 \begin{figure}[!t]
  \centering
\includegraphics[width=1\linewidth]{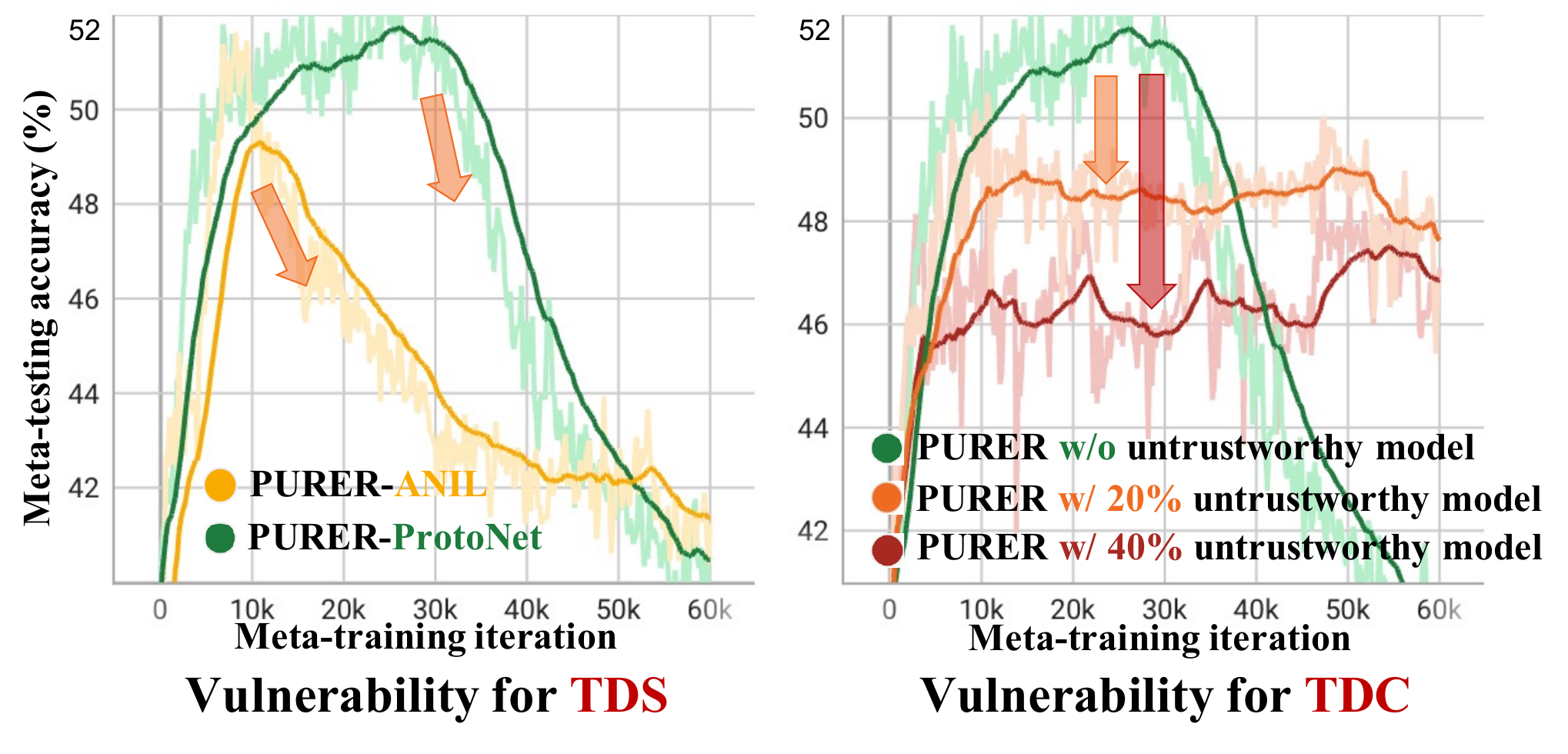}
\vspace{-0.6cm}
   \caption{Vulnerabilities of PURER \cite{hu2023architecture} for TDS and TDC. (Left) TDS causes the forgetting of previously learned meta-knowledge, resulting in a progressive decline in meta-testing accuracy alongside the meta-training process. (Right) The inclusion of untrustworthy models will significantly undermine the effectiveness of DFML. As the proportion of untrustworthy models in the model pool increases, the performance degradation of DFML becomes more severe.}
   \label{fig:limitation}
   \vspace{-0.4cm}
\end{figure}

In this paper, we fill this significant gap by systematically investigating the robustness of DFML, including its failure modes and vulnerability to attacks.
Specifically, we identify two critical but previously overlooked vulnerabilities of DFML: Task-Distribution Shift (TDS) and Task-Distribution Corruption (TDC). \textbf{TDS refers to the sequential shifts in the evolving task distribution, leading to the catastrophic forgetting of previously learned meta-knowledge.} As illustrated in \cref{fig:tds}, we visualize how the task distribution evolves over time, where the synthetic data gradually transitions from noisy to clear, and the classification task progressively shifts from easy to more difficult.
Such shifts can lead to catastrophic forgetting of previously learned knowledge when transferring a machine learning model to a new context \cite{wang2022meta,son2023meta}. In DFML, the impact of TDS is evident as shown in  \cref{fig:limitation} (left): we observe a significant decline in meta-testing accuracy over time for PURER \cite{hu2023architecture}—a classic DFML method—under the CIFAR-FS 5-way 5-shot setting.

\textbf{TDC exposes a security flaw of DFML, revealing its susceptibility to attacks when the pre-trained model pool includes untrustworthy models that deceptively claim to be beneficial but are actually harmful.}
Existing DFML methods assume that all models selected by DFML users are inherently trustworthy \cite{wang2022metalearning,hu2023architecture,hu2023learning,wei2024task}. However, this assumption often does not hold in practical scenarios and exposes vulnerabilities that attackers can exploit.
Untrustworthy models can be deliberately disguised and sent to DFML users by attackers, or inadvertently released to public repositories by normal users. In this paper, we categorize two types of untrustworthy models: 
(i) models with deceptive labels (\textit{e.g.}, claiming to classify classes A and B, but actually classifying C and D), and (ii) models of deceptive accuracy (\textit{e.g.}, claiming a high accuracy when the true accuracy is low).  
Consider the example in \cref{fig:type}, which illustrates how DFML users falsely select those untrustworthy models. When employing DFML to solve tasks such as rare bird species classification, DFML users may actively collect high-accuracy models of interest, those for similar bird or animal classifiers, as meta-training resources. This is because meta-knowledge is more easily generalizable to similar tasks \cite{oh2022understanding}.
However, this subjective selection also exposes a security vulnerability: attackers can manipulate the labels or accuracy associated with the models, thereby misleading users into collecting harmful models.
Since existing DFML methods have not yet considered this situation, it introduces uncertain security risks for the real-world deployment of DFML. Our empirical analysis, as shown in \cref{fig:limitation} (right), demonstrates that the inclusion of untrustworthy models will significantly reduce the generalization performance of the meta-learner. In this case, these untrustworthy models are attached with deceptive labels, which self-claim to classify classes in CIFAR-FS but actually classify classes in datasets including EuroSAT \cite{helber2019eurosat}, ISIC \cite{tschandl2018ham10000}, ChestX \cite{wang2017chestx}, Omniglot \cite{lake2015human}, and MNIST \cite{lecun2010mnist}.

To mitigate these vulnerabilities, we propose a trustworthy DFML framework comprising three components: synthetic task reconstruction, meta-learning with task memory interpolation, and automatic model selection.
Specifically, utilizing model inversion techiniques \cite{yin2020dreaming,fang2021contrastive}, we reconstruct synthetic tasks from multiple pre-trained models to perform meta-learning. To mitigate the forgetting of previously learned meta-knowledge caused by TDS, we introduce a strategy to replay interpolated historical tasks to efficiently recall previous meta-knowledge. We present two practical implementations of task memory interpolation, including \textit{task combination} and \textit{task mixup}.
Furthermore, we propose an automatic model selection mechanism to prevent the false selection of untrustworthy models as training resources for DFML. This is achieved by quantifying each model's trustworthiness with a learnable weight, which is optimized to enhance the generalization capacity of the meta-learner. The optimization is performed by the policy gradient algorithm, thus addressing the non-differentiable challenge posed by model selection. 
 We outline our contributions as follows:
\begin{itemize}
    \item \textbf{Innovative perspective:} To the best of our knowledge, this is the first work that systematically analyzes the robustness of DFML, particularly focusing on its failure modes and vulnerability to potential attacks. We identify two critical but overlooked issues in DFML: Task-Distribution Shift (TDS) and Task-Distribution Corruption (TDC), detailing their causes, impacts, and associated challenges. We believe such a robustness analysis is crucial for the development of DFML, since algorithms often need to operate in uncertain real-world environments.
    \item \textbf{Technical contributions:} We propose the first trustworthy DFML framework to effectively enhance the robustness against TDS and TDC. 
   These contributions extend beyond mere performance enhancements, providing fresh insights into the robustness and reliability of DFML. Additionally, our framework is versatile and can be seamlessly integrated with both
gradient-based and metric-based meta-learning algorithms, thus enabling integration with more advanced future algorithms for improved results.
    \item \textbf{Evaluation:} Comprehensive experiments across four datasets with two types of untrustworthy models validate the superiority of our method in significantly enhancing the robustness of DFML.
\end{itemize}
\section{Related Work}
\label{sec:related}

\subsection{Meta-Learning \& Data-Free Meta-Learning}
Gradient-based meta-learning \cite{finn2017model} aims to learn the initial model parameters or hyperparameters of deep neural networks that can be efficiently fine-tuned to similar tasks. 
Black-box meta-learning \cite{santoro2016meta} aims to meta-learn a hyper neural network across multiple tasks. This hypernetwork is designed to directly output the task-specific model parameters, conditioned on the support set of each task.
Metric-based meta-learning \cite{snell2017prototypical,vinyals2016matching} focuses on learning a function to compute a distance metric, measuring the similarity between data points and using proximity to make classifications. 

In contrast to  traditional data-based meta-learning methods that rely on task-specific data \cite{vettoruzzo2024advances}, DFML \cite{hu2023architecture,hu2023learning,wang2022metalearning,wei2024free,wei2024task} assumes only a pre-trained model is available for each task.
DFL2L \cite{wang2022metalearning} trains a hyper-network, taking all models as inputs and directly outputting the meta-learner in the parameter space. However, this method requires all pre-trained models to share the same architecture and can not scale to large-scale pre-trained models.
PURER \cite{hu2023architecture} and FREE \cite{wei2024free} achieve significant improvements by training the meta-learner with synthetic tasks. BiDf-MKD \cite{hu2023learning} further extends the DFML to the black-box scenario by introducing the concept of learning to learn from APIs, aiming to extract the prior meta-knowledge from multiple APIs.
Despite these advancements, a comprehensive analysis of DFML’s robustness—its failure modes and vulnerability to attacks—remains notably absent. We believe such a robustness analysis is crucial as algorithms often need to operate in complex and uncertain real-world environments.

\subsection{Synthetic Data Generation via Model Inversion.}
Existing DFML methods typically employ model inversion to generate synthetic data from pre-trained models, validating the effectiveness of using synthetic data for meta-learning. Typically, model inversion techniques can be categorized into pixel-based and generator-based approaches. Pixel-based methods \cite{ liu2021data, wang2021data} optimize each pixel individually, resulting in a significant increase in learnable parameters with more images. In contrast, generator-based methods \cite{fang2022up, chen2019data} optimize a generator (or with the latent vector) to produce images, maintaining a fixed (or slowly increasing) number of learnable parameters as image count grows. Additionally, the architecture of a generator (even randomly initialized) can implicitly incorporate the Deep Image Prior \cite{ulyanov2018deep} as regularization. 

\subsection{When Meta-Learning Meets Sequential Task Stream}
Traditional meta-learning methods \cite{finn2017model} assume the task distribution is stationary. 
Online meta-learning \cite{finn2019online}, however, assumes the tasks are provided as a stream, where each task becomes available sequentially and cannot be accessed again. Continual meta-learning \cite{wang2022meta,wang2021meta,son2023meta}  shares many similarities with online
meta-learning, including the streaming tasks and continual updating of the meta-learner. However, continual meta-learning assumes the task distribution is not stationary, \textit{i.e.}, tasks are sampled from heterogeneous task distributions. 
Similar to continual meta-learning, we also observe the sequential shifts in the evolving task distribution in DFML (see \cref{fig:tds}).
However, the task-distribution shift in DFML is fundamentally different and more challenging than that in the data-based setting. One key difference is that previous methods assume tasks are densely sampled from each task distribution, whereas in DFML, tasks are sparsely sampled. 
For instance, data-based meta-learning can easily construct up to ${\rm C}_{60}^5$ 5-way tasks using a dataset with 60 classes, but it is challenging to collect such a large number of pre-trained models in DFML. Consequently, directly applying these methods does not yield satisfactory results, as there is insufficient information from previous distributions
to prevent forgetting and recall previously learned meta-knowledge. 

\section{Rethinking Data-Free Meta-Learning}
\subsection{Preliminaries \& Problem Setup}
\label{sec:preliminary}
\textbf{Goal of meta-learning.}\ The goal of meta-training is to train a meta-learner (parameterized by $\boldsymbol{\theta}_{\mathcal{A}}$) for the best performance over a distribution of tasks $p_{\mathcal{T}}$:
\begin{equation}
\min_{\boldsymbol{\theta}_{\mathcal{A}}}\mathbb{E}_{\mathcal{T} \sim {p}_{\mathcal{T}}}\mathcal{L}_{\rm task}\left({\mathbf{D}}^{\rm q};\mathcal{A}[{\mathbf{D}}^{\rm s};\boldsymbol{\theta}_{\mathcal{A}}]\right),
    \label{eq:metaobjective}
\end{equation}
Here, ${\mathcal{T}}$ represents the task sampled from the task distribution ${p}_{\mathcal{T}}$. Each task $\mathcal{T}$ comprises a task-specific training set (\textit{i.e.,} support set) ${\mathbf{D}}^{\rm s}=(\mathbf{X}^s, \mathbf{Y}^s)$ and a task-specific testing set (\textit{i.e.,} query set) ${\mathbf{D}}^{\rm q}=(\mathbf{X}^q, \mathbf{Y}^q)$.  $\mathcal{A}[{\mathbf{D}}^{\rm s};\boldsymbol{\theta}_{\mathcal{A}}]$ denote the task-specific adaption process: the meta-learner $\boldsymbol{\theta}_{\mathcal{A}}$ uses the support set ${\mathbf{D}}^{\rm s}$ to perform adaptation (\textit{e.g.,} a few steps of gradient descent) and obtain the task-specific model $\mathcal{A}\left[\mathbf{D}^{\mathrm{s}} ; \boldsymbol{\theta}_{\mathcal{A}}\right]$.
$\mathcal{L}_{\rm task}$ refers to the task-level loss, which assesses the performance of the task-specific model on the query set ${\mathbf{D}}^{\rm q}=(\mathbf{X}^q, \mathbf{Y}^q)$ of each task. 
To simulate the few-shot setting, each task $\mathcal{T}$ is $N$-way $K$-shot, where ${\mathbf{D}}^{\rm s}$ contain $N$ classes and $K$ instances per class (K is typically set as 1 or 5). 
The rationale behind \cref{eq:metaobjective} is training a meta-learner across various tasks sampled from the task distribution, so that it can acquire the general meta-knowledge to enable fast adaption to similar but unseen tasks sampled from the same task distribution.

\textbf{Extension to data-free setting.} In the context of DFML, for each task $\mathcal{T} \sim p_{\mathcal{T}}$, we only have access to its pre-trained model $M$, with no access to its original support set and query set. We denote the set of pre-trained models as $\mathcal{M}_{\text{pool}}$. Note that the models may possess heterogeneous architectures. Existing DFML methods \cite{hu2023architecture,hu2023learning} typically address this data-free constraint by synthesizing data directly from each pre-trained model. To align with \cref{eq:metaobjective}, the optimization objective of DFML can be formally expressed as:
\begin{equation}
\begin{aligned}
\min_{\boldsymbol{\theta}_{\mathcal{A}}}&\mathbb{E}_{M \sim \mathcal{M}_{\text{pool}}}\mathcal{L}_{\rm task}\big({\mathbf{D}}^{\rm q}; \mathcal{A}[{\mathbf{D}}^{\rm s}; \boldsymbol{\theta}_{\mathcal{A}}]\big),\\ &\text{where}\ ({\mathbf{D}}^{\rm s}, {\mathbf{D}}^{\rm q})=\operatorname{Syn}(M).
\end{aligned}
\label{eq:dfmlobjective_reb}
\end{equation}
Here, with a slight abuse of notation, we use $({\mathbf{D}}^{\rm s}, {\mathbf{D}}^{\rm q})$ to denote the synthetic support and query sets generated from the model $M$ using the function $\operatorname{Syn(\cdot)}$. This formulation explicitly parallels \cref{eq:metaobjective} by replacing the data-based tasks $\mathcal{T}$ with synthetic data generated from pre-trained models $M$, thereby addressing the data-free constraints of DFML.

\textbf{Meta-testing.}\ We evaluate the meta-learner on 600 unseen $N$-way $K$-shot tasks. ``Unseen'' indicates the classes in meta-testing tasks are not seen during meta-training. Existing meta-learning methods \cite{finn2017model,hu2023architecture} typically divide the dataset into two disjoint sets of classes, separately used for constructing meta-training and meta-testing tasks.
Each $N$-way $K$-shot meta-testing task $\mathcal{T}_{\rm test} = (\mathbf{D}_{\rm test}^{\rm s}, \mathbf{D}_{\rm test}^{\rm q})$ consists of a support set $\mathbf{D}_{\rm test}^{\rm s} = (\mathbf{X}_{\rm test}^{\rm s}, \mathbf{Y}_{\rm test}^{\rm s})$ with $N$ classes and $K$ instances per class. For the evaluation on each meta-testing task, we conduct two steps: \textbf{(i) Adapt.} The meta-learner (parameterized by $\boldsymbol{\theta}_{\mathcal{A}}$) uses the the task-specific support set $\mathbf{D}_{\rm test}^{\rm s}$ to perform adaptation and obtain the task-specific model $\mathcal{A}[\mathbf{D}_{\rm test}^{\rm s}; \boldsymbol{\theta}_{\mathcal{A}}]$; \textbf{(ii) Predict.} The task-specific model $\mathcal{A}[\mathbf{D}_{\rm test}^{\rm s}; \boldsymbol{\theta}_{\mathcal{A}}]$ then makes predictions on the query set $\mathbf{D}_{\rm test}^{\rm q}$, which is what we ultimately use to test accuracy. The overall accuracy is measured by averaging the accuracy across all the meta-testing tasks. Note that the meta-testing procedure of data-free meta-learning is the same as the traditional data-based meta-learning \cite{finn2017model}.

\textbf{Case study of DFML: PURER \cite{hu2023architecture}.} To address the DFML objective in \cref{eq:dfmlobjective_reb}, PURER \cite{hu2023architecture} instantiates the data-generation function $\operatorname{Syn}(\cdot)$  via model inversion \cite{yin2020dreaming} and concurrently trains the meta-learner using the synthetic data. This can be formulated as a dual optimization objective that jointly optimizes data synthesis and meta-learning:
\begin{equation}
\max_{\boldsymbol{\theta}_{\mathcal{A}}}\underset{M}{\mathbb{E}}\min_{\mathbf{X}}\mathcal{L}_{\rm cls}\left(M\left(\mathbf{X}\right),\mathbf{Y}\right)+\mathcal{R}(\mathbf{X})-\mathcal{L}_{\rm task}\left(\mathbf{X};\boldsymbol{\theta}_{\mathcal{A}}\right).
\label{eq:purer}
\end{equation}
At each iteration, it involves three steps: 
\textbf{(i) Sample models} (${\underset{M\sim \mathcal{M}_{\rm pool}}{\mathbb{E}}}$). It first randomly samples a model $M$ (or a batch of models) from the given model pool $\mathcal{M}_{\rm pool}$. 
\textbf{(ii) Synthesize data} (${\underset{\mathbf{X}}{\min}}$). Given the sampled model $M$ and the self-defined target output $\mathbf{Y}$ (\textit{e.g.}, $[1, 0]$ for a binary classifier), it inversely reconstructs the input data via model inversion \cite{yin2020dreaming}: It first randomly initializes the input $\mathbf{X}$ with Gaussian noise, and then optimizes $\mathbf{X}$ by minimizing \cref{eq:purer}. Specifically, minimizing the classification loss $\mathcal{L}_{\rm cls}$ ensures $\mathbf{X}$ can be correctly predicted as the target label $\mathbf{Y}$; maximizing the meta-learning loss $\mathcal{L}_{\rm task}$ can synthesize $\mathbf{X}$ of higher difficulty, thus enhancing diversity \cite{hu2023architecture}. $\mathcal{R}$ is the regularization term used to enhance data quality, detailed in \cref{sec:task_reconstruction}. 
\textbf{(iii) Optimize the meta-learner} (${\underset{\boldsymbol{\theta}_{\mathcal{A}}}{\max}}$). With the synthetic $\mathbf{X}$, it trains the meta-learner (parameterized by $\boldsymbol{\theta}_{\mathcal{A}}$) by maximizing the negative meta-learning loss $-\mathcal{L}_{\rm task}$. To maintain clarity, we have not explicitly shown the support set and query set, as well as the adaptation process within the parentheses of $\mathcal{L}_{\rm task}$ defined in \cref{eq:metaobjective}.

\subsection{A closer look at TDS in DFML}
\begin{figure*}[!t]
  \centering
    \includegraphics[width=1\linewidth]{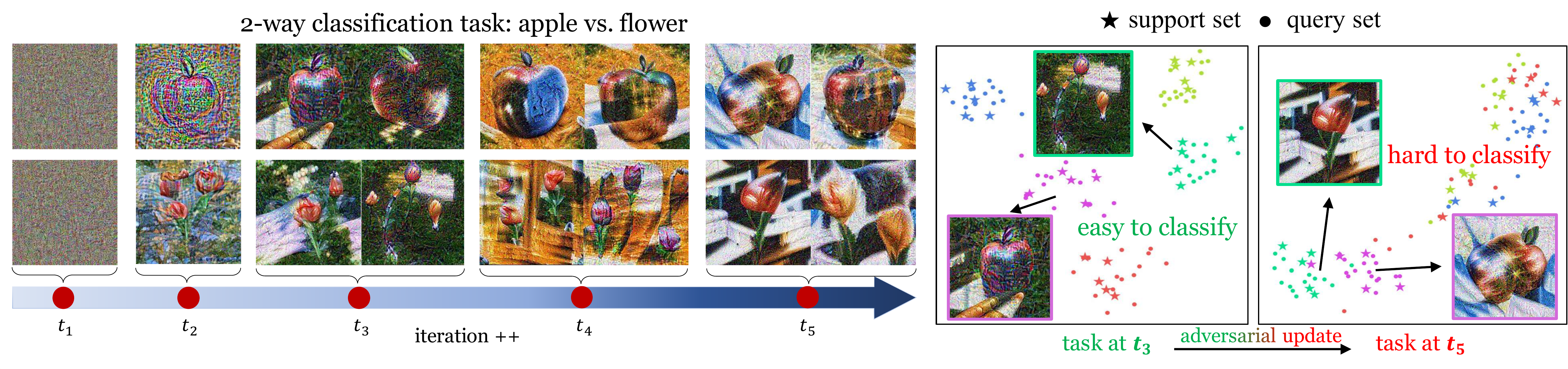}
  \vspace{-0.5cm}
   \caption{ Visualization of TDS. TDS refers to the sequential shifts in the evolving task distribution, where the synthetic data gradually transitions from noisy to clear, and the classification task progressively shifts from easy to more difficult (\textit{e.g.}, an apple and a flower of similar shape, color and background).}
   \label{fig:tds}
   \vspace{-0.35cm}
\end{figure*}

\textbf{Cause of TDS.} TDS refers to the sequential shifts in the evolving task distribution, leading to the catastrophic forgetting of previously learned meta-knowledge. As illustrated in \cref{fig:tds}, we visualize the evolving task distribution, where the synthetic data gradually transitions from noisy to clear, and the classification task progressively shifts from easy to more difficult.
This shift is attributed to the min-max optimization of synthetic data by minimizing the classification loss and adversarially maximizing the meta-learning loss (see \cref{eq:purer}). 
Specifically, (i) Noisy $\rightarrow$ Clear: By minimizing $\mathcal{L}_{\rm cls}$ towards the target label $\mathbf{Y}$, $\mathbf{X}$ is gradually optimized to display discriminative label-specific features (\textit{e.g.}, apple and flower) so that  $\mathbf{X}$ can be correctly predicted as the target label $\mathbf{Y}$.
(ii) Easy $\rightarrow$ Hard: By maximizing the meta-learning loss $\mathcal{L}_{\rm task}$, $\mathbf{X}$ is gradually optimized to become harder to classify by the meta-learner. For example, synthetic data tends to display features that are difficult to classify (\textit{e.g.}, an apple and a flower of similar shape, color and background).

\textbf{Impact of TDS.} Recent studies \cite{son2023meta} suggest that when a machine learning model transfers to a new environment, it often catastrophically forgets knowledge relevant to previous contexts. Our empirical results in \cref{fig:limitation} (left) demonstrate such catastrophic forgetting also occurs in the context of DFML:  the performance of the meta-learner deteriorates significantly over time.
Such accuracy degradation is highly undesirable in practical scenarios for two main reasons: (i) It indicates severe performance instability and significant performance declines. (ii) It renders snapshotting the best meta-learner impractical, particularly in the absence of validation set monitoring. This underscores the necessity for a stable DFML method, ensuring that the meta-learner maintains consistently high accuracy over time and that the meta-training phase can be safely terminated after a predetermined number of iterations.

\textbf{Challenges in addressing TDS.} Although there are existing studies \cite{wang2022meta,son2023meta} aimed at addressing distribution shifts, tackling TDS in the context of DFML presents unique challenges: (i) Sequential shifts in the evolving task distribution. The meta-learner need to adapt to the evolving task distribution where shifts appear sequentially throughout the meta-training process. (ii) Sparsely-sampled nature of synthetic task distribution. In the context of DFML, each pre-trained model can be considered as a sample drawn from the current task distribution.
However, unlike the densely-sampled nature of data-based meta-learning \cite{finn2017model}, sampling in DFML is sparse. For instance, data-based meta-learning can easily sample up to ${\rm C}_{60}^5$ 5-way tasks using a dataset with 60 classes, but it is challenging to collect such a large number of pre-trained models in DFML. This sparsely-sampled nature limits the effectiveness of many existing methods for addressing distribution shifts, as there is insufficient information from prior distributions to prevent forgetting and to facilitate the recall of previously learned knowledge. Moreover, since meta-learning typically requires a vast number of tasks to foster generalizable  meta-knowledge (\textit{e.g.}, over 60K tasks in MAML), the sparsely-sampled nature of DFML makes addressing TDS more challenging.

\begin{figure*}[!t]
  \centering
    \includegraphics[width=1\linewidth]{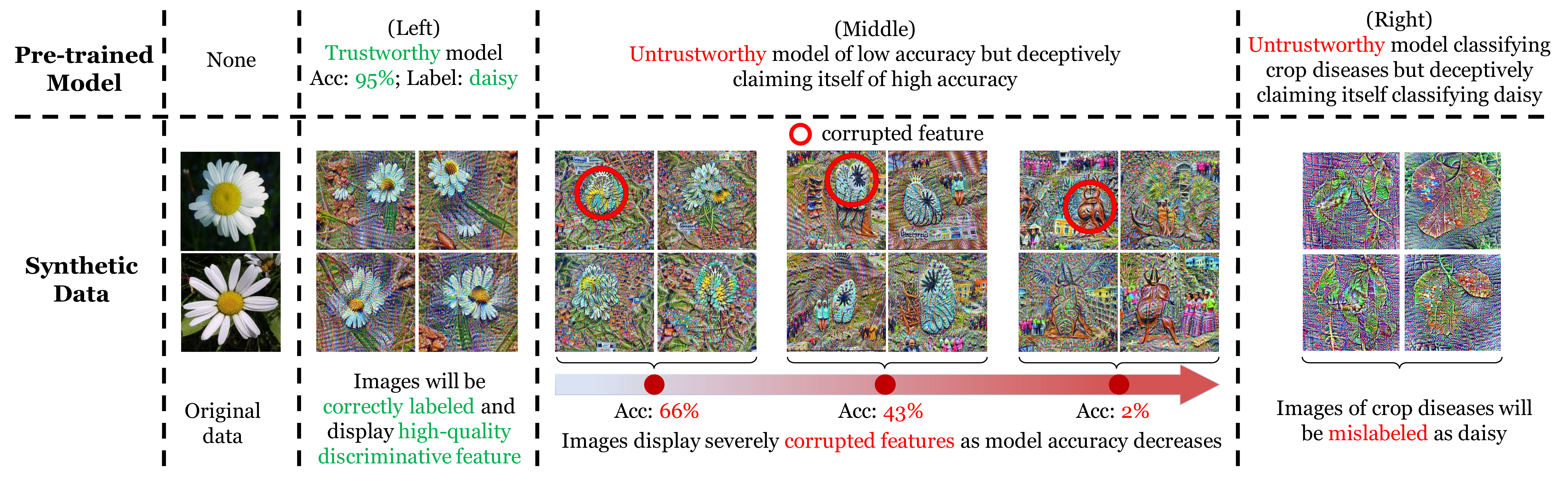}
  \vspace{-0.5cm}
   \caption{Visualization of TDC. (Left) Images synthesized from a trustworthy model will be correctly labeled and display high-quality discriminative features, such as white petals outside, yellow stamens inside, and green stems below.
(Middle) Models of deceptive accuracy can lead to synthetic tasks
with feature corruption.  As model accuracy decreases, images tend to display severely corrupted
features like white blobs and even incorrect features.
(Right) Models with deceptive labels can lead to synthetic tasks
with label noise. If a ``crop diseases" model deceptively claims itself as a ``daisy" model, images synthesized from this model  will be mislabelled as ``daisy", conflicting with the true ``daisy" images.}
   \label{fig:tdc_reason}
   \vspace{-0.3cm}
\end{figure*}

\subsection{A closer look at TDC in DFML}

\textbf{Cause of TDC.} 
Existing DFML methods assume that all models selected by DFML users are inherently trustworthy. However, this assumption often does not hold in practical scenarios and exposes vulnerabilities that attackers can exploit.
Untrustworthy models can be deliberately disguised and sent to DFML users by attackers, or inadvertently released to public repositories by normal users. As illustrated in \cref{fig:type}, we categorize two types of untrustworthy models: 
(i) models with deceptive labels (\textit{e.g.}, claiming to classify classes A and B, but actually classifying C and D), and (ii) models of deceptive accuracy (\textit{e.g.}, claiming a high accuracy when the true accuracy is low).
In \cref{fig:type}, we illustrate through an intuitive example how DFML users may mistakenly collect those untrustworthy models.


\textbf{Impact of TDC.} 
To figure out how different types of untrustworthy models affect the DFML, we take a closer look on the synthetic tasks inversely generated from those models:
\textbf{Models with deceptive labels can lead to synthetic tasks with label noise \cite{yao2021meta,zhan2024data}.} Consider a scenario where a ``crop diseases" classifier deceptively claims itself as a ``flower" classifier. In that case, if the DFML user intends to collect flower-relevant models, the  crop diseases classifier could be falsely collected as a flower classifier. 
As illustrated in \cref{fig:tdc_reason} (right), when we inversely synthesize images from this classifier, the synthesized ``crop diseases" images will be falsely labeled as ``flower", conflicting with the true ``flower" images and other false ``flower" images. In another word, there will be data labeled as ``flower" but not belonging to that label. This aligns with the definition of label noise \cite{frenay2014comprehensive}, which refers to the presence of incorrect labels in a machine learning dataset. More precisely, it could be considered as label noise completely at random (NCAR) \cite{frenay2014comprehensive}, where label errors are independent of the input data's features or the actual label.
We further explore how label noise affects meta-learning. Suppose some tasks  with noisy labels are introduced into the meta-training process. If mislabeled data appears in the support set, it will hinder the adaptation phase, leading to ineffective adaptation and consequently inaccurate predictions on the query set. Additionally, if mislabeled data  is also present in the query set, it will cause misleading loss and gradients, which will be erroneously updated for the meta-learner.

\textbf{Models of deceptive accuracy can lead to synthetic tasks with feature corruption.} In this paper, we informally define ``feature corruption" as the degree of difference between synthetic features and real features. We find that images synthesized from low-accuracy models tend to display severely corrupted features. Moreover, as the accuracy of the model decreases, the degree of corruption increases. For example, as shown in \cref{fig:tdc_reason} (middle), features synthesized from a model with 95\% accuracy are quite similar to real features. However, as model accuracy decreases, synthetic features become distorted, ambiguous, and even incorrect. We investigate the reasons behind this from the data synthesis process: Given the pre-trained model $M$ and the target label $\mathbf{Y}$, we optimize the input $\mathbf{X}$ so that it can be predicted as $\mathbf{Y}$ (\textit{i.e.}, minimizing $\mathcal{L}_{\rm cls}$ in \cref{eq:purer}). If a model has high accuracy, this suggests it has likely learned the high-quality and discriminative features of $\mathbf{Y}$. Therefore, when we synthesize images from high-accuracy models, these high-quality features will be crafted into $\mathbf{X}$ to ensure $\mathbf{X}$ will be correctly predicted as $\mathbf{Y}$. However, if a model has low accuracy, ambiguous or even incorrect features learned by such models will be crafted into $\mathbf{X}$.


\textbf{Challenges in addressing TDC.} (i) Pre-evaluation of each model is impractical due to the absence of model-specific data and its time-consuming nature.
(ii) 
Manually crafting the model selection criteria lacks ground-truth supervision and may also introduce biases. Overall, we aim to develop a method to automatically select trustworthy models based on feedback from the meta-learner itself, thereby avoiding time-consuming pre-evaluations and manually crafted criteria.

\section{Methodology}
In this section, we detail our proposed trustworthy DFML (TDFML) framework. First, in \cref{sec:task_reconstruction}, we present how to generate synthetic tasks directly from pre-trained models. Next, in \cref{sec:tmi}, we present how to recall previously learned meta-knowledge by performing meta-learning with interpolated historical tasks. 
Finally, in \cref{sec:ams}, we present a mechanism to automatically prevent the false selection of untrustworthy models during the meta-training process.
A summary of the proposed methods can be found in \cref{alg:DFML}.

\subsection{Synthetic Task Reconstruction}
\label{sec:task_reconstruction}
\textbf{Synthetic data generation.} Recall that the objective of meta-learning in \cref{eq:metaobjective} is to train a meta-learner (parameterized by $\boldsymbol{\theta}_{\mathcal{A}}$) towards minimizing the expected task-level loss over a task distribution $p_{\mathcal{T}}$.
However, in DFML, for each task $\mathcal{T} \sim p_{\mathcal{T}}$, we have no access to its task-specific data but only a model pre-trained on $\mathcal{T}$. To tackle the data-free challenge, we reconstruct the task-specific data from each pre-trained model itself. Given the pre-trained model $M$ and the target label $\mathbf{Y}$ (\textit{e.g.}, $[1, 0]$ for a binary classifier), we optimize a randomly initialized generator $G(\cdot; \theta_{G})$ \cite{fang2021contrastive} to synthesize the task-specific data.
The generator takes standard Gaussian noise $\mathbf{Z}$ and target label ${\mathbf{Y}}$ as inputs, and output the data $\mathbf{X}=G(\mathbf{Z},{\mathbf{Y}};\boldsymbol{\theta}_{\mathcal{G}})$. The optimization objective of synthetic data generation can be formulated as follows.
\begin{equation}
\begin{split}
&\min_{\boldsymbol{\theta}_{G}}\mathcal{L}_{G}\triangleq \min_{\boldsymbol{\theta}_{G}}\mathcal{L}_{\rm cls}\left(M\left(\mathbf{X}\right),\mathbf{Y}\right)+\mathcal{R}(\mathbf{X}) - \mathcal{L}_{\rm task}(\mathbf{X}),\\
&\ \text{s.t.,} \  \mathbf{X}=G(\mathbf{Z},{\mathbf{Y}};\boldsymbol{\theta}_{\mathcal{G}}).
    \label{eq:tdfml}
\end{split}
\end{equation}

To maintain clarity, we omit the support set and query set, as well as the adaptation process within the parentheses of $\mathcal{L}_{\rm task}$ defined in \cref{eq:metaobjective}.
Minimizing the classification loss $\mathcal{L}_{\rm cls}$ ensures $\mathbf{X}$ can be correctly predicted as $\mathbf{Y}$ by the pre-trained model; maximizing $\mathcal{L}_{\rm task}$ lead to data of higher difficulty \cite{hu2023architecture}, thus enhancing diversity. To further improve the realism of the synthetic data, we impose a naturalness prior $\mathcal{R}$ \cite{yin2020dreaming}, which is defined as follows.
\begin{equation}
    \mathcal{R}(\mathbf{X})=\sum_l\left\|\mu^{(l)}(\mathbf{X})-\mu_{\mathrm{BN}}^{(l)}\right\|_2+\left\|\sigma^{(l)}(\mathbf{X})-\sigma_{\mathrm{BN}}^{(l)}\right\|_2 .
    \label{eq:bn}
\end{equation}
$\mu^{(l)}(\mathbf{X})$ and $\sigma^{(l)}(\mathbf{X})$ denote the mean and variance of feature maps calculated at the $l^{th}$ layer of the pre-trained model. $\mu_{\rm BN}^{(l)}$ and $\sigma_{\rm BN}^{(l)}$ denote the statistics initially stored in the $l^{th}$ batch normalization layer of the pre-trained model. Inspired by \cite{hatamizadeh2022gradvit}, for models without the batch normalization layers, we can borrow the batch normalization statistics stored in an open-source pre-trained ResNet50. Since  $\mu_{\rm BN}^{(l)}$ and $\sigma_{\rm BN}^{(l)}$ is calculated with real data, minimizing gaps in these statistics can align the distribution between the synthetic and real data, thus improving realism.

\textbf{$N$-way $K$-shot task construction.}\ After synthesizing a batch of task-specific
data $\mathbf{X}$, we split it into one support set $\mathbf{D}^s$ and one query set $\mathbf{D}^q$ based on definition of $N$-way $K$-shot (see \cref{sec:preliminary}). The overall synthetic task reconstruction algorithm is summarized in \cref{alg:generatetaskfrommodel}. 

\subsection{Meta-Learning with Task Memory Interpolation}
\label{sec:tmi}
\textbf{Task interpolation within memory buffer.} To prevent the forgetting of previously learned meta-knowledge, we introduce a strategy to replay interpolated historical tasks. We adopt reservoir sampling (RS) \cite{riemer2018learning} to dynamically maintain a task memory buffer from the sequential task stream, ensuring that previous tasks have a fair chance of being retained. Details of RS are provided in \cref{sec:app_details}.
Since meta-learning typically requires densely sampled tasks (e.g., over 60K tasks in MAML) but tasks are inherently sparsely sampled in DFML, we present two implementations of task memory interpolation:

(i) \textit{Task combination} forms a new interpolated task by combining different classes from different memory tasks. For example, given two 2-way memory tasks with label spaces of ($i_1$, $i_2$) and ($j_1$, $j_2$), an interpolated task could be ($i_1$, $j_2$). 

(ii) \textit{Task mixup} \cite{zhang2017mixup,verma2019manifold,yao2022meta} forms a new interpolated task by blending hidden representations among different memory tasks and assigning a new label to each class.  For example, given two 2-way memory tasks with label spaces of ($i_1$, $i_2$) and ($j_1$, $j_2$), an interpolated 2-way task could be $(e_1, e_2)$, where $e_1$ and $e_2$ are new classes. To construct a new class $e_1$, we first randomly sample a class pair (\textit{e.g.}, ($i_1$, $j_2$)) with each from different memory tasks. Then, we feed their data into the meta-learner and obtain their representations $(\mathbf{H}^{(l)}_{i_1}, \mathbf{H}^{(l)}_{j_2})$ at one randomly selected layer $l$. To create representations of the new class $e_1$, we blend $(\mathbf{H}^{(l)}_{i_1}, \mathbf{H}^{(l)}_{j_2})$ by performing feature-level mixup:
    $\mathbf{H}^{(l)}_{e_1}=\lambda \mathbf{H}^{(l)}_{i_1} + (1-\lambda) \mathbf{H}^{(l)}_{j_2}$,
where $\lambda \in [0, 1]$ sampled from a Beta distribution $\texttt{Beta($\alpha$,$\beta$)}$. 

For a detailed algorithm description of tasks involving more than two classes, please refer to \cref{alg:taskcombination} and \cref{alg:taskmixup} in \cref{sec:app_details}. Complexity analysis is provided in \cref{tab:complexity} in \cref{app:ex}.

\textbf{Different $\boldsymbol{\mathcal{L}_{\rm task}}$ for new tasks and interpolated memory tasks.}\
For newly synthesized tasks, we employ a distillation loss that minimizes the Kullback–Leibler (KL) divergence between the predictions of the task-specific model $\mathcal{A}[{\mathbf{D}}^{\rm s};\boldsymbol{\theta}_{\mathcal{A}}]$ and the pre-trained model $M$. This approach leverages the \textit{dark knowledge} embedded in the soft-label outputs of pre-trained models \cite{hu2023learning}.
For interpolated memory tasks (marked with $\Tilde{\ \ }$ in \cref{eq:ltask}), we use a cross-entropy (CE) loss with hard labels, due to the mismatch between label spaces of interpolated tasks and those of the pre-trained models.
\begin{equation}
\boldsymbol{\theta}_{\mathcal{A}}^{(k+1)} \leftarrow
\begin{cases}
\boldsymbol{\theta}_{\mathcal{A}}^{(k)} - \nabla_{\boldsymbol{\theta}_{\mathcal{A}}^{(k)}} {\rm KL}\left(\mathcal{A}[{\mathbf{D}}^{\rm s};\boldsymbol{\theta}_{\mathcal{A}}]({\mathbf{X}}^{q}),M({\mathbf{X}}^{q})\right),&\\
\boldsymbol{\theta}_{\mathcal{A}}^{(k)} - \nabla_{\boldsymbol{\theta}_{\mathcal{A}}^{(k)}}{\rm CE}\left(\mathcal{A}[\Tilde{\mathbf{D}}^{\rm s};\boldsymbol{\theta}_{\mathcal{A}}](\Tilde{\mathbf{X}}^{q}),\Tilde{\mathbf{Y}}^{q}\right).&
\end{cases}
 \label{eq:ltask}
\end{equation}

\textbf{Flexible choice of the meta-learner $\boldsymbol{\mathcal{A}}$.}\ Our framework is flexibily compatible to both gradient-based (\textit{e.g.}, MAML \cite{finn2017model}), and metric-based (\textit{e.g.}, ProtoNet \cite{snell2017prototypical}) meta-learning algorithms. 
In MAML, the task-specific model $\mathcal{A}_{\rm MAML}[{\mathbf{D}}^{\rm s};\boldsymbol{\theta}_{\mathcal{A}}]$ is defined as a classification model $F$ parameterized by $\boldsymbol{\phi}$. The meta-learner parameterized by $\boldsymbol{\theta}_{\mathcal{A}}$ functions as the initialization of $\boldsymbol{\phi}$. The adaptation process refers to calculating $\boldsymbol{\phi}$ by performing one-step gradient descent on the support set $\mathbf{D}^{\rm s}$ strating from $\boldsymbol{\theta}_{\mathcal{A}}$:
\begin{equation}
\begin{split}
&\mathcal{A}_{\rm MAML}[\mathbf{D}^{s};\boldsymbol{\theta}_{\mathcal{A}}]\left(\mathbf{X}^{q}\right)=F\left(\mathbf{X}^{q};\boldsymbol{\psi}\right),\\
&{\rm s.t.} \quad \boldsymbol{\psi}=\boldsymbol{\theta}_{\mathcal{A}}-\nabla_{\boldsymbol{\theta}_{\mathcal{A}}}{\rm CE}\left(F(\mathbf{X}^{s};\boldsymbol{\theta}_{\mathcal{A}}),\mathbf{Y}^{s}\right).
    \label{eq:maml}
\end{split}
\end{equation}
For ProtoNet, the task-specific model $\mathcal{A}_{\rm ProtoNet}[\mathbf{X}^{s};\boldsymbol{\theta}_{\mathcal{A}}]$ is defined as a non-parametric classifier based on Nearest Neighbor. The meta-learner functions as a feature extractor $f(\cdot;\boldsymbol{\theta}_{\mathcal{A}})$. The adaptation process refers to calculating each class center using the support set $\mathbf{X}^{s}$. The probability of a query example being classified into class $c$ is calculated as:
\begin{equation}
\left[\mathcal{A}_{\rm ProtoNet}[\mathbf{X}^{s};\boldsymbol{\theta}_{\mathcal{A}}]\left(\mathbf{X}^{q}\right)\right]_c=\frac{\exp \left(-\Vert f(\mathbf{X}^{q};\boldsymbol{\theta}_{\mathcal{A}}) - \mathbf{C}_c \Vert\right)}{\sum_{c^{\prime}} \exp \left(-\Vert f(\mathbf{X}^{q};\boldsymbol{\theta}_{\mathcal{A}}) - \mathbf{C}_{c^{\prime}} \Vert\right)},
    \label{eq:protonet}
\end{equation}
where the class center $\mathbf{C}_c$ is the average feature embedding of samples in $\mathbf{D}^{\rm s}$ that belong to class $c$. The flexible compatibility with both gradient-based and metric-based meta-learning algorithms enables integration with more advanced future algorithms for improved results.

\subsection{Automatic Model Selection}
\label{sec:ams}

\textbf{Objective.} To further handle TDC, we seamlessly incorporate a mechanism along with the meta-training process to automatically select trustworthy models before each iteration.
We parameterize the trustworthiness of given models as a learnable weight vector $\boldsymbol{W} \in \mathbb{R}^{|\mathcal{M}_{\rm pool}|}$, where $w_i$ (\textit{i.e.}, the $i^{th}$ element in $\boldsymbol{W}$ ) characterizes the trustworthiness of model $M_i$. At each iteration, we take an action selecting a batch of models $\mathcal{M}_{\rm select}$ with the highest weights from $\mathcal{M}_{\rm pool}$. We use $\pi(\mathcal{M}_{\rm select}|\mathcal{M}_{\rm pool};\boldsymbol{W})$ to denote the probability of taking this action, \textit{i.e.},  selecting $\mathcal{M}_{\rm select}$ from $\mathcal{M}_{\rm pool}$:
\begin{equation}
\pi(\mathcal{M}_{\rm select}|\mathcal{M}_{\rm pool};\boldsymbol{W})=\prod \limits_{i \in \textsc{Index}(\mathcal{M}_{\rm select})} \left(\frac{e^{w_i}}{\sum_{i^{\prime}=1}^{|\mathcal{M}_{\rm pool}|}e^{w_{i^{\prime}}}}\right),
    \label{eq:policy}
\end{equation}
where \textsc{Index}($\mathcal{M}_{\rm select}$) returns the entry indexes of $\mathcal{M}_{\rm select}$ in $\mathcal{M}_{\rm pool}$.
Our goal is to adaptively optimize the selection policy so that the meta-learner trained with the selected models $\mathcal{M}_{\rm select}$ can enhance its generalization ability, which is quantified  as the performance on a minimal set of validation tasks $\mathcal{T}^{val}=(\mathbf{D}^{val,s},\mathbf{D}^{val,q})$. Following the standard setup \cite{hu2023architecture,finn2017model}, we have access to several validation tasks with real data. The optimization objection can be formulated as follows:
\begin{equation}
\begin{split}
&\min_{\boldsymbol{W}} \frac{1}{N_v} \sum_{v=1}^{N_v} \mathcal{L}_{\rm task}\left(\mathbf{D}_{v}^{val,q} ; \mathcal{A}[\mathbf{D}_{v}^{val,s};{\boldsymbol{\theta}}_{\mathcal{A}}^{*}(\boldsymbol{W})]\right),\\
&\ \text{where} \quad {\boldsymbol{\theta}}_{\mathcal{A}}^{*}=\textsc{DFML}(\mathcal{M}_{\rm select};\boldsymbol{W}),
    \label{eq:defendingobjective}
\end{split}
\end{equation}
where $\textsc{DFML}(\mathcal{M}_{\rm select})$ return the meta-learner trained with selected models via \cref{eq:ltask}.

\begin{algorithm}[!t]
\small
\DontPrintSemicolon
\SetKwInOut{Input}{Input}\SetKwInOut{Output}{Output}\SetKwInOut{Require}{Require}
\textbf{Input: }{Pre-trained model pool $\mathcal{M}_{\rm pool}$; generator $G(\cdot;\boldsymbol{\theta}_{G})$; meta-learner $\mathcal{A}[\cdot;\boldsymbol{\theta}_{\mathcal{A}}^{(0)}]$; memory buffer $\mathcal{B}$.
}\\
\textbf{Output: }{Meta-learner $\boldsymbol{\theta}_{\mathcal{A}}$}\\
Randomly initialize $\boldsymbol{\theta}_{\mathcal{A}}^{(0)}$ and trustworthiness weights $\boldsymbol{W}^{(0)}$\\
Clear memory bank $\mathcal{B} \leftarrow [\ ]$\\
\For{ {\rm each meta-iteration} $k \leftarrow 0$ \KwTo $N$}{
\If{{{\rm \textbf{not}}}\  {\rm task memory interpolation}}{
\tcp{\small \slshape Automatic model selection}
$\boldsymbol{P}^{(k)} \leftarrow \texttt{SoftMax}(\boldsymbol{W}^{(k)})$\\
Select a batch of models $\mathcal{M}_{\rm select}$ based on $\boldsymbol{P}^{(k)}$\\
\tcp{\small \slshape Meta-train on new tasks}
Synthesize a batch of pseudo tasks ${\mathcal{T}}$\\
Update memory buffer via reservoir sampling\\
Update $\boldsymbol{\theta}_{\mathcal{A}}^{(k+1)} \leftarrow \boldsymbol{\theta}_{\mathcal{A}}^{(k)}$ via \cref{eq:ltask}\\
Calculate rewards of $\boldsymbol{\theta}_{\mathcal{A}}^{(k+1)}$ on $\mathcal{T}^{val}$\\
Update $\boldsymbol{W}^{(k+1)} \leftarrow \boldsymbol{W}^{(k)}$ via \cref{eq:policyGradient}\\
}
\Else{
\tcp{\footnotesize \slshape Meta-train on memory task}
Construct a batch of interpolated tasks $\Tilde{\mathcal{T}}$ from $\mathcal{B}$\\
Update $\boldsymbol{\theta}_{\mathcal{A}}^{(k+1)} \leftarrow \boldsymbol{\theta}_{\mathcal{A}}^{(k)}$ via \cref{eq:ltask}\\
}
}
\caption{ \textsc{Trustworthy DFML (TDFML)}}
\label{alg:DFML}
\end{algorithm}

\textbf{Optimization via RL.}\ The model selection operation (i.e.,\ $\mathcal{M}_{\rm select} \leftarrow \mathcal{M}_{\rm pool}$) is non-differentiable, making \cref{eq:defendingobjective} intractable. Therefore, we adopt the REINFORCE policy gradient method \cite{williams1992simple} to reformulate \cref{eq:defendingobjective} to a differentiable form \cref{eq:policyGradient}. Specifically, at meta-iteration $k$, we regard the average accuracy on $N_v$ validation tasks as rewards $\sum_{v=1}^{N_v}R_v^{(k)}$.  We empirically use two tasks for fast reward computation, which is significantly fewer than the number of tasks typically required for meta-training. Intuitively, if the action $\mathcal{M}_{\rm select}^{(k)}\leftarrow \mathcal{M}_{\rm pool}$ leads to an increasing reward, we will optimize the selection policy to increase the probability of taking this action $\pi(\mathcal{M}_{\rm select}^{(k)}|\mathcal{M}_{\rm pool})$, and vice versa. To reduce the gradient variance and stabilize the optimization process, we introduce the baseline function $b$ as the moving average of all past rewards:
\begin{equation}
\begin{split}
&\boldsymbol{W}^{(k+1)} \leftarrow \boldsymbol{W}^{(k)} + \\
&\nabla_{ \boldsymbol{W}^{(k)}}\left[\log \pi(\mathcal{M}_{\rm select}^{(k)}|\mathcal{M}_{\rm pool};\boldsymbol{W}^{(k)})\times \left(\frac{1}{N_v} \sum_{v=1}^{N_v} R_v^{(k)}-b\right)\right].
    \label{eq:policyGradient}
\end{split}
\end{equation}

\section{Experiments}
\label{sec:experiment}

\begin{table*}[tbp]
  \centering
   \caption{Compare with SOTA DFML baselines with trustworthy models.} 
   \scalebox{0.85}{
  \begin{tabular}{lcccccccc}
    \toprule
   \multirow{3}{*}{\textbf{Method}} & \multicolumn{4}{c}{\textbf{CIFAR-FS} \cite{bertinetto2018meta}}&\multicolumn{4}{c}{\textbf{MiniImageNet} \cite{vinyals2016matching}}\\
    \cmidrule(r){2-5}
    \cmidrule(r){6-9}
    &\multicolumn{2}{c}{\textbf{5-way 1-shot}}&\multicolumn{2}{c}{\textbf{5-way 5-shot}}&\multicolumn{2}{c}{\textbf{5-way 1-shot}}&\multicolumn{2}{c}{\textbf{5-way 5-shot}}\\
    \cmidrule(r){2-3}
    \cmidrule(r){4-5}
    \cmidrule(r){6-7}
    \cmidrule(r){8-9}
     &\textbf{BEST} &\textbf{LAST} &\textbf{BEST} &\textbf{LAST}&\textbf{BEST} &\textbf{LAST}&\textbf{BEST} &\textbf{LAST} \\
    \midrule
    RANDOM & 28.59 $\pm$ 0.56&28.59 $\pm$ 0.56&  34.77 $\pm$ 0.62&34.77 $\pm$ 0.62&  25.06 $\pm$ 0.50&25.06 $\pm$ 0.50&  28.10 $\pm$ 0.52&28.10 $\pm$ 0.52 \\
    AVERAGE &23.96 $\pm$ 0.53&23.96 $\pm$ 0.53&  27.04 $\pm$ 0.51&27.04 $\pm$ 0.51&  23.79 $\pm$ 0.48&23.79 $\pm$ 0.48&  27.49 $\pm$ 0.50&27.49 $\pm$ 0.50 \\
    DFL2L &30.43 $\pm$ 0.43& 29.35 $\pm$ 0.41 &  36.21 $\pm$ 0.51&35.28 $\pm$ 0.49&  27.56 $\pm$ 0.48&25.22 $\pm$ 0.42& 30.19 $\pm$ 0.43&28.43 $\pm$ 0.44 \\
    PURER-ANIL &35.31 $\pm$ 0.70& 26.40 $\pm$ 0.43&  51.63 $\pm$ 0.78&41.24 $\pm$ 0.68&  30.20 $\pm$ 0.61&23.05 $\pm$ 0.36&  40.78 $\pm$ 0.62& 29.60 $\pm$ 0.53 \\
    PURER-ProtoNet &36.26 $\pm$ 0.62&27.01 $\pm$ 0.58&  52.67 $\pm$ 0.68&40.53 $\pm$ 0.67&   30.46 $\pm$ 0.64&24.00 $\pm$ 0.52&  41.00 $\pm$ 0.58&31.32 $\pm$ 0.52 \\
    \midrule
    \textbf{\textsc{TDFML}-ANIL}& 40.39 $\pm$ 0.79&39.69 $\pm$ 0.79&  55.31 $\pm$ 0.75&52.92 $\pm$ 0.75&  32.58 $\pm$ 0.68&29.76 $\pm$ 0.61&  {43.63 $\pm$ 0.72}&{42.45 $\pm$ 0.67} \\
    \textbf{\textsc{TDFML}-ProtoNet} &{40.80 $\pm$ 0.78}&{40.28 $\pm$ 0.79}&  {57.11 $\pm$ 0.78}&{55.69 $\pm$ 0.76}&  {32.61 $\pm$ 0.64}& {31.97 $\pm$ 0.61}&  42.93 $\pm$ 0.65& 41.28 $\pm$ 0.64 \\
    \textbf{\textsc{TDFML}-ProtoNet} (+mixup) &\textbf{41.91 $\pm$ 0.72}&\textbf{41.78 $\pm$ 0.69}&  \textbf{58.25 $\pm$ 0.72}&\textbf{56.12 $\pm$ 0.74}&  \textbf{33.82 $\pm$ 0.68}& \textbf{32.75 $\pm$ 0.64}&  \textbf{43.97 $\pm$ 0.67}& \textbf{42.88 $\pm$ 0.62} \\
     \bottomrule
     \toprule
   \multirow{3}{*}{\textbf{Method}} & \multicolumn{4}{c}{\textbf{VGG-Flower}  \cite{flower}}&\multicolumn{4}{c}{\textbf{CUB}  \cite{cub}}\\
    \cmidrule(r){2-5}
    \cmidrule(r){6-9}
    &\multicolumn{2}{c}{\textbf{5-way 1-shot}}&\multicolumn{2}{c}{\textbf{5-way 5-shot}}&\multicolumn{2}{c}{\textbf{5-way 1-shot}}&\multicolumn{2}{c}{\textbf{5-way 5-shot}}\\
    \cmidrule(r){2-3}
    \cmidrule(r){4-5}
    \cmidrule(r){6-7}
    \cmidrule(r){8-9}
     &\textbf{BEST} &\textbf{LAST} &\textbf{BEST} &\textbf{LAST}&\textbf{BEST} &\textbf{LAST}&\textbf{BEST} &\textbf{LAST} \\
    \midrule
    RANDOM & 38.39 $\pm$ 0.71&38.39 $\pm$ 0.71&  48.18 $\pm$ 0.65& 48.18 $\pm$ 0.65 &  26.26 $\pm$ 0.48& 26.26 $\pm$ 0.48 &  29.89 $\pm$ 0.55& 29.89 $\pm$ 0.55  \\
    AVERAGE & 24.52 $\pm$ 0.46& 24.52 $\pm$ 0.46 &  32.78 $\pm$ 0.53& 32.78 $\pm$ 0.53 &  24.53 $\pm$ 0.46& 24.53 $\pm$ 0.46 &  28.00 $\pm$ 0.47& 28.00 $\pm$ 0.47  \\
    DFL2L & 40.02 $\pm$ 0.72&  38.98 $\pm$ 0.74&50.22 $\pm$ 0.68&  49.13 $\pm$ 0.70&28.33 $\pm$ 0.69&  26.01 $\pm$ 0.68&31.24 $\pm$ 0.76&29.39 $\pm$ 0.70 \\
    PURER-ANIL &51.34 $\pm$ 0.80& 45.02 $\pm$ 0.68& 67.26 $\pm$ 0.75&62.54 $\pm$ 0.72&  31.29 $\pm$ 0.64&25.05 $\pm$ 0.62&  43.34 $\pm$ 0.59& 32.08 $\pm$ 0.60 \\
        PURER-ProtoNet & 53.90 $\pm$ 0.76&47.12 $\pm$ 0.71&  68.01 $\pm$ 0.68&64.51 $\pm$ 0.67&  31.62 $\pm$ 0.63&27.23 $\pm$ 0.61& 45.36  $\pm$ 0.71&35.32 $\pm$ 0.66 \\
    \midrule
    \textbf{\textsc{TDFML}-ANIL$^{\dag}$} & 55.28 $\pm$ 0.79&54.86 $\pm$ 0.76&  69.03 $\pm$ 0.78&68.52 $\pm$ 0.75&  35.65 $\pm$ 0.72&34.32 $\pm$ 0.69&  47.24 $\pm$ 0.72&46.28 $\pm$ 0.65 \\
    \textbf{\textsc{TDFML}-ProtoNet} &{57.31 $\pm$ 0.85}&{56.79 $\pm$ 0.80}&  {71.12 $\pm$ 0.71}&{70.60 $\pm$ 0.69}&  {37.47 $\pm$ 0.73}& {36.67 $\pm$ 0.71}&  {48.68 $\pm$ 0.71}& {47.64 $\pm$ 0.68} \\
    \textbf{\textsc{TDFML}-ProtoNet} (+mixup) &\textbf{58.23 $\pm$ 0.78}&\textbf{57.56 $\pm$ 0.77}&  \textbf{72.22 $\pm$ 0.69}&\textbf{71.57 $\pm$ 0.68}&  \textbf{37.93 $\pm$ 0.70}& \textbf{37.12 $\pm$ 0.72}&  \textbf{49.54 $\pm$ 0.68}& \textbf{48.78 $\pm$ 0.71} \\
     \bottomrule
  \end{tabular}}
  \label{tab:main_no_attack}
\end{table*}
\begin{figure*}[!t]
     \vspace{-0.4cm}
    \centering
    \includegraphics[width=0.9\linewidth]{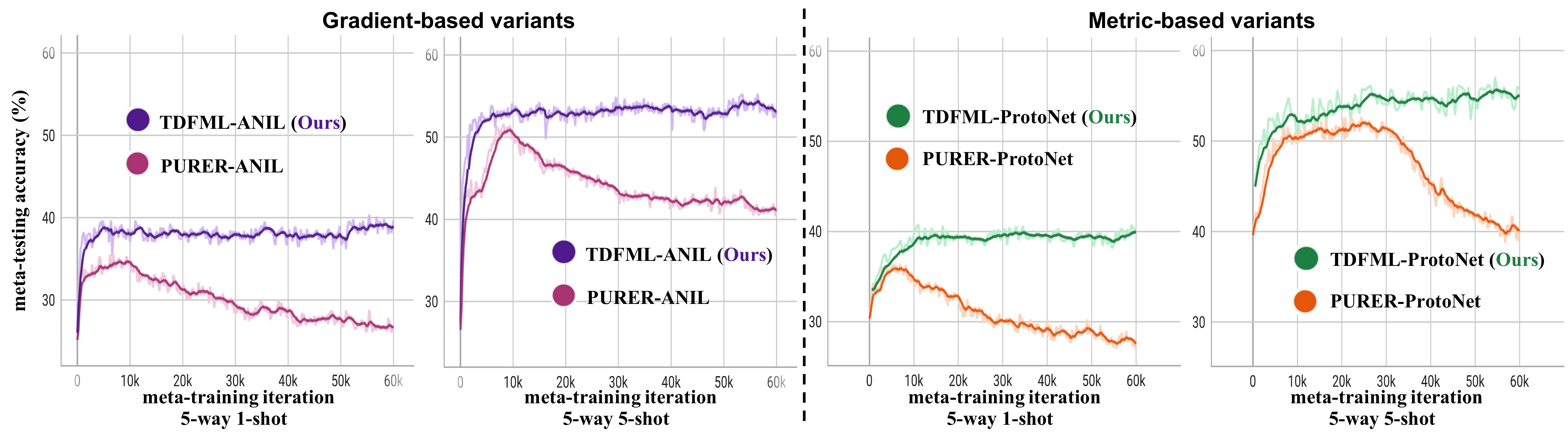}
    \vspace{-0.2cm}
    \caption{TDFML maintains stable and higher meta-testing accuracy throughout the training phase, contrasting sharply
with PURER’s degradation over time.}
    \label{fig:overfitting}
     \vspace{-0.2cm}
\end{figure*}

We compare with various state-of-the-art (SOTA) baselines on multiple datasets. We present experiments with trustworthy models in \cref{sec:expnoattack}. We further present the experiments involving exposure to untrustworthy models in \cref{sec:exattack}, which include configurations with two types and various percentages of untrustworthy models.
We perform comprehensive ablation studies and complexity analysis in \cref{sec:ablation} and \cref{app:ex}.

\textbf{Datasets.} We compare different methods on the following
most commonly used datasets for meta-learning, including: CIFAR-FS \cite{bertinetto2018meta}, MiniImageNet \cite{vinyals2016matching}, VGG-Flower \cite{flower}, and CUB-200-2011 (CUB) \cite{cub}. These datasets cover a broad range, encompassing general natural images (CIFAR-FS and MiniImageNet) as well as specific categories such as birds (CUB) and flowers (VGG-Flower). Following standard splits \cite{metadataset}, each dataset is divided into meta-training, meta-validation, and meta-testing subsets with disjoint label spaces.

\begin{figure*}[!t]
	\centering
	\includegraphics[width=0.9\linewidth]{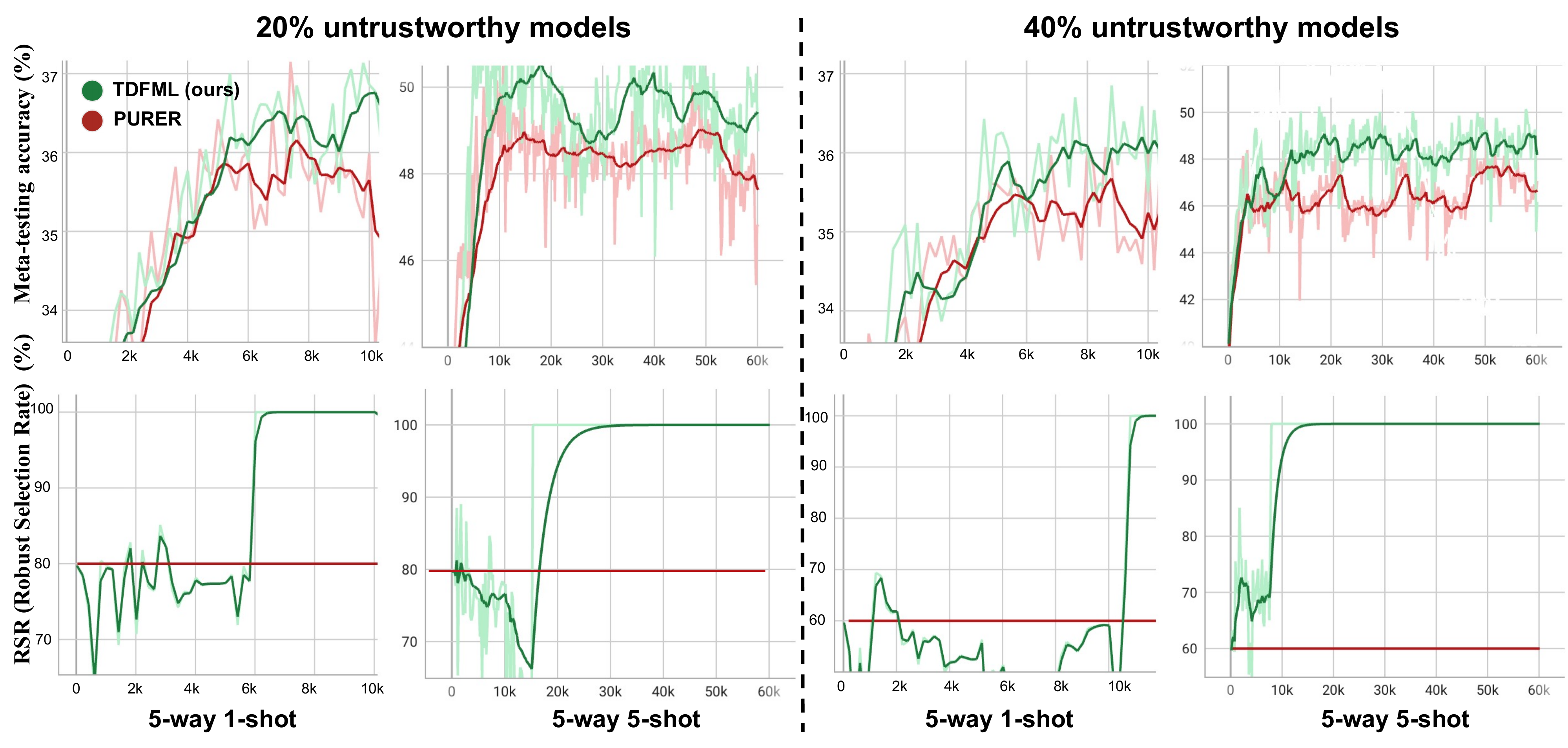}
 \vspace{-0.2cm}
	\caption{(Top) Our framework consistently improves robustness against various pollution rates of untrustworthy models.
 (Bottom) Increasing trends in RSR (up to 100\%) indicate that it  truly selects trustworthy models and avoids selecting untrustworthy ones.}
	\label{fig:ablation}
 \vspace{-0.3cm}
\end{figure*}

  \begin{table*}[!h]
    \centering
   \caption{ Results with two types of untrustworthy models across various pollution rates (CIFAR-FS, 5-way 5-shot).} 
  \scalebox{0.82}{\begin{tabular}{ccccccc}
    \toprule
\multirow{2}{*}{\textbf{Type}}&\multirow{2}{*}{\textbf{Method}}&\multicolumn{5}{c}{\textbf{Pollution Rate}}\\
\cmidrule{3-7}
&&10\% &20\%&40\%&60\%&80\%  \\
\midrule
\multirow{3}{*}{{Deceptive Label}}&PURER& 48.81$\pm$ 0.70&47.72 $\pm$ 0.78&46.28 $\pm$ 0.73&43.06 $\pm$ 0.78&41.02 $\pm$ 0.67\\
&TDFML (w/o AMS)&49.21 $\pm$ 0.74&48.11 $\pm$ 0.72&47.03 $\pm$ 0.74&43.56 $\pm$ 0.72&41.72 $\pm$ 0.70\\
&TDFML&51.08 $\pm$ 0.72$_{\textcolor{blue}{+2.27\%}}$&50.66 $\pm$ 0.72$_{\textcolor{blue}{+2.94\%}}$&49.72 $\pm$ 0.72$_{\textcolor{blue}{+3.44\%}}$&46.71 $\pm$ 0.69$_{\textcolor{blue}{+3.65\%}}$&44.84 $\pm$ 0.65$_{\textcolor{blue}{+3.82\%}}$\\
\midrule
\multirow{3}{*}{{Deceptive Accuracy}}&PURER&49.33 $\pm$ 0.71&48.09 $\pm$ 0.78&46.68 $\pm$ 0.73&44.17 $\pm$ 0.78&42.05 $\pm$ 0.67\\
&TDFML (w/o AMS)&49.94 $\pm$ 0.71&48.52 $\pm$ 0.78&47.18 $\pm$ 0.73&44.87 $\pm$ 0.78&42.95 $\pm$ 0.67\\
&TDFML&51.49 $\pm$ 0.68$_{\textcolor{blue}{+2.16\%}}$&50.96 $\pm$ 0.76$_{\textcolor{blue}{+2.87\%}}$&50.24 $\pm$ 0.74$_{\textcolor{blue}{+3.56\%}}$&47.95 $\pm$ 0.72$_{\textcolor{blue}{+3.78\%}}$&45.85 $\pm$ 0.67$_{\textcolor{blue}{+3.80\%}}$\\
\bottomrule
    \end{tabular}}
    \label{tab:main_attack}
    \vspace{-0.2cm}
\end{table*}

\textbf{Baselines.}\ We perform comprehensive experiments by
comparing to the following representative baselines: \textbf{(a) RANDOM.} Train a classification model from scratch using the support set of each meta-testing task. \textbf{(b) AVERAGE.} Average all pre-trained models and then fine-tune the average one using the support set. \textbf{(c) Baselines of data-free meta-learning:} \textbf{DFL2L} \cite{wang2022metalearning} obtains the meta-learner by fusing all pre-trained models via a hyper-network. \textbf{PURER-[$\cdot$]} \cite{hu2023architecture} meta-trains the meta-learner with synthetic tasks generated from pre-trained models. [$\cdot$] represents the integrated meta-learning algorithm such as  ANIL \cite{raghu2019rapid} or ProtoNet \cite{snell2017prototypical}.
\textbf{BiDf-MKD} \cite{hu2023learning} focuses on the black-box setting, training the meta-learner with synthetic tasks generated from APIs.  \textbf{(d) Baselines of data-free knowledge distillation:}
\textbf{MAD} \cite{do2022momentum} maintain an exponential moving average of the generator to prevent forgetting.
\textbf{LRA} \cite{patel2023learning} leverages the bi-level optimization as a regularization to prevent forgetting previous knowledge after acquiring new knowledge.
\textbf{(e) Baselines of online \& continual meta-learning:}
\textbf{FTML} \cite{finn2019online} is an online meta-learning baseline, adapting the meta-learner to the stationary task stream. \textbf{SDML} \cite{Wang_2022_CVPR} is a continual meta-learning baseline, adapting the meta-learner to sequential domains.

\textbf{Metrics.}\ {(i) BEST} denotes the meta-testing accuracy obtained by the checkpoints selected via the meta-validation subset. {(ii) LAST} denotes the meta-testing accuracy at the final iteration. {(iii) VARIATION} denotes the value of ``LAST - BEST", indicating the stability of performance.

\textbf{Implementation details.}\ 
In line with \cite{hu2023architecture}, we collect 100 models pre-trained on 100 $N$-way tasks sampled from the meta-training subset. 
Evaluation is conducted on 600 tasks sampled from the meta-testing subset.
Both the meta-learner and the pre-trained models use a Conv4 architecture \cite{wang2022metalearning,hu2023architecture} for a fair comparison with existing baselines. We use the Adam optimizer with an inner learning rate of 0.01, an outer learning rate of 0.001 for gradient-based meta-learning, and 0.001 for metric-based meta-learning and generator training. 
The coefficients for $\mathcal{R}$ and $\mathcal{L}_{\rm task}$ in \cref{eq:tdfml} are set empirically as 0.001 and 0.1, respectively. 
Following \cite{hu2023learning}, new tasks are synthesized every 200 iterations.
Unless otherwise specified, we adopt the task combination. Alternatively, we adopt a hybrid strategy denoted as ``+mixup'': performing task combination before 4000 iterations as a warmup following \cite{hu2023architecture}, then switching to task mixup. More details are provided in \cref{sec:app_details}. 

\subsection{DFML with Trustworthy Models}
\label{sec:expnoattack}

\textbf{Compare with SOTA DFML baselines with trustworthy models.}\ We first explore our proposed framework's efficacy when trained exclusively with trustworthy models. \cref{tab:main_no_attack} shows results for 5-way classification compared with existing DFML baselines. We summarize our key findings as follows. \textbf{(i)} TDFML achieves significantly {higher BEST accuracy}. Compared with the best baseline, it achieves 3.36\% $\sim$ 6.31\% performance gains for 1-shot learning and 2.97\% $\sim$ 5.58\% gains for 5-shot learning w.r.t. the BEST accuracy. \textbf{(ii)} \textsc{TDFML} achieve significantly {higher LAST accuracy and less VARIATION}. Compared with the best baseline, it achieves 7.53\% $\sim$ 12.43\% performance gains for 1-shot learning and 7.06\% $\sim$ 14.88\% gains for 5-shot learning w.r.t. the LAST accuracy. As \cref{fig:overfitting} indicates, \textsc{TDFML} maintains stable meta-testing accuracy throughout 60k meta-training iterations, contrasting sharply with PURER's degradation over time.
\textbf{(iii)} {Simply AVERAGE all models, surprisingly, even performs worse than RANDOM.}  This is due to the lack of alignment in the parameter space among models of different tasks (see \cref{tab:modelnumber} in \cref{sec:app_details} for detailed discussions).


\subsection{DFML with Untrustworthy Models}
\label{sec:exattack}

\textbf{Setup of untrustworthy models.}\ We further explore the robustness of our proposed framework when exposed to untrustworthy models. To simulate attackers injecting models with deceptive labels, we pre-train models on various datasets including EuroSAT \cite{helber2019eurosat}, ISIC \cite{tschandl2018ham10000}, ChestX \cite{wang2017chestx}, Omniglot \cite{lake2015human}, and MNIST \cite{lecun2010mnist}, but falsely claim they are pre-trained on CIFAR-FS. To simulate attackers injecting models with deceptive accuracy, we randomly alter the pre-training configuration by varying the learning rate \(\{1, 0.1, 0.001, 0.00001\}\) and the number of epochs \(\{10, 20, \ldots, 60\}\). We inject models with testing accuracies below 25\% to the model pool. Additionally, we introduce a metric termed ``pollution rate" to quantify the proportion of untrustworthy models in the model pool.


\textbf{Compare with SOTA DFML baselines with untrustworthy models.}\ 
\cref{tab:main_attack} and \cref{fig:ablation} show that our framework consistently improves performance across various pollution rates and untrustworthy models, with gains ranging from 2.16\% to 3.82\%. Notably, within a certain range, as the pollution rate increases, TDFML with automatic model selection (AMS) demonstrates more improvements, suggesting the AMS does enhance the framework's robustness to untrustworthy models. Other baselines, which have not considered the presence of untrustworthy models, are significantly undermined in their effectiveness by the inclusion of such models.

\textbf{Analysis of the learnable trustworthiness weights.}\ To analyze our effectiveness in selecting trustworthy models, 
we introduce an indicator named Robust Sampling Rate (RSR):
${{\rm RSR}= \textstyle \sum_{i \in \texttt{INDEX}(\mathcal{M}_{\rm trustworthy})}\left(\frac{e^{w_i}}{\sum_{i^{\prime}=1}^{|\mathcal{M}_{\rm pool}|}e^{w_{i^{\prime}}}}\right)}$, 
where \texttt{INDEX}($\mathcal{M}_{\rm trustworthy}$) returns the entry indexes of all trustworthy models in $\mathcal{M}_{\rm pool}$.
RSR is a metric that quantifies the effectiveness of the model selection by measuring the percentage of selected models that are trustworthy. As depicted in \cref{fig:ablation} (Bottom), the RSR increases over time to 100\%, indicating that it  truly selects trustworthy models and avoids selecting untrustworthy ones.


\subsection{Ablation Studies}
\label{sec:ablation}

\textbf{Effect of each component.}\ \cref{tab:ablation} analyzes the effectiveness of each component evaluated on CIFARF-FS with trustworthy models. 
The ``\underline{V}anilla'' (V) approach, which utilizes synthetic tasks without a memory buffer, exhibits significant performance degradation due to the TDS issue. Solely incorporating a \underline{M}emory buffer (M) does not yield satisfactory benefits due to the inherent sparse task sampling bottleneck discussed in \cref{sec:related}.
The addition of task \underline{I}nterpolation (I) results in significantly higher BEST and LAST performance. 
Utilizing \underline{S}oft-label (S) supervision from pre-trained models (\cref{eq:ltask}) further enhances the framework's performance. With all components, we achieve the best with a boosting improvement, demonstrating the effectiveness of the joint schema. 

\begin{table}[!h]
\vspace{-0.3cm}
\centering
   \caption{Effect of each component.} 
  \scalebox{0.85}{\begin{tabular}{lcccc}
    \toprule
     \multirow{2}{*}{\textbf{Method}} & \multicolumn{2}{c}{\textbf{5-way 1-shot}}&\multicolumn{2}{c}{\textbf{5-way 5-shot}}\\
    \cmidrule(r){2-3}
    \cmidrule(r){4-5}
       &BEST &LAST &BEST &LAST \\
    \midrule
     V & 35.34 $\pm$ 0.68 &  26.76 $\pm$ 0.64&  50.31 $\pm$ 0.72&  39.22 $\pm$ 0.74\\
     V + M & 36.02 $\pm$ 0.70 &  28.33 $\pm$ 0.68& 51.23 $\pm$ 0.72&  41.54 $\pm$ 0.70\\
     V + M + I & 38.76 $\pm$ 0.74 &  38.42 $\pm$ 0.72&  54.91 $\pm$ 0.74&  53.52 $\pm$ 0.74\\
     V + M +I + S& \textbf{40.80 $\pm$ 0.78} &  \textbf{40.28 $\pm$ 0.79}&  \textbf{57.11 $\pm$ 0.78}&  \textbf{55.69 $\pm$ 0.76}\\
     \bottomrule
  \end{tabular}}
  \label{tab:ablation}
\end{table}

\textbf{Comparison with data-free knowledge distillation (DFKD).} DFKD leverages the synthetic data to transfer knowledge from the teacher model to the lightweight student model. 
Recent studies \cite{do2022momentum,yu2023data} attempt to address the distribution shift issue in DFKD. Although they do not target DFML, to show the effectiveness of our method, we compare to the SOTA methods aiming to address distribution shift, including MAD \cite{do2022momentum} and LRA \cite{yu2023data}. The results on CIFAR-FS are shown in \cref{tab:dfkd}.
Our method substantially outperforms these baselines. We believe that the effectiveness of the moving average used in MAD significantly decays over a long-term task stream in DFML. Similarly, LRA faces limitations where the bi-level optimization is confined to short-term remembering.

\begin{table}[!h]
\vspace{-0.3cm}
\centering
   \caption{Comparison with data-free knowledge distillation.} 
  \scalebox{0.85}{\begin{tabular}{lcccc}
    \toprule
     \multirow{2}{*}{\textbf{Method}} & \multicolumn{2}{c}{\textbf{5-way 1-shot}}&\multicolumn{2}{c}{\textbf{5-way 5-shot}}\\
    \cmidrule(r){2-3}
    \cmidrule(r){4-5}
       &BEST &LAST &BEST &LAST \\
    \midrule
     MAD \cite{do2022momentum}& 29.24 $\pm$ 0.72 &   26.16 $\pm$ 0.75& 35.19 $\pm$ 0.69 &  32.57 $\pm$ 0.63\\
     LRA \cite{yu2023data}& 30.12 $\pm$ 0.68 &  28.03 $\pm$ 0.62& 36.41 $\pm$ 0.71&  33.78 $\pm$ 0.65\\
     TDFML (ours)& \textbf{41.91 $\pm$ 0.72}&\textbf{41.78 $\pm$ 0.69}&  \textbf{58.25 $\pm$ 0.72}&\textbf{56.12 $\pm$ 0.74}\\
     \bottomrule
  \end{tabular}}
  \label{tab:dfkd}
\end{table}

\textbf{Comparison with online \& continual meta-learning.}
To confirm our superiority in dealing with sequential task streams, we compared our method with online and continual meta-learning baselines, including FTML \cite{finn2019online} and SDML \cite{wang2022meta}. The results on CIFAR-FS, presented in \cref{tab:sequential}, show that our method outperforms these baselines. Our superiority lies in our adaptability to sparsely-sampled task distributions, where baselines fail to achieve satisfactory results due to the lack of sufficient information from previous distributions to effectively recall previously learned knowledge.



\begin{table}[!h]
\vspace{-0.3cm}
\centering
   \caption{Comparison with online \& continual meta-learning.} 
  \scalebox{0.85}{\begin{tabular}{lcccc}
    \toprule
     \multirow{2}{*}{\textbf{Method}} & \multicolumn{2}{c}{\textbf{5-way 1-shot}}&\multicolumn{2}{c}{\textbf{5-way 5-shot}}\\
    \cmidrule(r){2-3}
    \cmidrule(r){4-5}
       &BEST &LAST &BEST &LAST \\
    \midrule
     FTML \cite{do2022momentum}& 34.12 $\pm$ 0.68 &   30.35 $\pm$ 0.73& 47.34 $\pm$ 0.66 &  41.62 $\pm$ 0.63\\
     SDML \cite{yu2023data}& 35.15 $\pm$ 0.69 &  32.15 $\pm$ 0.67& 49.25 $\pm$ 0.71&  43.63 $\pm$ 0.62\\
     TDFML (ours)& \textbf{41.91 $\pm$ 0.72}&\textbf{41.78 $\pm$ 0.69}&  \textbf{58.25 $\pm$ 0.72}&\textbf{56.12 $\pm$ 0.74}\\
     \bottomrule
  \end{tabular}}
  \label{tab:sequential}
\end{table}


\section{Conclusion}
For the first time, we conduct a systematic robustness analysis of DFML, focusing on its failure modes and vulnerability to potential attacks. We identify two critical but previously overlooked vulnerabilities: Task-Distribution Shift (TDS) and Task-Distribution Corruption (TDC), detailing their causes, impacts, and associated challenges. We believe such a robustness analysis of DFML is crucial since algorithms often need to operate in uncertain and complex environments. To address these vulnerabilities, we propose a trustworthy DFML framework comprising three components: synthetic task reconstruction, meta-learning with task memory interpolation, and automatic model selection. 
Extensive experiments across four datasets with two types of untrustworthy models demonstrate the superior effectiveness of our method in significantly enhancing the robustness.



{
\bibliographystyle{IEEEtran}
\bibliography{files/ref}
}

\section{Biography Section}
 

\vspace{-1.4cm}
\begin{IEEEbiography}
[{\includegraphics[width=0.9in,height=1.25in,clip,keepaspectratio]{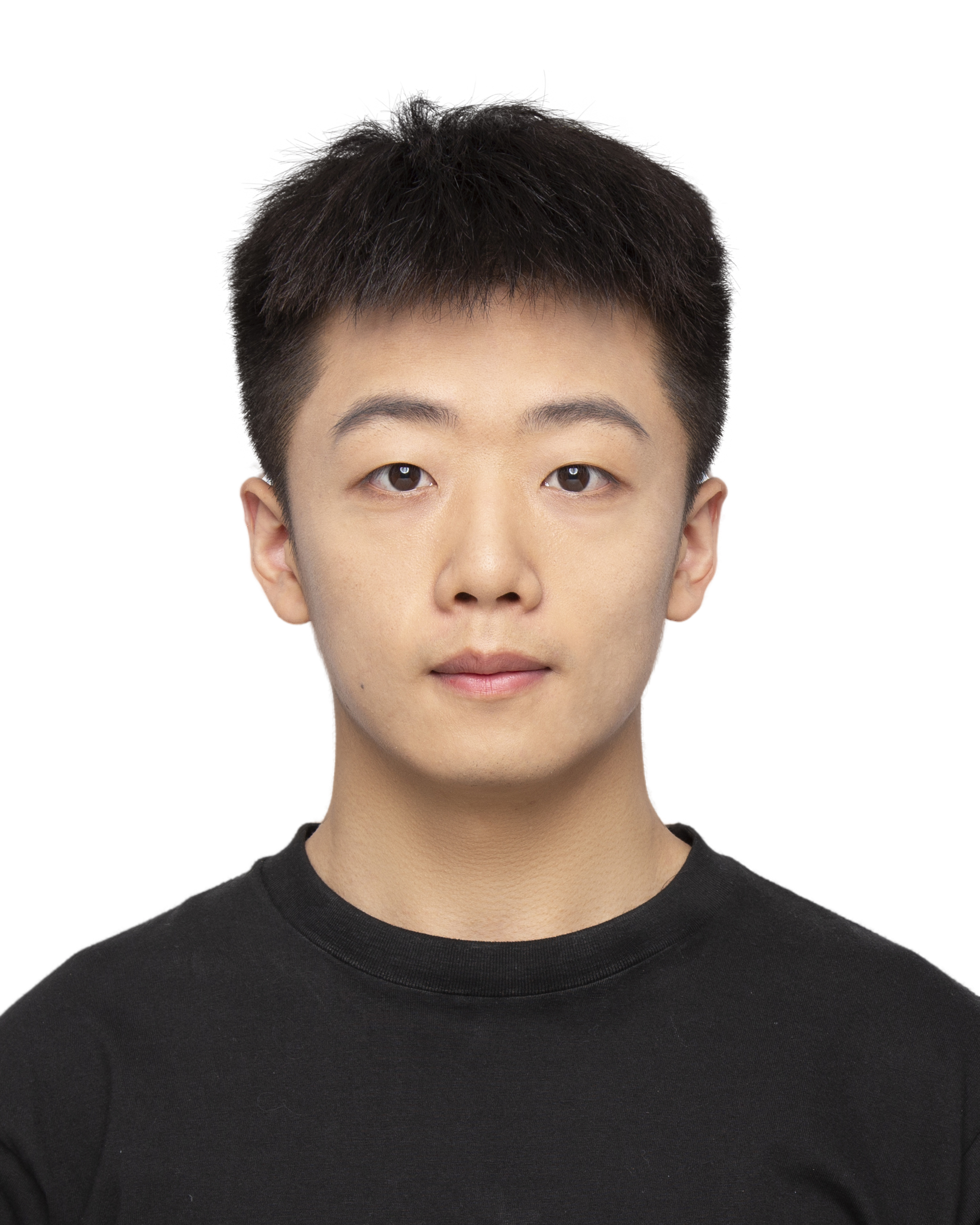}}]{Zixuan Hu}is currently pursuing a PhD in the College of Computing and Data Science at Nanyang Technological University (NTU), Singapore. He received his B.E. degree from the Harbin Institute of Technology (HIT) in 2021, and his M.S. degree from the Shenzhen International Graduate School at Tsinghua University (THU) in 2024. His research interests include machine learning and computer vision, with a particular focus on meta-learning, synthetic data generation, and model inversion.
\end{IEEEbiography}
\vspace{-1.4cm}
\begin{IEEEbiography}
[{\includegraphics[width=0.9in,height=1.05in,clip,keepaspectratio]{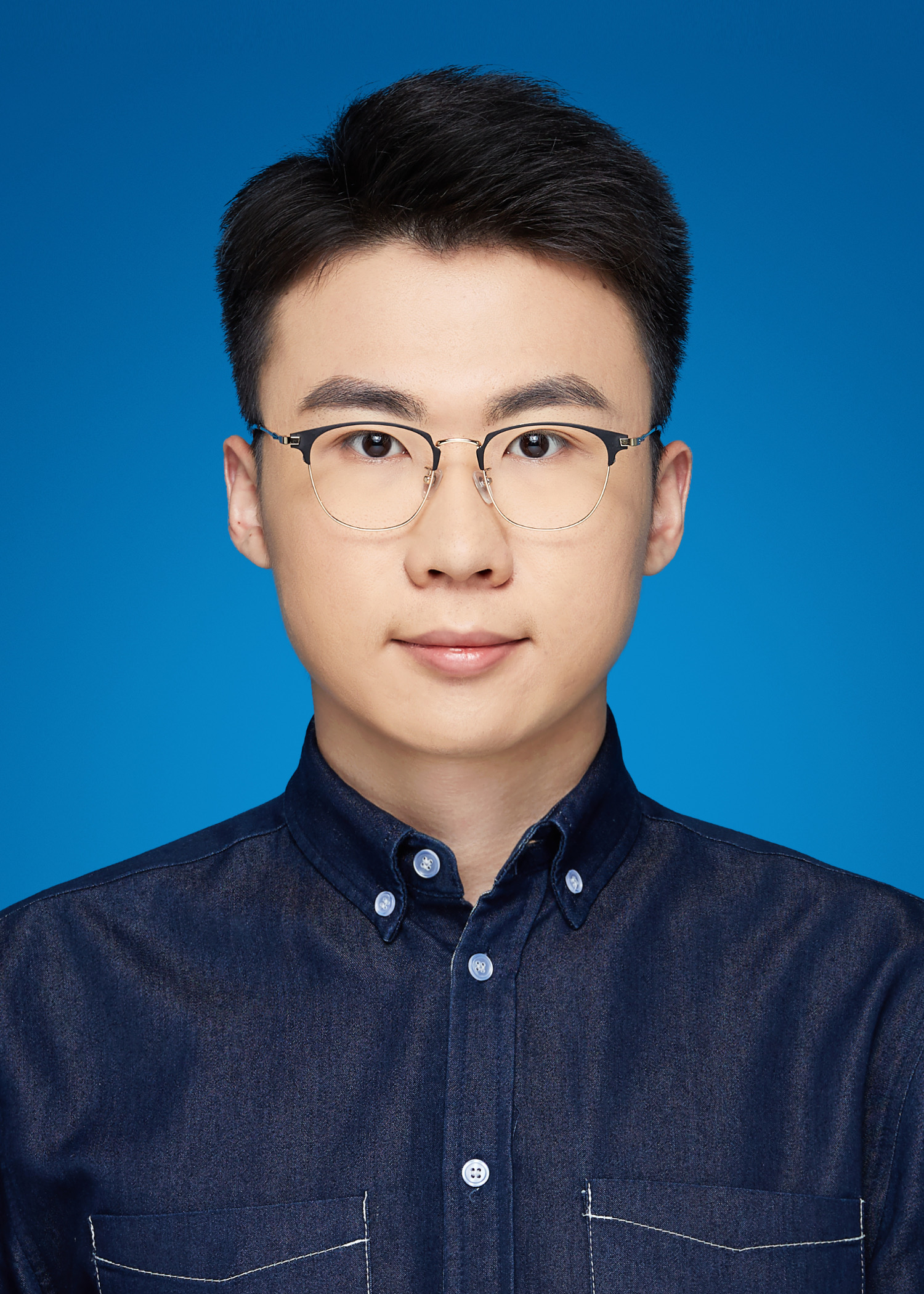}}]{Yongxian Wei} received the B.Sc. degree in computer
science and technology from Nanjing University of
Science and Technology, China in 2023. Currently,
he is a master student in Shenzhen International
Graduate School, Tsinghua University, China. His
research interests include few-shot learning, data-free
learning and machine learning.
\end{IEEEbiography}
\vspace{-1.5cm}

\begin{IEEEbiography}
[{\includegraphics[height=1.16in,clip,keepaspectratio]{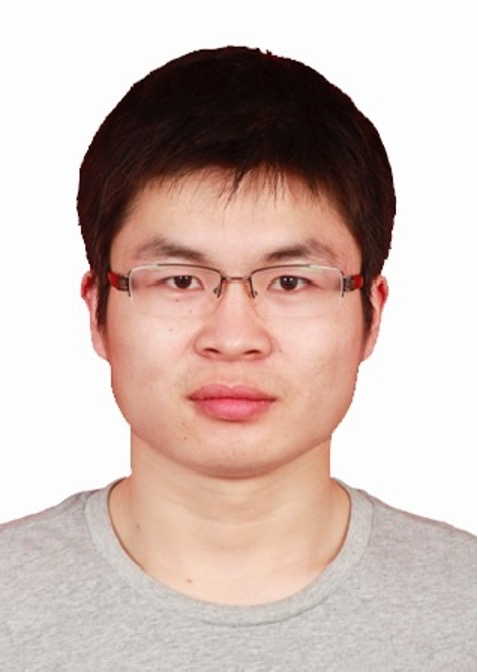}}]
{Li Shen} is currently an associate professor at Sun
Yat-sen University. Previously, he was a research
scientist at JD Explore Academy, Beijing, and a
senior researcher at Tencent AI Lab, Shenzhen. He
received his bachelor’s degree and Ph.D. from the
School of Mathematics, South China University of
Technology. His research interests include theory and
algorithms for nonsmooth convex and nonconvex
optimization, and their applications in trustworthy artificial intelligence, deep learning, and reinforcement
learning. He has published more than 100 papers in
peer-reviewed top-tier journal papers (JMLR, IEEE TPAMI, IJCV, IEEE TSP,
IEEE TIP, IEEE TKDE, etc.) and conference papers (ICML, NeurIPS, ICLR,
CVPR, ICCV, etc.). He has also served as the senior program committee for
AAAI and area chairs for ICML, ICLR, ICPR, and ACML.
\end{IEEEbiography}
\vspace{-1.2cm}

\begin{IEEEbiography}
[{\includegraphics[width=0.8in,height=1.05in,clip,keepaspectratio]{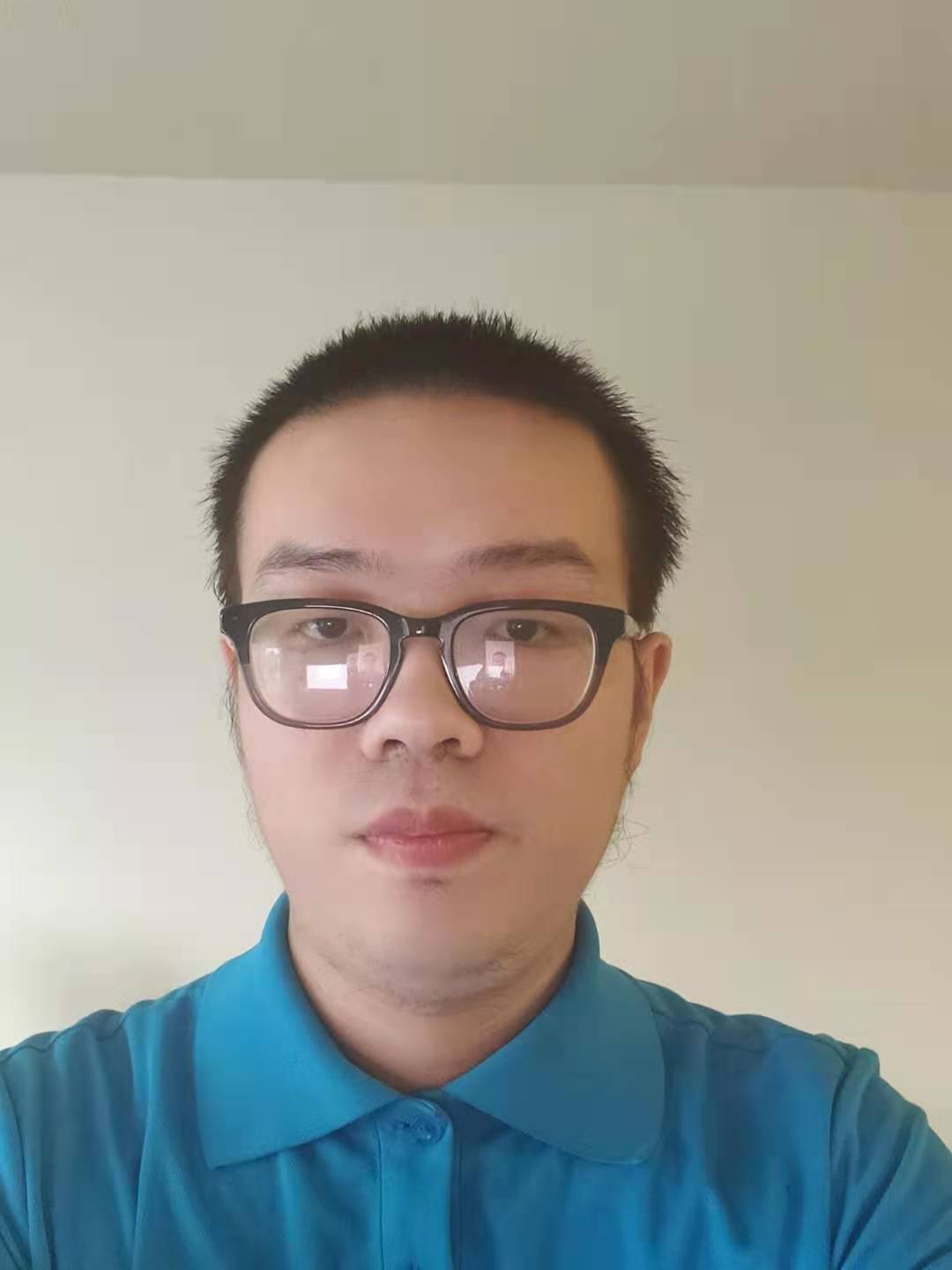}}]{Zhenyi Wang} is currently an assistant professor in the department of computer science and AI Institute, University of Central Florida, USA. He was a Postdoctoral Associate
at University of Maryland, College Park. He received
PhD degree from The State University of New York
at Buffalo. He received the Bachelor degree from
Northeastern University, China. His research interests
include continual learning, trustworthy ML and data-efficient learning.
\end{IEEEbiography}
\vspace{-1.5cm}

\begin{IEEEbiography}
[{\includegraphics[width=0.85in,height=1.25in,clip,keepaspectratio]{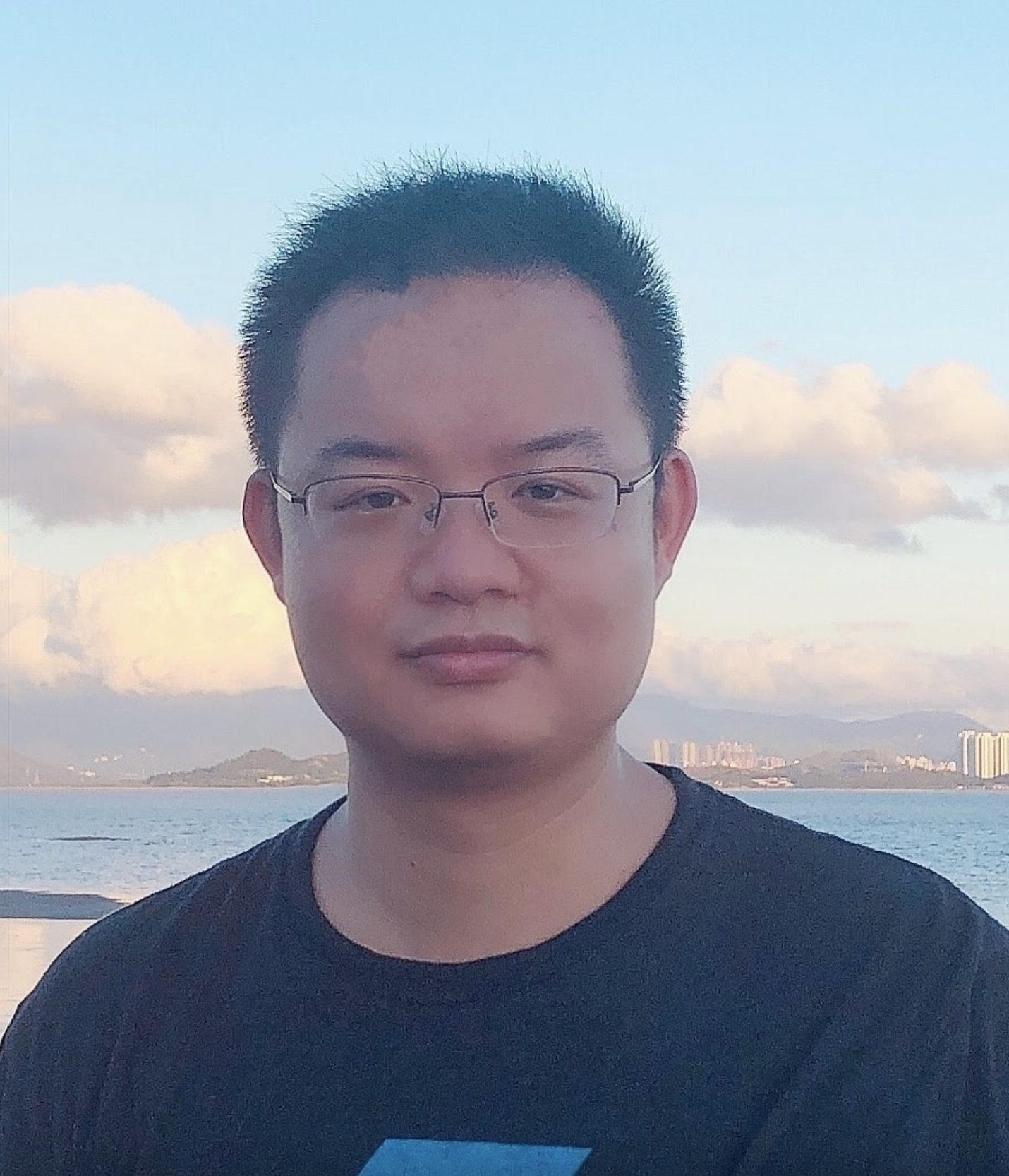}}]{Baoyuan Wu} 
is a  Tenured Associate Professor, and Assistant Dean (research) of School of Data Science, the Chinese University
of Hong Kong, Shenzhen (CUHK-Shenzhen).
He received
the PhD degree from the National Laboratory
of Pattern Recognition, Institute of Automation,
Chinese Academy of Sciences. His research interests are AI security and
privacy, machine learning, computer vision and optimization. 
He has published more than 70 top-tier
conference and journal papers, including IEEE TPAMI, IJCV, NeurIPS, ICML, CVPR,ICLR, and one
paper was selected as the Best Paper Finalist of CVPR 2019.
He is
serving as an associate editor of Neurocomputing, organizing chair of PRCV
2022, area chair of NeurIPS, ICLR, ICML, and
AAAI.

\end{IEEEbiography}
\vspace{-1.2cm}

\begin{IEEEbiography}
[{\includegraphics[width=0.8in,height=1.05in,clip,keepaspectratio]{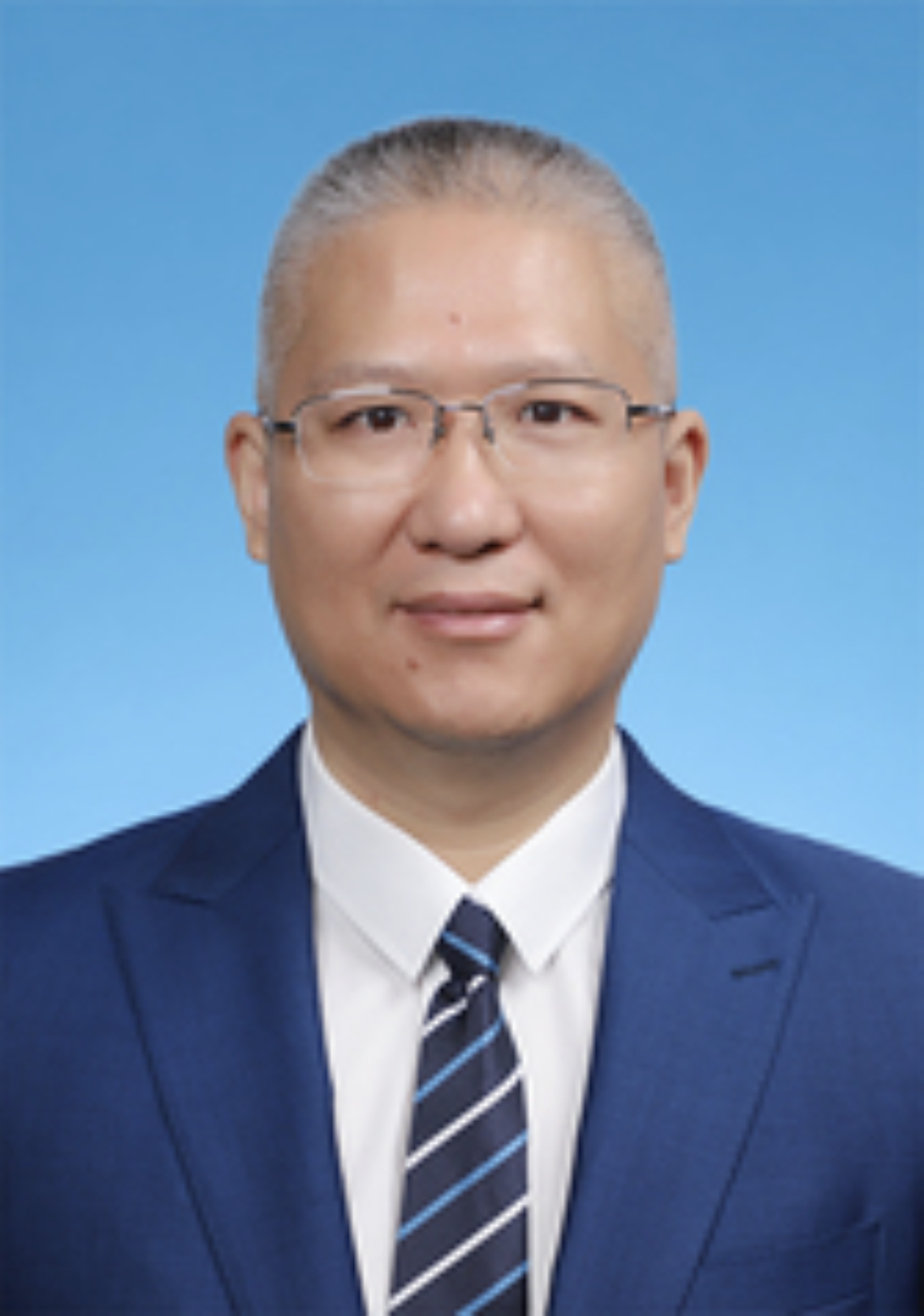}}]{Chun Yuan} (Senior Member, IEEE) received the
M.S. and Ph.D. degrees from the Department of Computer Science and Technology, Tsinghua University,
in 1999 and 2003, respectively. He is currently a
Professor with the Tsinghua Shenzhen International
Graduate School, Tsinghua University. His research
interests include computer vision, machine learning,
and reinforcement learning
\end{IEEEbiography}
\vspace{-1.5cm}

\begin{IEEEbiography}
[{\includegraphics[width=0.8in,height=1.05in,clip,keepaspectratio]{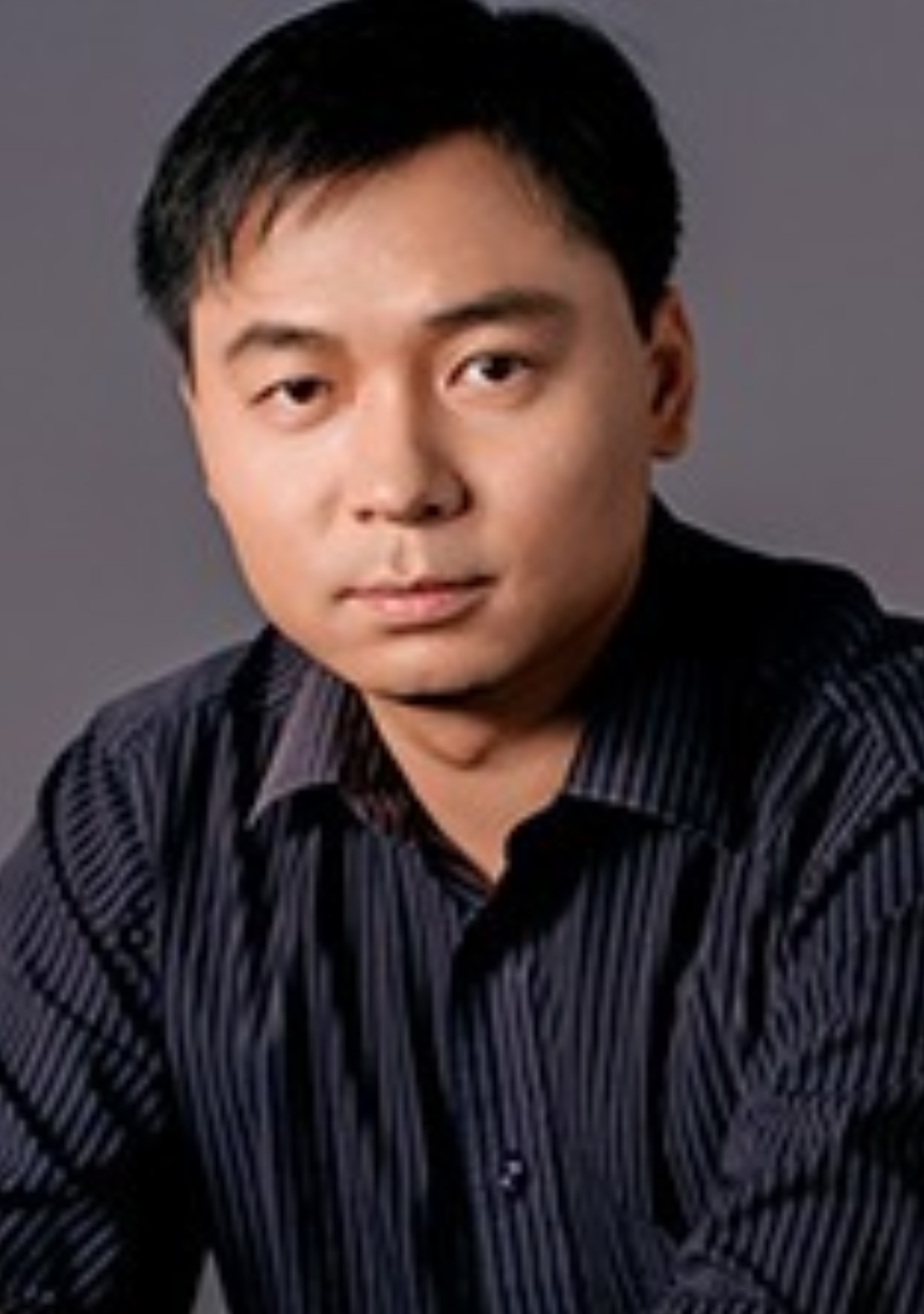}}]{Dacheng Tao} (Fellow, IEEE) is currently a Distinguished University
Professor in the College of Computing and Data
Science at Nanyang Technological University. He
mainly applies statistics and mathematics to artificial
intelligence and data science, and his research is
detailed in one monograph and over 200 publications
in prestigious journals and proceedings at leading
conferences, with best paper awards, best student
paper awards, and test-of-time awards. His publications have been cited over 130K times and he has
an h-index 170+ in Google Scholar. He received the
2015 and 2020 Australian Eureka Prize, the 2018 IEEE ICDM Research
Contributions Award, and the 2021 IEEE Computer Society McCluskey
Technical Achievement Award. He is a Fellow of the Australian Academy
of Science, AAAS, ACM and IEEE.
\end{IEEEbiography}



\clearpage
\appendix
\label{sec:appendix}
\subsection{More Discussions}
\label{app:addtionaldiscussions}
\textbf{Limitations and future work.}\ A limitation of our proposed TDFML is that we focus on meta-learning from models trained on the same type of task, specifically classification. This reflects a long-standing limitation in meta-learning: the focus on learning priors from the same type of tasks to facilitate solving unseen tasks. Until recently, Meta Omnium \cite{bohdal2023meta} proposes the concept of multi-task meta-learning, where tasks are diverse and different, including classification, semantic segmentation, pose estimation, and beyond. Unfortunately, they do not propose methods to address multi-task meta-learning. This is because learning a general meta-learner that can generalize to different types of tasks is very challenging due to differences in model architecture design and other factors. In the data-free scenario, exploring multi-task DFML is practically significant, \textit{i.e.}, meta-learning models pre-trained on various types of tasks to facilitate learning of unseen tasks. We leave this limitation to be addressed in future work.

\textbf{Broader impacts.} We explored the potential positive and negative societal impacts of our framework.
Regarding positive societal impacts, our method can effectively defend against attacks on DFML, \textit{i.e.}, attackers can actively send untrustworthy models with deceptive model information to DFML users. For negative societal impacts, the model inversion technique used in our method may leak information about the original training data. However, the objective of data generation is to minimize the classification loss, not to accurately generate high-quality images like the works in the generation community. Therefore, the synthetic data will not accurately  reproduce the original images.

\textbf{The presence of untrustworthy models can somewhat mitigate TDS.} By comparing the left and right panels of \cref{fig:limitation}, we observe that the appearance of untrustworthy models can alleviate the performance drop over time caused by TDS. However, this also significantly reduces peak accuracy. We explain this phenomenon from the perspective of distributionally robust optimization (DRO) \cite{rahimian2019distributionally}. The basic idea of DRO is to construct an ambiguity set of probability distributions and optimize the worst-case performance within this set. DRO has been used to improve robustness against group shift \cite{sagawa2019distributionally}, subpopulation shift \cite{zhai2021doro}, and class imbalances \cite{wang2022improving}. In our scenario, these untrustworthy models can be seen as worst-case samples from a largely shifted task distribution, thus improving the meta-learner's robustness to TDS to some extent. However, the severity of these untrustworthy models is not controlled, thereby severely compromising the peak performance of DFML.

\textbf{Initialization strategy.} We initialize the trustworthiness weights of the pre-trained models with random values to avoid introducing bias and to encourage initial exploration. We compare this choice with two alternative initialization strategies, including uniform initialization (where all models start with equal weights) and heuristic-based initialization using external validation. Our rationale for choosing random initialization can be summarized as follows:
\begin{itemize}
    \item First, random initialization avoids introducing bias. In contrast, heuristic-based initialization, which relies on external validation, may not be available and mislead the following model selection process. Instead, random initialization avoids introducing bias without relying on unreliable external verification.
    \item Second, random initialization encourages initial exploration. Compared to uniform initialization, random initialization introduces slight stochastic differences among the models' weights. This breaks the symmetry  and allows the algorithm to rapidly identify promising models based on early reward signals \cite{houthooft2016vime}. In contrast, uniform initialization can lead to slower adaptation because it does not introduce any initial diversity for the algorithm to exploit in the early stages.
\end{itemize}

\textbf{Comparison among ML, CL, OML, CML, and DFML.} We provide a more formal and detailed comparison among meta-learning, continual learning, online meta-learning, continual meta-learning, and data-free meta-learning. We first present a concise formulation for each of these relevant fields below to better illustrate their similarities and differences.
\begin{itemize}
  \item \textbf{Meta-learning (ML) \cite{finn2017model}.} The goal of ML is to train a meta-learner (parameterized by $\boldsymbol{\theta}_{\mathcal{A}}$) for the best performance over a distribution of tasks $p_{\mathcal{T}}$, so that it can acquire the adaptable meta-knowledge to similar but unseen tasks sampled from the same task distribution. The optimization objective can be formulated as:
\begin{equation}
\min_{\boldsymbol{\theta}_{\mathcal{A}}}\mathbb{E}_{\mathcal{T} \sim {p}_{\mathcal{T}}}\mathcal{L}_{\rm task}\left({\mathbf{D}}^{\rm q};\mathcal{A}[{\mathbf{D}}^{\rm s};\boldsymbol{\theta}_{\mathcal{A}}]\right),
\end{equation}
Here, ${\mathcal{T}}$ represents the task sampled from the task distribution ${p}_{\mathcal{T}}$. Each task $\mathcal{T}$ comprises a task-specific training set (\textit{i.e.,} support set) ${\mathbf{D}}^{\rm s}=(\mathbf{X}^{\rm s}, \mathbf{Y}^{\rm s})$ and a task-specific testing set (\textit{i.e.,} query set) ${\mathbf{D}}^{\rm q}=(\mathbf{X}^{\rm q}, \mathbf{Y}^{\rm q})$.  $\mathcal{A}[{\mathbf{D}}^{\rm s};\boldsymbol{\theta}_{\mathcal{A}}]$ denote the task-specific adaption process: the meta-learner $\boldsymbol{\theta}_{\mathcal{A}}$ uses the support set ${\mathbf{D}}^{\rm s}$ to perform adaptation (\textit{e.g.,} a few steps of gradient descent) and obtain the task-specific model $\mathcal{A}\left[\mathbf{D}^{\mathrm{s}} ; \boldsymbol{\theta}_{\mathcal{A}}\right]$.
$\mathcal{L}_{\rm task}$ refers to the task-level loss, which assesses the performance of the task-specific model on the query set ${\mathbf{D}}^{\rm q}=(\mathbf{X}^{\rm q}, \mathbf{Y}^{\rm q})$ of each task. 

\item \textbf{Continual learning (CL) \cite{wang2024comprehensive}.}  
The goal of CL is to learn on a sequence
of tasks $\{\mathcal{T}_1, \mathcal{T}_2, \cdots, \mathcal{T}_T\}$ without forgetting the knowledge
on previous tasks. It can be formulated with the following
optimization objective. Suppose when learning task $t$, the
goal is to minimize the risk on all the seen tasks so far, \textit{i.e.},
\begin{equation}
\min_{\boldsymbol{\phi}} \sum_{t'=1}^t \mathcal{L}_{\rm task}(\mathbf{D}_{t'}^{\rm s}; \boldsymbol{\phi}),
\end{equation}
where $\mathcal{L}_{\rm task}$ refers to the task-level loss. $\boldsymbol{\phi}$ is the model parameters. $\mathbf{D}_{t}^{\rm s}$ represents the training data of task $t$.

\item  \textbf{Online meta-learning (OML) \cite{finn2019online}.} OML shares the same objective as ML. However, in OML, the offline set of meta-training tasks $\{\mathcal{T}_t\}_{t=1}^T$ is replaced by a stream of tasks $\{\mathcal{T}_1, \mathcal{T}_2, \cdots, \mathcal{T}_T\}$. The evaluation protocol in OML also differs: instead of relying on a separate meta-test set, OML directly uses the performance of the meta-learner in each new meta-training task as the evaluation metric.
\item  \textbf{Continual meta-learning (CML) \cite{son2023meta}.} CML has the same setting as OML: it replaces the offline meta-training task set $\{\mathcal{T}_t\}_{t=1}^T$ with a stream of tasks $\{\mathcal{T}_1, \mathcal{T}_2, \cdots, \mathcal{T}_T\}$. The key difference lies in the stationarity of the meta-training stream. In OML, the task distribution is stationary (tasks are sequentially sampled from the same distribution), while in CML, the task distribution shifts during meta-training (tasks are sequentially sampled from different distributions). Furthermore, the evaluation protocol for CML is the same as that of traditional ML.
\item \textbf{Data-free meta-learning (DFML) \cite{hu2023architecture}.} DFML also shares the same objective as ML: learning adaptable meta-knowledge to new tasks. However, in DFML, we do not have access to the datasets of meta-training tasks and instead rely solely on the corresponding pre-trained models (\textit{i.e.}, a collection of pre-trained models $\{M_t\}_{t=1}^T$). The evaluation protocol for DFML is the same as that of traditional ML.
\end{itemize}

We summarize differences among ML, CL, OML, CML, and DFML in \cref{tab:comparison} and their source of forgetting in \cref{tab:forgetting}.

\begin{table*}[!h]
\centering
\caption{Comparison of ML, CL, OML, CML, and DFML.}
\label{tab:comparison}
\resizebox{\textwidth}{!}{ 
\begin{tabular}{@{}l c c c@{}}
\toprule
\textbf{Problem} & \textbf{Goal} & \multicolumn{2}{c}{\textbf{Setting}} \\
\cmidrule(lr){3-4}
 & & \textbf{Training} & \textbf{Testing} \\
\midrule
\textbf{CL} & Learn without forgetting previous tasks & $\{\mathcal{T}_t\}_{t=1}^T$, where $\mathcal{T}=(\mathbf{D}^{\rm s})$ & $\{\mathcal{T}_t\}_{t=1}^T$, where $\mathcal{T}=(\mathbf{D}^{\rm q})$ \\
\textbf{ML} & Learn adaptable meta-knowledge to new tasks & $\{\mathcal{T}_t\}_{t=1}^T$, where $\mathcal{T}=(\mathbf{D}^{\rm s}, \mathbf{D}^{\rm q})$ & $\{\mathcal{T}_t^{\rm test}\}_{t=1}^T$, where $\mathcal{T}^{\rm test}=(\mathbf{D}^{\rm s}_{\rm test}, \mathbf{D}^{\rm q}_{\rm test})$ \\
\textbf{OML} & Learn adaptable meta-knowledge to new tasks & $\{\mathcal{T}_1, \mathcal{T}_2, \cdots, \mathcal{T}_T\}$, where $\mathcal{T}=(\mathbf{D}^{\rm s}, \mathbf{D}^{\rm q})$ & $\{\mathcal{T}_1, \mathcal{T}_2, \cdots, \mathcal{T}_T\}$, where $\mathcal{T}=(\mathbf{D}^{\rm s}, \mathbf{D}^{\rm q})$\\
\textbf{CML} & Learn adaptable meta-knowledge to new tasks & $\{\mathcal{T}_1, \mathcal{T}_2, \cdots, \mathcal{T}_T\}$, where $\mathcal{T}=(\mathbf{D}^{\rm s}, \mathbf{D}^{\rm q})$ & $\{\mathcal{T}_t^{\rm test}\}_{t=1}^T$, where $\mathcal{T}^{\rm test}=(\mathbf{D}^{\rm s}_{\rm test}, \mathbf{D}^{\rm q}_{\rm test})$ \\
\textbf{DFML} & Learn adaptable meta-knowledge to new tasks & $\{\mathcal{T}_t\}_{t=1}^T$, where $\mathcal{T}=(M)$ & $\{\mathcal{T}_t^{\rm test}\}_{t=1}^T$, where $\mathcal{T}^{\rm test}=(\mathbf{D}^{\rm s}_{\rm test}, \mathbf{D}^{\rm q}_{\rm test})$ \\
\bottomrule
\end{tabular}
}
\end{table*}

\begin{table}[!h]
\centering
\caption{Source of Forgetting for Different Problems}
\label{tab:forgetting}
\resizebox{0.9\columnwidth}{!}{ 
\begin{tabular}{@{}l c@{}}
\toprule
\textbf{Problem} & \textbf{Source of Forgetting} \\
\midrule
\textbf{ML} & -- \\
\textbf{CL} & sequential training tasks \\
\textbf{OML} & sequential meta-training tasks \\
\textbf{CML} & sequential meta-training tasks with task-distribution shift \\
\textbf{DFML} & task-distribution shift due to evolving generation of synthetic data\\
\bottomrule
\end{tabular}
}
\end{table}

\textbf{Extension to generative tasks.} Our method primarily consists of three components: synthetic data generation, meta-learning, and model selection. Below, we discuss how each of these components can be extended to generative tasks.
\begin{itemize}
    \item For synthetic data generation, if we have access to a collection of generative models, we can directly perform inference to obtain synthetic data without relying on model inversion.
    \item For meta-learning and model selection, their optimization objectives are inherently task-agnostic. Specifically, in the case of generative tasks, we can adapt the task-level loss (\textit{i.e.}, $\mathcal{L}_{\rm task}$) to the generative objectives. Relevant works on meta-learning with generative models can be found in \cite{sridhar2022meta, 9424414, min2021metaicl, coda2023meta}.
\end{itemize}

\subsection{More Details}
\label{sec:app_details}

\textbf{Implementation details.} We utilize four datasets: CIFAR-FS \cite{bertinetto2018meta}, MiniImageNet \cite{vinyals2016matching}, VGG-Flower \cite{flower}, and CUB-200-2011 (CUB) \cite{cub}, encompassing general natural images and specific categories like birds (CUB) and flowers (VGG-Flower). Following standard splits \cite{metadataset}, each dataset is divided into meta-training, meta-validation, and meta-testing subsets with disjoint label spaces.
In line with \cite{hu2023architecture}, we collect 100 models pre-trained on 100 $N$-way tasks sampled from the meta-training subset. Those pre-trained models are trained using an Adam optimizer with a learning rate of 0.01 for 60 epochs. The meta-learner is evaluated on 600 tasks sampled from the meta-testing subset. Both the meta-learner and the pre-trained models use a Conv4 architecture \cite{finn2017model} for a fair comparison with existing baselines. Conv4 consists of four convolutional blocks. Each block consists of 32 3 × 3 filters, a BatchNorm, a ReLU and a 2 × 2 max-pooling. We use the Adam optimizer with an inner learning rate of 0.01 and an outer learning rate of 0.001 for gradient-based meta-learning. We use the Adam optimizer with a learning rate of 0.001 for metric-based meta-learning and generator training. The coefficients for $\mathcal{R}$ and $\mathcal{L}_{\rm task}$ in \cref{eq:tdfml} are set empirically as 0.001 and 0.1, respectively. New tasks are synthesized every 200 iterations following \cite{hu2023learning}. $\alpha$ and $\beta$ of the Beta distribution are set as 0.5. The memory bank was limited to 20 tasks. 
Unless otherwise specified, we use task combination. Alternatively, we adopt a hybrid strategy denoted as “+mixup”: performing task combination before 4000 iterations as a warmup following \cite{hu2023architecture}, then switching to task mixup. The structure of the generator is provided in \cref{tab:sturcture_of_generator}. To facilitate readers' better understanding of our motivation using the synthetic images (in \cref{fig:tds} and \cref{fig:tdc_reason}), we inversely synthesize images from the pre-trained CLIP model. This is because features learned and crafted by large models tend to align more closely with human perception compared to those from smaller models \cite{ilyas2019adversarial}. All experiments are conducted on one NVIDIA GeForce RTX 3090 GPU.

\begin{algorithm}[!t]
\small
\DontPrintSemicolon
\SetKwInOut{Input}{Input}\SetKwInOut{Output}{Output}\SetKwInOut{Require}{Require}
\textbf{Input: }{Max iterations $N_{G}$ for task generation; the pre-trained model $M$ and the corresponding generator $G(\cdot;\boldsymbol{\theta}_{G})$.
}\\
\textbf{Output: }{Synthetic task $synTask$}
\SetKwFunction{GTFM}{SynTaskReconstruction}
  \SetKwProg{Fn}{Function}{:}{}
  \Fn{\GTFM{$M$, $G$}}{
        Initialize $minLoss \leftarrow +\infty ,\  synTask \leftarrow None$\\
Randomly initialize $\boldsymbol{\theta}_{G}$\\
Set the target labels $\mathbf{Y}$\\
\For{{\rm each iteration $n_g$} $\leftarrow 1$ \KwTo $N_{G}$}{
Randomly sample a batch of noise $\mathbf{Z}$ from the standard Gaussian distribution\\
$\mathbf{X}^{(n_g)} \leftarrow G(\mathbf{Z},\mathbf{Y};\boldsymbol{\theta_{G}}^{(n_g)})$ \\
Calculate the loss value $L_{G}^{(n_g)} \leftarrow \mathcal{L}_{G}(\mathbf{Z},\mathbf{Y};\boldsymbol{\theta_{G}}^{(n_g)})$\\
$\boldsymbol{\theta_{G}}^{(n_g+1)} \leftarrow \boldsymbol{\theta_{G}}^{(n_g)} - \nabla_{\boldsymbol{\theta}_{G}}\mathcal{L}_{G}^{(n_g)}$ (\cref{eq:tdfml})\\
\If{$\mathcal{L}_{G}^{(n_g)} < minLoss$}{
$minLoss \leftarrow \mathcal{L}_{G}^{(n_g)}$\\
$synTask \leftarrow \mathbf{X}^{(n_g)}$\\
}
}
Randomly split $synTask$ into the support set ${\mathbf{D}}^s$ and the query set ${\mathbf{D}}^q$\\
        \KwRet $synTask$ with ${\mathbf{D}}^s$ and ${\mathbf{D}}^q$\\
  }
\caption{Synthetic Task Reconstruction}
\label{alg:generatetaskfrommodel}
\end{algorithm}

\begin{algorithm}[!t]
\small
\DontPrintSemicolon
\SetKwInOut{Input}{Input}\SetKwInOut{Output}{Output}\SetKwInOut{Require}{Require}
\textbf{Input: }{Task memory buffer $\mathcal{B}$}\\
\textbf{Output: }{$N$-way interpolated task $\mathcal{T}$}

\SetKwFunction{TaskCombination}{TaskCombination}
\SetKwProg{Fn}{Function}{:}{}
\Fn{\TaskCombination{$\mathcal{B}$}}{
    Initialize $\mathcal{T} \leftarrow \emptyset$\\
    \For{$i \leftarrow 1$ \KwTo $|N|$}{
        \Repeat{class $c$ is not in $\mathcal{T}$}{
            $t \leftarrow$ Randomly select a task from $\mathcal{B}$\\
            $c \leftarrow$ Randomly select a class from $t$\\
            \If{$c$ is not in $\mathcal{T}$}{
                Add $c$ to $\mathcal{T}$ as the $i$-th class\\
            }
        }
    }
    \KwRet $\mathcal{T}$\\
}
\caption{Task Combination}
\label{alg:taskcombination}
\end{algorithm}

\begin{algorithm}[!t]
\small
\DontPrintSemicolon
\SetKwInOut{Input}{Input}\SetKwInOut{Output}{Output}\SetKwInOut{Require}{Require}
\textbf{Input: }{Task memory buffer $\mathcal{B}$, Random layer $l$, $\lambda \in[0,1]$ sampled from a Beta distribution $\operatorname{Beta}(\alpha, \beta)$}\\
\textbf{Output: }{$N$-way interpolated task $\mathcal{T}$}
\SetKwFunction{TaskMixup}{TaskMixup}

\SetKwProg{Fn}{Function}{:}{}
\Fn{\TaskMixup{$\mathcal{B}$}}{
    Initialize $\mathcal{T} \leftarrow \emptyset$\\
    \For{$i \leftarrow 1$ \KwTo $N$}{
        \Repeat{pair $(c_1, c_2)$ and $(c_2, c_1)$ has not been selected for $\mathcal{T}$}{
            $t_1, t_2 \leftarrow$ Randomly select two tasks from $\mathcal{B}$\\
            $c_1 \leftarrow$ Randomly select a class from $t_1$\\
            $c_2 \leftarrow$ Randomly select a class from $t_2$\\
            \If{$(c_1, c_2)$ and $(c_2, c_1)$ has not been selected for $\mathcal{T}$}{
                Obtain representations of new class $e$ at layer $l$ by performing latent mixup between $(c_1, c_2)$, $\mathbf{H}_{e}^{(l)}=\lambda \mathbf{H}_{c_1}^{(l)}+(1-\lambda) \mathbf{H}_{c_2}^{(l)}$\\
                Add $e$ to $\mathcal{T}$ as the $i$-th class\\
            }
        }
    }
    \KwRet $\mathcal{T}$\\
}
\caption{ Task Mixup }
\label{alg:taskmixup}
\end{algorithm}

\textbf{Algorithm of synthetic task reconstruction.} We provide the algorithm of synthetic task reconstruction in \cref{alg:generatetaskfrommodel}.

\textbf{Reservoir sampling (RS).} We adopt reservoir sampling (RS) to dynamically maintain a task memory buffer from a task stream, ensuring that previous tasks have a fair chance of being retained. Initially, the first $k$ tasks will be stored until the buffer fills. Upon filling, as task $i$ arrives, RS  generates a random number $j$ between 1 and $i$. If $j$ is at most $k$, task $i$ replaces the task in position $j$ of the reservoir; otherwise, it is discarded.

\begin{table*}[!t]
    \centering
    \caption{\small Detailed structure of conditional generator. We highlight the dimension change in \textcolor{blue}{blue}.}
    \scalebox{0.9}{
    \begin{tabular}{ccc}
    \toprule
    \textbf{Notion} &\multicolumn{2}{c}{\textbf{Description}}\\
    \midrule
    $img\_size$ $\times$ $img\_size$&\multicolumn{2}{c}{ resolution of generated image}\\
    $bs$ & \multicolumn{2}{c}{ batch size}\\
    $nc$ & \multicolumn{2}{c}{ number of channels of generated image}\\
    $nf$ & \multicolumn{2}{c}{ number of convolutional filters}\\
    FC($\cdot$) & \multicolumn{2}{c}{ fully connected layer;}\\
    BN($\cdot$)&\multicolumn{2}{c}{ batch normalization layer}\\
    Conv2D($input$,\ $output$,$filter\_size$,\ $stride$,\ $padding$) & \multicolumn{2}{c}{ convolutional layer}\\
    \toprule
    \multirow{2}{*}{\textbf{Structure}} & \multicolumn{2}{c}{\textbf{Dimension}}\\
    & \textbf{Before} & \textbf{After}\\
     \midrule
         $\boldsymbol{z} \in \mathbb{R}_{d_{\boldsymbol{z}}} \sim \mathcal{N}(\boldsymbol{0},\boldsymbol{1})$&--- & $\left[\ bs, d_{\boldsymbol{z}}\ \right]$  \\
         $\boldsymbol{y} \in \mathbb{R}_{d_{\boldsymbol{y}}}$&--- & $\left[\ bs, d_{\boldsymbol{y}}\ \right]$  \\
        FC$_1$($\boldsymbol{z}$) & $\left[\ bs, \textcolor{blue}{d_{\boldsymbol{z}}}\ \right] $ & $ \left[\ bs, \textcolor{blue}{nf \times (img\_size//4) \times (img\_size//4)}\ \right]$ \\
        FC$_2$($\boldsymbol{y}$) & $\left[\ bs, \textcolor{blue}{d_{\boldsymbol{y}}}\ \right] $ & $ \left[\ bs, \textcolor{blue}{nf \times (img\_size//4) \times (img\_size//4)}\ \right]$\\
        Concatenate(FC$_1$($\boldsymbol{z}$),FC$_2$($\boldsymbol{y}$))&---&$ \left[\ bs, \textcolor{blue}{2 \times nf \times (img\_size//4) \times (img\_size//4)}\ \right]$\\
        Reshape & $\left[\ bs, \textcolor{blue}{2 \times nf \times (img\_size//4) \times (img\_size//4)}\ \right] $ & $ \left[\ bs, \textcolor{blue}{2 \times nf}, \textcolor{blue}{(img\_size//4)}, \textcolor{blue}{(img\_size//4)}\ \right]$\\
        \midrule
         BN & $ \left[\ bs, 2 \times nf, (img\_size//4), (img\_size//4)\ \right]$&$ \left[\ bs, 2 \times nf, (img\_size//4), (img\_size//4)\ \right]$\\
        Upsampling &$ \left[\ bs, 2 \times nf, (img\_size//\textcolor{blue}{4}), (img\_size//\textcolor{blue}{4}))\ \right]$ & $ \left[\ bs, 2 \times nf, (img\_size//\textcolor{blue}{2}), (img\_size//\textcolor{blue}{2}))\ \right]$\\
        \midrule
        Conv2D($2 \times nf,\ 2 \times nf,\ 3,\ 1,\ 1$) & $ \left[\ bs, 2 \times nf, (img\_size//2), (img\_size//2))\ \right]$ & $ \left[\ bs, 2 \times nf, (img\_size//2), (img\_size//2))\ \right]$\\
        BN, LeakyReLU & $ \left[\ bs, 2 \times nf, (img\_size//2), (img\_size//2))\ \right]$&$ \left[\ bs, 2 \times nf, (img\_size//2), (img\_size//2))\ \right]$ \\
        Upsampling& $ \left[\ bs, 2 \times nf, (img\_size//\textcolor{blue}{2}), (img\_size//\textcolor{blue}{2}))\ \right]$ & $ \left[\ bs, 2 \times nf, img\_size, img\_size\ \right]$\\
        \midrule
        Conv2D($2 \times nf,\ nf,\ 3,\ 1,\ 1$) & $ \left[\ bs, \textcolor{blue}{2 \times nf}, img\_size, img\_size\ \right]$ & $ \left[\ bs, \textcolor{blue}{nf}, img\_size, img\_size\ \right]$\\
        BN, LeakyReLU& $ \left[\ bs, nf, img\_size, img\_size\ \right]$ & $ \left[\ bs, nf, img\_size, img\_size\ \right]$\\
        Conv2D($nf,\ nc,\ 3,\ 1,\ 1$) & $ \left[\ bs, \textcolor{blue}{nf}, img\_size, img\_size\ \right]$ & $ \left[\ bs, \textcolor{blue}{nc}, img\_size, img\_size\ \right]$\\
        Sigmoid & $ \left[\ bs, nc, img\_size, img\_size\ \right]$& $ \left[\ bs, nc, img\_size, img\_size\ \right]$\\
        \bottomrule
    \end{tabular}
    }
    \label{tab:sturcture_of_generator}
\end{table*}

\textbf{Algorithm of task memory interpolation} We provide the algorithm of two implementations of task memory interpolation in \cref{alg:taskcombination} and \cref{alg:taskmixup}, respectively.

\textbf{Details of the architecture of generator.} \cref{tab:sturcture_of_generator} lists the structure of the conditional generator. The generator takes the standard Gaussian noise and the one-hot label embedding as inputs and outputs the generated data.  Here, $d_{\boldsymbol{z}}$ is the dimension of Gaussian noise data $\boldsymbol{z}$, which is set as 256 in practice. 
We set $img\_size$ as 32. We set the number of channels $nc$ as 3 for color image recovery and the number of convolutional filters $nf$ as 64.

\subsection{More Experiments}
\label{app:ex}

\textbf{Complexity analysis of each component.} We conduct a comprehensive analysis of each component of our framework in terms of time and memory. Specifically, we factorize each element into several factors for a better comparison:
\begin{itemize}
    \item $FF_G$ represents the forward FLOPs performed during the generation phase for each task, quantifying the computational cost of processing data in the forward pass within the generator and pre-trained models. When we use the Conv4 as the architecture of pre-trained models, the value of $FF_G$ is about $117,709,152$.
    \item $FB_G \approx 3 \times FF_G$ indicates that the backward FLOPs during the generation phase are approximately three times the forward FLOPs. This is used to estimate the computational effort required for the backward pass, which involves gradient computation and backpropagation.
    \item $FF_M $ signifies the forward FLOPs during the meta-training phase per task, tracking the FLOPs involved in one forward pass within the meta-learner. When we use the Conv4 as the architecture of the meta-learner, the value of $FF_G$ is about $4,199,552$.
    \item $FB_M  \approx 3 \times FF_M $ suggests that the backward FLOPs during the meta-training phase, which are also about three times the forward FLOPs.
    \item $I_G$ refers to the number of iterations required to update the generator.
     $I_M$ denotes the number of iterations for updating the meta-learner.
     $I_W$ is the number of iterations to update the weight vector.
     $I_Z$ is the number of queries performed in the zero-order gradient estimation, which is specifically used in black-box settings for approximating gradients.
    \item  $N$ stands for the number of synthetic images per task.
\end{itemize}

\cref{tab:complexity} demonstrates that our method only introduces a minor increase in time—merely an additional 0.05 hours—compared to PURER \cite{hu2023architecture}, while offering a significant enhancement in robustness. This efficiency is achieved through the rapid computation of rewards and policy gradients, which bypasses the need for complex gradient calculations and backpropagation. In contrast, BiDf-MKD \cite{hu2023learning} requires more time due to the necessity of performing multiple queries to estimate gradients. Furthermore, the memory overhead of our method also remains comparable to other approaches.

\begin{table*}[!t]
  \centering
  \small
   \caption{Complexity analysis of each component in terms of time and memory.} 
  \scalebox{0.65}{\begin{tabular}{lccccc}
    \toprule
     Method &Synthetic Task Reconstruction (FLOPs)&Meta-Learning with Task Memory Interpolation (FLOPs)&Automatic Model Selection (FLOPs)&Time (h)&Memory\\
     \midrule
     PURER & $I_G \times N \times (FF_G+FB_G)$ &$I_M \times N \times  (FF_M+FB_M)$&NA&2.65&$O(N)$\\
     BiDf-MKD &$I_G \times N \times  (FF_G+FB_G) \times I_Z$&$I_M \times N \times  (FF_M+FB_M)$&NA&5.71&$O(N)$ \\
     TDFML &$I_G \times N \times  (FF_G+FB_G)$&$\begin{cases}
         I_M \times N \times  (FF_M+FB_M)\ \text{for task combination}\\k \times I_M \times N \times  (FF_M+FB_M), k \leq 2 \ \text{for task mixup}
     \end{cases}$  &
     $I_P \times N \times FF_M$&2.70&$\begin{cases}
         O(N)\ \text{for task combination}\\O(k N), k \leq 2\ \text{for task mixup}
     \end{cases}$ \\
     \bottomrule
  \end{tabular}}
  \label{tab:complexity}
  \vspace{-0.2cm}
\end{table*}

\begin{figure*}[!t]
	\centering
	\includegraphics[width=0.86\linewidth]{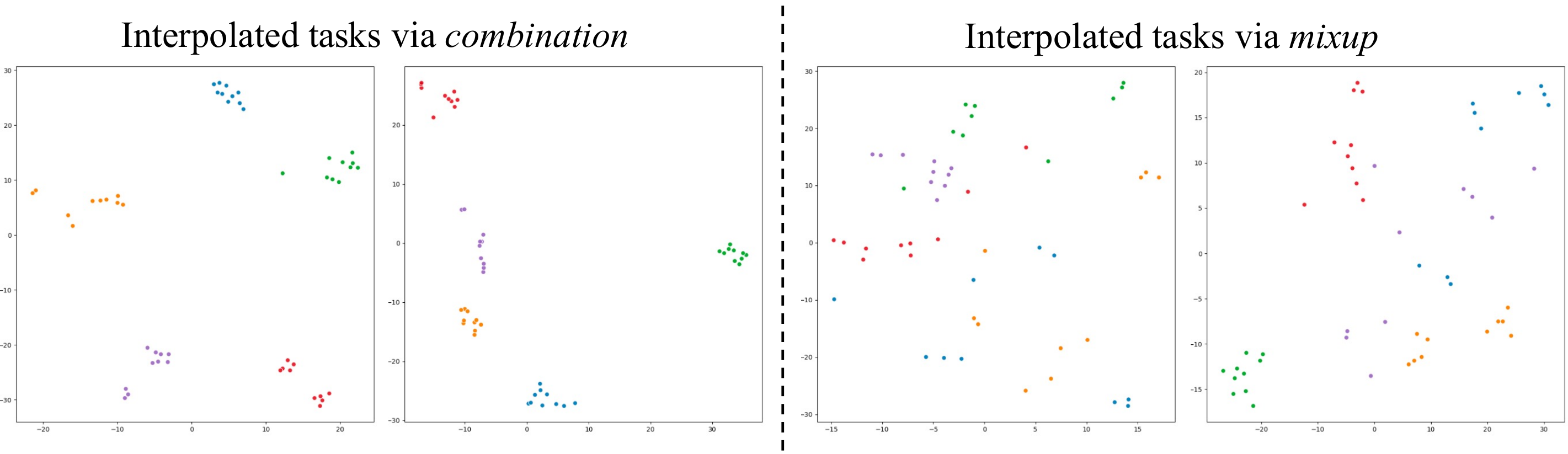}
	\captionsetup{font={small}}
	\caption{T-SNE visualization of interpolated tasks. Different colors represent different classes.}
	\label{fig:tsne}
\end{figure*}

\textbf{Effect of different task interpolation methods.}\ 
The effectiveness of different task interpolation methods is evaluated on CIFAR-FS and presented in \cref{tab:interpolation}. 
As shown in \cref{tab:interpolation}, solely using \textit{mixup} is suboptimal because tasks interpolated by \textit{mixup} tend to be more ``difficult'' compared with \textit{combination}, making them less effective for meta-training purposes. 
Inspired by \cite{hu2023architecture} and curriculum learning \cite{zhang2021curriculum}, we adopt a hybrid strategy denoted as “combination+mixup” in \cref{tab:interpolation}: performing task combination before 4000 iterations as a warmup \cite{hu2023architecture}, then switching to task mixup.
In this way, the meta-learner can learn easy tasks interpolated by combination in the early stage and then learn harder tasks for better generalization in the later stage, similarly to curriculum learning \cite{zhang2021curriculum}.
T-SNE visualizations of  tasks in \cref{fig:tsne} corroborate our findings and offer a deeper understanding of how different interpolation strategies influence task difficulty and diversity.
\begin{table}[!h]
  \centering
  \vspace{-0.25cm}
  \small
   \caption{Variants of task interpolation.} 
  \scalebox{0.7}{\begin{tabular}{ccccc}
    \toprule
     \multirow{2}{*}{\shortstack{\textbf{Interpolation}\\ \textbf{Method}}} & \multicolumn{2}{c}{\textbf{5-way 1-shot}}&\multicolumn{2}{c}{\textbf{5-way 5-shot}}\\
    \cmidrule(r){2-3}
    \cmidrule(r){4-5}
       &BEST &LAST &BEST &LAST \\
    \midrule
     combination & 40.80 $\pm$ 0.78 &  40.28 $\pm$ 0.79&  57.11 $\pm$ 0.78&  55.69 $\pm$ 0.76\\
     mixup & 39.67 $\pm$ 0.76 &  39.04 $\pm$ 0.77&  54.02 $\pm$ 0.78&  53.19 $\pm$ 0.75\\
     combination+mixup & \textbf{41.91 $\pm$ 0.72} &  \textbf{41.78 $\pm$ 0.69} & \textbf{58.25 $\pm$ 0.72}& \textbf{56.12 $\pm$ 0.74}\\
     \bottomrule
  \end{tabular}}
  \label{tab:interpolation}
  \vspace{-0.2cm}
\end{table}

\textbf{Effect of the generator.} \cref{tab:generator} illustrates the effectiveness of generator-based methods compared to pixel-based methods on CIFAR-FS. Pixel-based methods \cite{yin2020dreaming} optimize each pixel individually, leading to a significant increase in the number of learnable parameters as the number of images increases. Conversely, generator-based methods \cite{fang2022up, chen2019data, zhu2021data} train a generator (or include the latent vector) to produce images, thus offering a fixed (or slowly increasing) number of learnable parameters as the number of images increases. Besides, the architecture of a generator (even randomly initialized) can implicitly incorporate the Deep Image Prior \cite{ulyanov2018deep,cazenavette2023generalizing} as regularization.
\begin{table}[h]
  \centering
  \vspace{-0.25cm}
  \small
   \caption{Effect of the generator.} 
  \scalebox{0.9}{\begin{tabular}{ccc}
    \toprule
     Type &{\textbf{5-way 1-shot}}&{\textbf{5-way 5-shot}}\\
    \midrule
     pixel-based & 39.78 $\pm$ 0.72 &   56.43 $\pm$ 0.79\\
     generator-based & 40.80 $\pm$ 0.78 &   57.11 $\pm$ 0.78\\
     \bottomrule
  \end{tabular}}
  \label{tab:generator}
  \vspace{-0.2cm}
\end{table}

\textbf{Effect of the budgets of memory buffer.}\ \cref{tab:buffer} shows the effect of the memory buffer size on CIFAR-FS. 
With the incorporation of task memory interpolation, our framework can achieve a diverse task distribution even when constrained by limited memory budgets. This suggests that the memory buffer's effectiveness is not solely dependent on its size but also on how the memory is utilized.
The ability to maintain performance with limited memory buffers highlights the real-world applicability of our framework.
\begin{table}[!h]
  \centering
  \vspace{-0.2cm}
  \small
   \caption{Effect of the budgets of memory buffer.} 
  \scalebox{0.73}{\begin{tabular}{ccccc}
    \toprule
     \multirow{2}{*}{\shortstack{\textbf{Memory Buffer}\\ \textbf{Size (task)}}} & \multicolumn{2}{c}{\textbf{5-way 1-shot}}&\multicolumn{2}{c}{\textbf{5-way 5-shot}}\\
    \cmidrule(r){2-3}
    \cmidrule(r){4-5}
       &BEST &LAST &BEST &LAST \\
    \midrule
     1 & 34.06 $\pm$ 0.66 &  26.08 $\pm$ 0.72&  48.13 $\pm$ 0.76&  40.12 $\pm$ 0.73\\
     10 & 39.26 $\pm$ 0.73 & 37.65  $\pm$0.75 & 53.99 $\pm$ 0.76&  50.32 $\pm$ 0.76\\
     20 & \textbf{40.80 $\pm$ 0.78} & \textbf{  40.28 $\pm$ 0.79}& \textbf{ 57.11 $\pm$ 0.78} &   \textbf{55.69 $\pm$ 0.76}\\
     \bottomrule
  \end{tabular}}
  \label{tab:buffer}
  \vspace{-0.2cm}
\end{table}

\begin{table*}[!t]
    \centering
    \small
   \caption{ DFML with untrustworthy models across various pollution rates (CIFAR-FS, 5-way 1-shot).} 
  \scalebox{0.8}{\begin{tabular}{ccccccc}
    \toprule
\multirow{2}{*}{\textbf{Type}}&\multirow{2}{*}{\textbf{Method}}&\multicolumn{5}{c}{\textbf{Pollution Rate}}\\
\cmidrule{3-7}
&&10\% &20\%&40\%&60\%&80\%  \\
\midrule
\multirow{3}{*}{{Deceptive Label}}&PURER&36.73 $\pm$ 0.70&35.42 $\pm$ 0.70&34.37 $\pm$ 0.67&33.02 $\pm$ 0.71&31.87 $\pm$ 0.70\\
&TDFML (w/o AMS)&37.21 $\pm$ 0.70&35.92 $\pm$ 0.73&34.97 $\pm$ 0.74&33.86 $\pm$ 0.68&32.51 $\pm$ 0.72\\
&TDFML&38.41 $\pm$ 0.72$_{\textcolor{blue}{+1.68\%}}$&37.23 $\pm$ 0.72$_{\textcolor{blue}{+1.81\%}}$&36.52 $\pm$ 0.72$_{\textcolor{blue}{+2.15\%}}$&35.81 $\pm$ 0.69$_{\textcolor{blue}{+2.79\%}}$&34.90 $\pm$ 0.65$_{\textcolor{blue}{3.03\%}}$\\
\midrule
\multirow{3}{*}{{Deceptive Accuracy}}&PURER&37.21 $\pm$ 0.69&36.59 $\pm$ 0.72&35.18 $\pm$ 0.70&34.24 $\pm$ 0.74&32.75 $\pm$ 0.69\\
&TDFML (w/o AMS)&37.79 $\pm$ 0.70&36.92 $\pm$ 0.72&35.88 $\pm$ 0.74&34.92 $\pm$ 0.71&33.34 $\pm$ 0.68\\
&TDFML&38.97 $\pm$ 0.71$_{\textcolor{blue}{+1.76\%}}$&38.43 $\pm$ 0.72$_{\textcolor{blue}{+1.84\%}}$&37.49 $\pm$ 0.76$_{\textcolor{blue}{+2.31\%}}$&36.91 $\pm$ 0.70$_{\textcolor{blue}{+2.67\%}}$&35.76 $\pm$ 0.68$_{\textcolor{blue}{+3.01\%}}$\\
\bottomrule
    \end{tabular}}
    \label{tab:main_attack2}
\end{table*}

\textbf{DFML with heterogeneous pre-trained models.}\ In order to illustrate the model-agnostic nature of our proposed framework, we conducte experiments using different neural network architectures on CIFAR-FS. The results, as shown in \cref{tab:architecture}, indicate how our framework performs across varied architectures in both 5-way 1-shot and 5-way 5-shot learning scenarios. With a homogeneous architecture (100\% Conv4), the framework achieves 40.80\% in 5-way 1-shot learning and 57.11\% in 5-way 5-shot learning.
In the heterogeneous setting, comprising 33\% Conv4, 33\% ResNet10, and 33\% ResNet18, the framework achieves 41.94\%  in 5-way 1-shot learning and 57.02\% in 5-way 5-shot learning.
These findings demonstrate that our framework maintains consistent performance regardless of the underlying neural network architectures, indicating its effective model-agnostic properties.
\begin{table}[!h]
  \centering
  \small
   \caption{TDFML with heterogeneous pre-trained models.} 
  \scalebox{0.8}{\begin{tabular}{ccccc}
    \toprule
     \multirow{2}{*}{\textbf{Architecture}} & \multicolumn{2}{c}{\textbf{5-way 1-shot}}&\multicolumn{2}{c}{\textbf{5-way 5-shot}}\\
    \cmidrule(r){2-3}
    \cmidrule(r){4-5}
       &BEST &LAST &BEST &LAST \\
    \midrule
   Homogeneous& 40.80 $\pm$ 0.78 & 40.28 $\pm$ 0.79 & 57.11 $\pm$ 0.78& 55.69 $\pm$ 0.76\\
    \shortstack{Heterogeneous}&41.94 $\pm$ 0.73& 41.56 $\pm$ 0.74& 57.02 $\pm$ 0.76& 56.04 $\pm$ 0.75\\
     \bottomrule
  \end{tabular}}
  \label{tab:architecture}
\end{table}

\textbf{Effect of the number of pre-trained models.}\ The effect of varying the number of pre-trained models on our method (TDFML), as well as on RANDOM and AVERAGE baselines  on CIFAR-FS, is detailed in \cref{tab:modelnumber}. We observed that within a certain range, an increase in the number of pre-trained models positively impacts TDFML's performance. This suggests that TDFML effectively leverages additional models to enhance its learning capability.
However, adding more models damages AVERAGE. The reason why AVERAGE even performs worse than RANDOM is that each
pre-trained model trains on different tasks, thus lacking precise correspondence  in parameter space. 
The poor performance of AVERAGE also highlights the challenges of utilizing a large volume of models for practical applications.
\begin{table}[!h]
  \centering
  \vspace{-0.2cm}
  \small
   \caption{Effect of the number of pre-trained models.} 
  \scalebox{0.7}{\begin{tabular}{cccccc}
    \toprule
     \multirow{2}{*}{\shortstack{\textbf{Method}}}&\multirow{2}{*}{\shortstack{\textbf{Number}}} & \multicolumn{2}{c}{\textbf{5-way 1-shot}}&\multicolumn{2}{c}{\textbf{5-way 5-shot}}\\
    \cmidrule(r){3-4}
    \cmidrule(r){5-6}
       &&BEST &LAST &BEST &LAST \\
       \midrule
       RANDOM &0& \textbf{28.59 $\pm$ 0.56} & \textbf{28.59 $\pm$ 0.56} & 34.77 $\pm$ 0.62& 34.77 $\pm$ 0.62 \\
           \midrule
       \multirow{3}{*}{AVERAGE}&10 & 27.99 $\pm$ 0.59 & 27.99 $\pm$ 0.59 & \textbf{36.92 $\pm$ 0.67}& \textbf{36.92 $\pm$ 0.67} \\
    &50 &24.05 $\pm$ 0.51& 24.05 $\pm$ 0.51 & 28.16 $\pm$ 0.50& 28.16 $\pm$ 0.50 \\
    &100 &23.96 $\pm$ 0.53 & 23.96 $\pm$ 0.53 & 27.04 $\pm$ 0.51& 27.04 $\pm$ 0.51 \\
    \midrule
    \multirow{3}{*}{TDFML}&10 & 38.32 $\pm$ 0.75 & 37.14$\pm$ 0.76& 49.83 $\pm$ 0.74&47.22 $\pm$ 0.78\\
    &50 &39.15 $\pm$ 0.75&38.98 $\pm$ 0.76& 52.41 $\pm$ 0.76&51.27 $\pm$ 0.76\\
    &100 &\textbf{40.80 $\pm$ 0.78} &\textbf{40.28 $\pm$ 0.79}& \textbf{57.11 $\pm$ 0.78}&\textbf{55.69 $\pm$ 0.76}\\
     \bottomrule
  \end{tabular}}
  \label{tab:modelnumber}
\end{table}

\begin{figure*}[!h]
    \centering
    \includegraphics[width=0.6\linewidth]{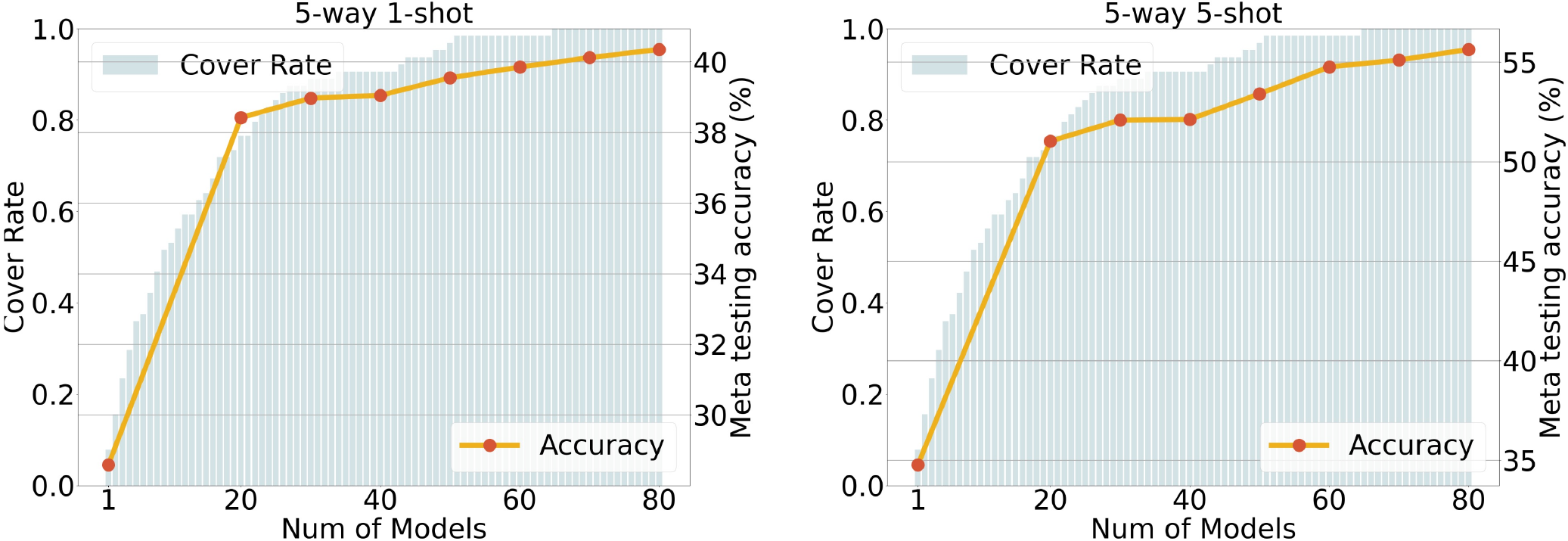}
    \caption{Effect of the model pool size on CIFAR-FS dataset. We observe that the growth trend of the cover rate closely matches that of the meta-learning performance as the number of models increases.}
    \label{fig:num_model}
\end{figure*}

To further investigate how the model pool size influences meta-learning performance, we introduce a metric called the \textit{cover rate}, which measures the ratio of the number of unique classes covered by the current model pool to the maximum number of classes.
As shown in \cref{fig:num_model}, the growth trend of the cover rate closely matches that of the meta-learning performance as the number of models increases. Specifically, increasing the number of models in the pool leads to a higher cover rate and better meta-learning performance. This finding suggests that including a greater variety of classes enhances the meta-learner’s generalization ability. Moreover, even after the cover rate reaches 100\%, we observe an improvement in performance as the number of models continues to grow. This is because the additional models offer richer supervision of the semantic relationships among different classes.

\textbf{Using pre-trained models from real repositories.} To ensure our experiments are grounded in real-world scenarios, we select a variety of pre-trained models from public repositories, covering multiple datasets and model architectures. These include: 
\begin{itemize}
    \item ResNet34 pre-trained on CIFAR100 dataset (https://huggingface.co/edadaltocg/resnet34\_cifar100)
    \item  ViT-L pre-trained on Tiny ImageNet dataset (https://github.com/ehuynh1106/TinyImageNet-Transformers)
    \item ResNet18 pre-trained on CUB dataset (https://github.com/PRIS-CV/Mutual-Channel-Loss/tree/master)
    \item VGG19 pre-trained on VGG Flower dataset (https://github.com/TrasperJ/102-flowers-classfication-with-PyTorch/tree/master)
\end{itemize}

The evaluation is conducted on 600 tasks evenly sampled  from additional datasets (Aircraft, Quick Draw, Textures and MSCOCO).
The results in \cref{tab:real} indicate that our method remains effective and outperforms other DFML baselines, when using pre-trained models from real-world repositories.

\begin{table}[!h]
\vspace{-0.2cm}
\centering
   \caption{Using pre-trained models from real repositories.} 
  \scalebox{0.86}{\begin{tabular}{lcccc}
    \toprule
     \multirow{2}{*}{\textbf{Method}} & \multicolumn{2}{c}{\textbf{5-way 1-shot}}&\multicolumn{2}{c}{\textbf{5-way 5-shot}}\\
    \cmidrule(r){2-3}
    \cmidrule(r){4-5}
       &BEST &LAST &BEST &LAST \\
    \midrule
     DFL2L& 29.53 $\pm$ 0.56  &27.63 $\pm$ 0.69& 32.52 $\pm$ 0.62& 30.62 $\pm$ 0.69\\
      PURER& 35.78 $\pm$ 0.56 & 26.78 $\pm$ 0.61& 47.12 $\pm$ 0.67&39.71 $\pm$ 0.71 \\
     TDFML (ours)& \textbf{40.82 $\pm$ 0.61}&\textbf{40.12 $\pm$ 0.62}   & \textbf{52.55 $\pm$ 0.59}&\textbf{51.62 $\pm$ 0.59} \\
     \bottomrule
  \end{tabular}}
  \label{tab:real}
\end{table}

\textbf{Extension to black-box setting.} Our method can be easily extended to the black-box setting. Assuming the loss function of synthetic data generation as $\mathcal{L}_G$, with generator parameters $\boldsymbol{\theta}_ G$ and generated data $\mathbf{X}$, we follow BiDf-MKD \cite{hu2023learning} to split the gradient using the chain rule: $\nabla_ {\boldsymbol{\theta}_ G} \mathcal{L}_G=\frac{\partial \mathcal{L}_G}{\partial \boldsymbol{\theta}_ G}=\frac{\partial \mathcal{L}_G}{\partial \mathbf{X}} \times \frac{\partial \mathbf{X}}{\partial \boldsymbol{\theta}_ G}$.
While PyTorch's automatic differentiation handles the second term, the first term $\frac{\partial \mathcal{L}_G}{\partial \mathbf{X}}$ is intractable due to lack of access to the black-box model parameters. Like BiDf-MKD, we use zero-order optimization to estimate this term.
In \cref{tab:black}, we conduct experiments comparing our method with BiDf-MKD and other baselines in the black-box settings on CIFAR-FS, demonstrating its versatility and superiority.

\begin{table}[!h]
\vspace{-0.2cm}
\centering
   \caption{Extension to the black-box setting.} 
  \scalebox{0.8}{\begin{tabular}{lcccc}
    \toprule
     \multirow{2}{*}{\textbf{Method}} & \multicolumn{2}{c}{\textbf{5-way 1-shot}}&\multicolumn{2}{c}{\textbf{5-way 5-shot}}\\
    \cmidrule(r){2-3}
    \cmidrule(r){4-5}
       &BEST &LAST &BEST &LAST \\
    \midrule
     PURER (black-box)& 33.21 $\pm$ 0.74 &  24.52 $\pm$ 0.76 & 45.56 $\pm$ 0.64 &   34.51 $\pm$ 0.71\\
      BiDf-MKD (black-box)& 35.48 $\pm$ 0.67 &  26.42 $\pm$ 0.78 & 47.58 $\pm$ 0.74 &   27.31 $\pm$ 0.71 \\
     TDFML (black-box)& \textbf{38.23 $\pm$ 0.61} &  \textbf{37.83 $\pm$ 0.67} & \textbf{51.36 $\pm$ 0.68} &   \textbf{50.42 $\pm$ 0.65}\\
     \bottomrule
  \end{tabular}}
  \label{tab:black}
\end{table}

\textbf{Experiments with untrustworthy models on 5-way 1-shot.} \cref{tab:main_attack2} shows that our framework consistently improves performance across various pollution rates and untrustworthy models, which aligns with our findings in \cref{tab:main_attack}. Similarly, within a certain range, as the pollution rate increases, TDFML with automatic model selection (AMS) demonstrates more improvements, suggesting enhanced robustness of DFML.

\textbf{T-SNE visualization of interpolated tasks.}\ The t-SNE visualization in \cref{fig:tsne} provides insights into the effectiveness of task interpolation methods. The left panel illustrates interpolated tasks via combination, with each color representing a different task and points indicating individual examples within those tasks. This combination method reveals distinct clusters, signifying that tasks are well-separated in the feature space and thus easier to classify.
In contrast, the right panel depicts interpolated tasks via mixup. Here, we observe a more blended distribution of points, indicating that mixup creates tasks with overlapping features. This overlap can be advantageous for the meta-learner, as it increases difficulty levels and promotes better generalization \cite{zhang2021curriculum}.
Employing a hybrid approach—using combination for warmup followed by mixup—yields superior performance. The meta-learner can learn easy tasks interpolated by combination in the early stage and then learn more harder tasks for better generalization in the later stage, similarly to curriculum learning \cite{zhang2021curriculum}. Empirical results detailed in \cref{tab:interpolation} support this, demonstrating the enhanced performance of the hybrid approach.

\begin{figure}[!h]
    \centering
    \includegraphics[width=0.7\linewidth]{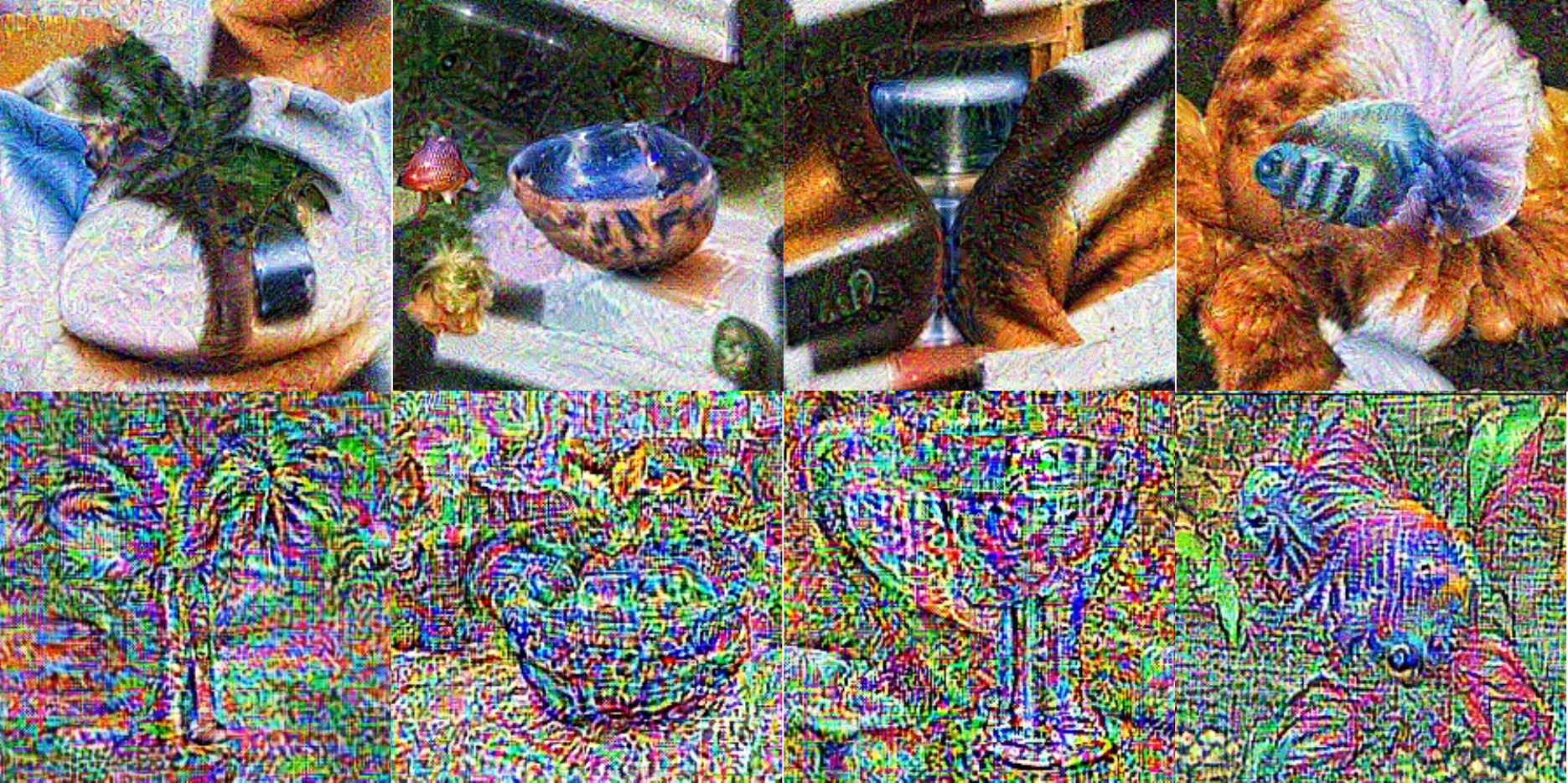}
    \caption{ Effect of the naturalness prior $\mathcal{R}$. (Top) Synthetic data with $\mathcal{R}$. (Bottom) Synthetic data without $\mathcal{R}$.}
    \label{fig:bn}
\end{figure}

\textbf{Effect of the naturalness prior $\mathcal{R}$.} \cref{fig:bn} demonstrates the effectiveness of the naturalness prior, $\mathcal{R}$ in \cref{eq:bn}, in enhancing the realism of images. This improves image realism by enriching natural colors and reducing noise, contributing to a more visually authentic representation.

\textbf{Effect of differences between synthetic and real data.} To evaluate the effect of differences between synthetic and real data, we compare the meta-learner’s performance when trained on synthetic data versus real data. As shown in \cref{tab:real_data}, meta-learners trained on real data outperform those trained on synthetic ones. This performance gap is expected, as real data better reflect the target distribution and typically contain richer structural and semantic information.
Nevertheless, our method is designed for the practical scenario in which only pre-trained models are accessible and real data is unavailable. In such cases, synthetic data serve as a feasible approximation. The performance achieved using real data can therefore be regarded as an ideal upper bound for DFML.

\begin{table}[!h]
  \centering
  \caption{Effect of differences between synthetic and real data.} 
  \scalebox{0.7}{
  \begin{tabular}{lccccc}
    \toprule
    \multirow{2}{*}{\textbf{Method}} & \multirow{2}{*}{\textbf{Data}} & \multicolumn{2}{c}{\textbf{CIFAR-FS} \cite{bertinetto2018meta}} & \multicolumn{2}{c}{\textbf{MiniImageNet} \cite{vinyals2016matching}}\\
    \cmidrule(r){3-4}
    \cmidrule(r){5-6}
    & & 5-way 1-shot & 5-way 5-shot & 5-way 1-shot & 5-way 5-shot \\
    \midrule
    \multirow{2}{*}{{\textsc{TDFML}-ANIL}}&Real&49.55 ± 0.78&64.42 ± 0.79&42.60 ± 0.79& 59.24 ± 0.98 \\
    &Synthetic&40.39 ± 0.79&55.31 ± 0.75&32.58 ± 0.68&43.63 ± 0.72\\
    \midrule
    \multirow{2}{*}{{\textsc{TDFML}-ProtoNet}}&Real&50.86  ± 0.78&67.62 ± 0.77&44.22 ± 0.78&62.23 ± 0.66\\
    &Synthetic&41.91 ± 0.72&58.25 ± 0.72&33.82 ± 0.68&43.97 ± 0.67\\
    \bottomrule
  \end{tabular}}
  \label{tab:real_data}
\end{table}

\textbf{Effect of task combination and task mixup.} To demonstrate the effectiveness of our proposed task combination and task mixup techniques, we compare them against the baseline techniques mentioned above. Results in \cref{tab:augmentation} demonstrate that our proposed techniques consistently outperform these baselines. Notably, the higher LAST performance suggests that our methods effectively mitigate the meta-learner’s tendency to forget previously acquired meta-knowledge, while other techniques fail to maintain high performance over time. We attribute this improvement to the fact that our methods perform \textit{inter-task augmentation}—by combining or mixing features across different tasks—which generates entirely new tasks across different task and thus significantly enrich the task diversity. This advantage is particularly important when the number of synthetic tasks is small, as intra-task augmentation alone is limited to manipulating data within each task and cannot provide sufficient diversity.

\begin{table*}[!h]
  \centering
   \caption{Comparison with more augmentation techniques.} 
   \scalebox{0.85}{
  \begin{tabular}{lcccccccc}
    \toprule
   \multirow{3}{*}{\textbf{Technique}} & \multicolumn{4}{c}{\textbf{CIFAR-FS} \cite{bertinetto2018meta}}&\multicolumn{4}{c}{\textbf{MiniImageNet} \cite{vinyals2016matching}}\\
    \cmidrule(r){2-5}
    \cmidrule(r){6-9}
    &\multicolumn{2}{c}{\textbf{5-way 1-shot}}&\multicolumn{2}{c}{\textbf{5-way 5-shot}}&\multicolumn{2}{c}{\textbf{5-way 1-shot}}&\multicolumn{2}{c}{\textbf{5-way 5-shot}}\\
    \cmidrule(r){2-3}
    \cmidrule(r){4-5}
    \cmidrule(r){6-7}
    \cmidrule(r){8-9}
     &\textbf{BEST} &\textbf{LAST} &\textbf{BEST} &\textbf{LAST}&\textbf{BEST} &\textbf{LAST}&\textbf{BEST} &\textbf{LAST} \\
    \midrule
    w/o Augmentation& 36.02 $\pm$ 0.70 &28.33 $\pm$ 0.68 &51.23 $\pm$ 0.72 & 41.54 $\pm$ 0.70 &29,84 $\pm$ 0.66 & 24.31 $\pm$ 0.68 & 40.12 $\pm$ 0.70 & 31.46 $\pm$ 0.71\\
    Standard Augmentation & 37.02 $\pm$ 0.78 & 28.19 $\pm$ 0.78 & 52.43 $\pm$ 0.76 & 41.40 $\pm$ 0.71 & 30.84 $\pm$ 0.74 & 23.14 $\pm$ 0.74 & 41.32 $\pm$ 0.75 & 32.37 $\pm$ 0.75  \\
    Meta-MaxUp ~\cite{ni2021data} & 38.02 $\pm$ 0.75 & 28.13 $\pm$ 0.74 & 53.43 $\pm$ 0.72 & 41.74 $\pm$ 0.70 & 31.84 $\pm$ 0.73 & 24.80 $\pm$ 0.77 & 42.32 $\pm$ 0.75 & 32.07 $\pm$ 0.76  \\
    Meta-Augmentation~\cite{rajendran2020meta} & 37.32 $\pm$ 0.75 & 28.88 $\pm$ 0.68 & 52.63 $\pm$ 0.74 & 40.41 $\pm$ 0.76 & 31.14 $\pm$ 0.75 & 24.83 $\pm$ 0.73 & 41.52 $\pm$ 0.70 & 31.29 $\pm$ 0.75  \\
    \midrule
    Task Combination (ours) &{40.80 $\pm$ 0.78}&{40.28 $\pm$ 0.79}&  {57.11 $\pm$ 0.78}&{55.69 $\pm$ 0.76}&  {32.61 $\pm$ 0.64}& {31.97 $\pm$ 0.61}&  42.93 $\pm$ 0.65& 41.28 $\pm$ 0.64 \\
Task MixUp  (ours) &\textbf{41.91 $\pm$ 0.72}&\textbf{41.78 $\pm$ 0.69}&  \textbf{58.25 $\pm$ 0.72}&\textbf{56.12 $\pm$ 0.74}&  \textbf{33.82 $\pm$ 0.68}& \textbf{32.75 $\pm$ 0.64}&  \textbf{43.97 $\pm$ 0.67}& \textbf{42.88 $\pm$ 0.62} \\
     \bottomrule
  \end{tabular}}
  \label{tab:augmentation}
\end{table*}

\textbf{Extension to adaptive difficulty mechanism.} In the current implementation, our framework increases the difficulty of the synthetic tasks by maximizing $\mathcal{L}_{\text{task}}$, which is formulated in \cref{eq:tdfml}. 
However, we acknowledge that this approach does not explicitly adapt the task difficulty based on the meta-learner’s learning dynamics. As a promising future direction, we incorporate an adaptive mechanism by dynamically adjusting the weight $\lambda_{\text{task}}$ of $\mathcal{L}_{\text{task}}$ using the meta-learner’s real-time feedback (\textit{i.e.}, training loss change), which also reflects the meta-learner’s ability to handle the current tasks. Specifically, we track the moving average of the training loss change over the past $K$ steps:
\begin{equation}
\Delta \bar{L}^{(t)} = \frac{1}{K} \sum_{k=0}^{K-1} (L^{(t-k)} - L^{(t-k-1)}),
\end{equation}
and update the weight $\lambda_{\text{task}}$ as:
\begin{equation}
\lambda_{\text{task}}^{(t+1)} = \max\left(0,\ \lambda_{\text{task}}^{(t)} - \eta \cdot \text{sign}(\Delta \bar{L}^{(t)})\right),
\end{equation}
where $\eta$ is a small step size.
If the training loss consistently decreases ($\Delta \bar{L}^{(t)} < 0$), this indicates the tasks are too easy, prompting us to increase $\lambda_{\text{task}}$ and generate more challenging tasks. Conversely, if the loss stagnates or increases, we decrease $\lambda_{\text{task}}$ to provide easier tasks that better align with the meta-learner’s current capacity.

To demonstrate the effectiveness of the adaptive difficulty mechanism, we conduct experiments and present the results in \cref{tab:adaptive}. We empirically set $K=20$ and $\eta=1\times10^{-4}$. These results show that the adaptive difficulty mechanism outperforms the non-adaptive variant across several datasets. This demonstrates that adaptively adjusting task difficulty according to the meta-learner's feedback leads to better meta-learning performance.

\begin{table}[!h]
  \centering
  \caption{Effect of adaptive difficulty mechanism.} 
  \scalebox{0.7}{
  \begin{tabular}{lccccc}
    \toprule
    \multirow{2}{*}{\textbf{Method}} & \multirow{2}{*}{\textbf{Task Difficulty}} & \multicolumn{2}{c}{\textbf{CIFAR-FS} \cite{bertinetto2018meta}} & \multicolumn{2}{c}{\textbf{MiniImageNet} \cite{vinyals2016matching}}\\
    \cmidrule(r){3-4}
    \cmidrule(r){5-6}
    & & 5-way 1-shot & 5-way 5-shot & 5-way 1-shot & 5-way 5-shot \\
    \midrule
    \multirow{2}{*}{{\textsc{TDFML}-ANIL}}&w/o feedback&40.39 ± 0.79&55.31 ± 0.75&32.58 ± 0.68&43.63 ± 0.72 \\
    & w/ feedback & \textbf{41.34 ± 0.80} & \textbf{56.49 ± 0.76} & \textbf{33.67 ± 0.68} & \textbf{44.67 ± 0.74} \\
    \midrule
    \multirow{2}{*}{{\textsc{TDFML}-ProtoNet}}&w/o feedback&41.91 ± 0.72&58.25 ± 0.72&33.82 ± 0.68&43.97 ± 0.67\\
    & w/ feedback & \textbf{42.77 ± 0.74} & \textbf{59.11 ± 0.74} & \textbf{34.64 ± 0.69} & \textbf{45.12 ± 0.69} \\
    \bottomrule
  \end{tabular}}
  \label{tab:adaptive}
\end{table}

\begin{table}[!t]
  \centering
  \caption{Effect of different image regularizations.} 
  \scalebox{0.7}{
  \begin{tabular}{llcccc}
    \toprule
    \multirow{2}{*}{\textbf{Method}} & \multirow{2}{*}{\textbf{Regularization}} & \multicolumn{2}{c}{\textbf{CIFAR-FS} \cite{bertinetto2018meta}} & \multicolumn{2}{c}{\textbf{MiniImageNet} \cite{vinyals2016matching}}\\
    \cmidrule(r){3-4}
    \cmidrule(r){5-6}
    & & 5-way 1-shot & 5-way 5-shot & 5-way 1-shot & 5-way 5-shot \\
    \midrule
    \multirow{4}{*}{{\textsc{TDFML}-ProtoNet}}
    & w/o regularization & 32.56 ± 0.74 & 49.62 ± 0.70 & 27.64  ± 0.72 & 36.42 ± 0.72\\
    & w/ $\mathcal{R}_{\rm l_2}$ & 32.02 ± 0.72 & 49.12 ± 0.76 & 27.66 ± 0.70 & 36.23 ± 0.72\\
    & w/ $\mathcal{R}_{\rm TV}$ & 33.21 ± 0.74 & 50.06 ± 0.73 & 27.07 ± 0.71 & 37.08 ± 0.74\\
    &w/ $\mathcal{R}_{\rm nature}$& \textbf{41.91 ± 0.72}&\textbf{58.25 ± 0.72}&\textbf{33.82 ± 0.68}&\textbf{43.97 ± 0.67}\\
    \bottomrule
  \end{tabular}}
  \label{tab:reg}
\end{table}

\textbf{Robustness to dynamic model pool.} The automatic model selection (AMS) mechanism remains effective even when the pre-trained model pool changes over time. This is because AMS does not rely on static assumptions about the pool; instead, it dynamically adapts based on real-time feedback. Concretely, at each meta-iteration, the selection weights $\boldsymbol{W}$ are updated according to the real-time rewards observed for the selected models in the current iteration. As new models are added to the pool or existing models change, AMS continuously re-evaluates and adjusts the weights to prioritize models that provide higher rewards. Therefore, the mechanism inherently adapts to any updates in the model pool and remains effective in selecting the most beneficial models.

To empirically verify the robustness of AMS against dynamic model pool, we conduct additional experiments where the model pool is expanded threefold during meta-training, while keeping the same poisoning ratios. The results in \cref{tab:incremental} confirm that incorporating AMS significantly outperforms the baseline without AMS, demonstrating the effectiveness of AMS even when the pre-trained model pool changes over time.

\begin{table*}[!h]
    \centering
    \caption{Effect of automatic model selection (AMS) under different pollution rates in the scenario of dynamic model pool (CIFAR-FS, 5-WAY 5-SHOT). The model pool is expanded threefold during meta-training and the results are reported at the final stage.}
    \scalebox{0.85}{\begin{tabular}{ccccccc}
    \toprule
    \multirow{2}{*}{\textbf{Type}} & \multirow{2}{*}{\textbf{Method}} & \multicolumn{5}{c}{\textbf{Pollution Rate}} \\
    \cmidrule{3-7}
    &&10\% &20\%&40\%&60\%&80\% \\
    \midrule
    \multirow{2}{*}{{Deceptive Label}} 
    & TDFML (w/o AMS) & 48.57 $\pm$ 0.73 & 47.19 $\pm$ 0.76 & 46.22 $\pm$ 0.73 & 42.81 $\pm$ 0.73 & 41.15 $\pm$ 0.70 \\
    & TDFML & \textbf{50.68 $\pm$ 0.72} & \textbf{49.84 $\pm$ 0.72} & \textbf{48.94 $\pm$ 0.72} & \textbf{46.27 $\pm$ 0.69} & \textbf{44.41 $\pm$ 0.68} \\
    \midrule
    \multirow{2}{*}{{Deceptive Accuracy}} 
    & TDFML (w/o AMS) & 49.21 $\pm$ 0.72 & 47.97 $\pm$ 0.74 & 46.33 $\pm$ 0.76 & 44.14 $\pm$ 0.79 & 42.12 $\pm$ 0.67 \\
    & TDFML & \textbf{51.05 $\pm$ 0.69} & \textbf{50.42 $\pm$ 0.78} & \textbf{49.59 $\pm$ 0.73} & \textbf{47.40 $\pm$ 0.71} & \textbf{45.33 $\pm$ 0.68} \\
    \bottomrule
    \end{tabular}}
    \label{tab:incremental}
\end{table*}

\textbf{Effect of image regularization.} To demonstrate the effect of different image regularization, we compare three different image regularizations \cite{mahendran2015understanding,nguyen2015deep,yin2020dreaming,ulyanov2018deep}, including $R_{\rm nature}$, $R_{\rm TV}$, and $R_{\rm l_2}$.
\begin{itemize}
  \item \textbf{Naturalness regularization $R_{\rm nature}$ (\textit{i.e.}, \cref{eq:bn})}: $R_{\rm nature}$ constrains the mean and variance of the generated feature maps at each layer to match the batch normalization statistics of the real data.
  \begin{equation}
    \mathcal{R}_{\rm nature}(\mathbf{X})=\sum_l\left\|\mu^{(l)}(\mathbf{X})-\mu_{\mathrm{BN}}^{(l)}\right\|_2+\left\|\sigma^{(l)}(\mathbf{X})-\sigma_{\mathrm{BN}}^{(l)}\right\|_2 .
\end{equation}
Here, $\mu^{(l)}(\mathbf{X})$ and $\sigma^{(l)}(\mathbf{X})$ denote the mean and variance of feature maps calculated at the $l^{th}$ layer of the pre-trained model. $\mu_{\rm BN}^{(l)}$ and $\sigma_{\rm BN}^{(l)}$ denote the statistics initially stored in the $l^{th}$ batch normalization layer of the pre-trained model. Since  $\mu_{\rm BN}^{(l)}$ and $\sigma_{\rm BN}^{(l)}$ is calculated with real data, minimizing gaps in these statistics can align the distribution between the synthetic and real data, thus improving realism.
  
  \item \textbf{Total variance regularization $R_{\rm TV}$}: $R_{\rm TV}$ encourages spatial smoothness by penalizing abrupt changes between neighboring pixels.
  \begin{equation}
  \begin{aligned}
 &\mathcal{R}_{\mathrm{TV}}(\mathbf{X})=\sum_{i=1}^H \sum_{j=1}^W \sum_{(a, b) \in \mathcal{N}}\left\|\mathbf{X}_{i, j}-\mathbf{X}_{i+a, j+b}\right\|_2,\\ &\text{where}\quad \mathcal{N}=\{(-1,0),(0,-1),(-1,-1),(-1,+1)\}.
   \end{aligned}
\end{equation}
  
\item \textbf{L2-norm regularization $\mathcal{R}_{\rm l_2}$}: $\mathcal{R}_{\rm l_2}$ controls the magnitude of the generated data by directly penalizing large pixel values.
\begin{equation}
\mathcal{R}_{\rm l_2}(\mathbf{X}) = \|\mathbf{X}\|_2
\end{equation}

\end{itemize}

We adopt the recommended weights for $\mathcal{R}_{\rm nature}$, $\mathcal{R}_{\rm TV}$, and $\mathcal{R}_{\rm l_2}$ as $1 \times 10^{-3}$, $2.5 \times 10^{-5}$, and $3 \times 10^{-8}$, as suggested by \cite{yin2020dreaming}. The ablation study in \cref{tab:reg} demonstrates that incorporating $\mathcal{R}_{\rm nature}$ significantly improves meta-learning performance, while adding $\mathcal{R}_{\rm TV}$ and $\mathcal{R}_{\rm l_2}$ has unstable or even negative effects. This finding is also consistent with the weak strength setting for $\mathcal{R}_{\rm TV}$ and $\mathcal{R}_{\rm l_2}$ in \cite{yin2020dreaming}. $\mathcal{R}_{\rm nature}$ is particularly effective because it directly matches the batch normalization statistics, thereby reducing the distribution shift between generated and real data. In contrast, $\mathcal{R}_{\rm TV}$ and $\mathcal{R}_{\rm l_2}$ mainly encourage smoothness and smaller magnitude of generated data, which do not directly address the domain gap and can potentially over-constrain the data generation process.

\begin{table*}[!t]
  \centering
  \caption{Robustness to model pool with varying difficulty level.} 
  \scalebox{0.94}{
  \begin{tabular}{ccccccccc}
    \toprule
    \multirow{3}{*}{\textbf{Selection Criteria}}&\multicolumn{8}{c}{\textbf{Meta-training: models pre-trained on VGG-Flower, CIFAR-FS, CUB, and MiniImageNet}} \\
    \cmidrule(r){2-9}
    & 
    \multicolumn{2}{c}{\textbf{Meta-testing: VGG-Flower} \cite{flower}} & 
    \multicolumn{2}{c}{\textbf{Meta-testing: CIFAR-FS} \cite{bertinetto2018meta}} &
    \multicolumn{2}{c}{\textbf{Meta-testing: CUB}\cite{cub}} &
    \multicolumn{2}{c}{\textbf{\textbf{Meta-testing: MiniImageNet} \cite{vinyals2016matching}}} \\
    \cmidrule(r){2-3}
    \cmidrule(r){4-5}
    \cmidrule(r){6-7}
    \cmidrule(r){8-9}
    & 5-way 1-shot & 5-way 5-shot 
    & 5-way 1-shot & 5-way 5-shot 
    & 5-way 1-shot & 5-way 5-shot 
    & 5-way 1-shot & 5-way 5-shot \\
    \midrule
    Accuracy-based & 51.63 ± 0.76 & 66.51 ± 0.72 & 38.64 ± 0.74 & 54.62 ± 0.76 & 31.46 ± 0.72 & 44.62 ± 0.72 & 28.24 ± 0.74 & 38.52 ± 0.72 \\
    AMS (ours) & \textbf{58.23 ± 0.78}& \textbf{72.22 ± 0.69} & \textbf{41.91 ± 0.72}&\textbf{58.25 ± 0.72}& \textbf{37.93 ± 0.70} & \textbf{49.54 ± 0.68} &\textbf{33.82 ± 0.68}&\textbf{43.97 ± 0.67 }   \\
    \bottomrule
  \end{tabular}}
  \label{tab:difficulty}
\end{table*}

\textbf{Hyperparameter sensitivity.} For the hyperparameters, we provide detailed motivation and empirical support below.
\begin{itemize}
\item For the meta-training setups, we follow the widely adopted settings from classic meta-learning works \cite{finn2017model,snell2017prototypical}. Specifically, we use the widely adopted learning rate and iteration numbers as recommended, which ensures fairness in comparisons.
\item For the strengths $\lambda_{\mathcal{R}}$ and $\lambda_{\rm task}$ of the loss terms $\mathcal{R}$ and $\mathcal{L}_{\rm task}$ in \cref{eq:purer}, we conduct thorough ablation studies to examine the influence of different values. The results in \cref{tab:ablation_1,tab:ablation_2} show that within a reasonable range, varying $\lambda_{\mathcal{R}}$ and $\lambda_{\rm task}$ causes minor fluctuations in performance, demonstrating that our method is not overly sensitive to these hyperparameters and is easy to implement in practice.
\item For the $\lambda$ hyperparameter of feature mixup ($\mathbf{H}_{e_1}^{(l)}=\lambda \mathbf{H}_{i_1}^{(l)}+(1-\lambda) \mathbf{H}_{j_2}^{(l)}$), we introduce randomness by sampling from a Beta distribution $\operatorname{Beta}(\alpha, \beta)$. In our implementation, we set $(\alpha, \beta)=(0.5, 0.5)$. Our motivation and considerations are as follows:  
    \begin{itemize}
        \item \textbf{Why set $\alpha=\beta$?}  
        This ensures symmetric exploration of both source features. In contrast, using $\alpha\neq\beta$ would bias the interpolation towards one source feature, which may reduce the balance and diversity of the generated tasks.

        \item \textbf{Why set $\alpha=\beta<1$?}  
        When $\alpha, \beta<1$, this distribution is U-shaped and assigns higher probabilities to $\lambda$ values near 0 and 1. This means that $\lambda$ is more likely to be close to 0 or 1, so the interpolated features are more similar to one of the two original features. In this way, the generated features cover a broader range between the two original features, increasing the diversity of the tasks. Although $\alpha=\beta=1$ theoretically provides uniform coverage over $[0,1]$, it does not sample extreme $\lambda$ values frequently enough, leading to more similar interpolated features and limiting the variability of generated tasks. In contrast, when $\alpha, \beta>1$, the Beta distribution concentrates $\lambda$ near 0.5, so the interpolated features are mostly similar to the average of the two sources, which further reduces the variability of the generated tasks.
\item The results in \cref{tab:beta_ablation} confirm the rationality of our hyperparameter choices for $\lambda$.
    \end{itemize}
\end{itemize}

\noindent
Overall, although our method has several hyperparameters, each design choice has been well motivated, carefully explained, and thoughtfully evaluated. Moreover, our experiments demonstrate that the final performance remains stable across reasonable variations of these hyperparameters.

\begin{table}[!h]
  \centering
  \caption{Impact of the strengths $\lambda_{\mathcal{R}}$ of the loss term $\mathcal{R}$.} 
  \scalebox{0.99}{
  \begin{tabular}{llcc}
    \toprule
    \multirow{2}{*}{\textbf{Method}} & \multirow{2}{*}{\textbf{Hyperparameter}} & \multicolumn{2}{c}{\textbf{CIFAR-FS} \cite{bertinetto2018meta}} \\
    \cmidrule(r){3-4}
    & & 5-way 1-shot & 5-way 5-shot  \\
    \midrule
    \multirow{4}{*}{{\textsc{TDFML}-ProtoNet}}
    & $\lambda_{\mathcal{R}}=0.1$ & 40.20 ± 0.70 & 57.02 ± 0.72 \\
    & $\lambda_{\mathcal{R}}=0.01$ & 40.85 ± 0.69 & 57.01 ± 0.74 \\
    & $\lambda_{\mathcal{R}}=0.001$ & \textbf{41.91 ± 0.72} & \textbf{58.25 ± 0.72} \\
    & $\lambda_{\mathcal{R}}=0.0001$ & 40.10 ± 0.69 & 57.20 ± 0.72 \\
    \bottomrule
  \end{tabular}}
  \label{tab:ablation_1}
\end{table}

\begin{table}[!h]
  \centering
  \caption{Impact of the strengths $\lambda_{\rm task}$ of the loss term $\mathcal{L}_{\rm task}$.} 
  \scalebox{0.99}{
  \begin{tabular}{llcc}
    \toprule
    \multirow{2}{*}{\textbf{Method}} & \multirow{2}{*}{\textbf{Hyperparameter}} & \multicolumn{2}{c}{\textbf{CIFAR-FS} \cite{bertinetto2018meta}} \\
    \cmidrule(r){3-4}
    & & 5-way 1-shot & 5-way 5-shot  \\
    \midrule
    \multirow{4}{*}{{\textsc{TDFML}-ProtoNet}}
    & $\lambda_{\rm task}=1$ & 40.02 ± 0.75 & 56.40 ± 0.72 \\
    & $\lambda_{\rm task}=0.1$ & \textbf{41.91 ± 0.72} & \textbf{58.25 ± 0.72} \\
    & $\lambda_{\rm task}=0.01$ & 40.25 ± 0.69 & 57.30 ± 0.73 \\
    & $\lambda_{\rm task}=0.001$ & 40.57 ± 0.74 & 56.90 ± 0.70 \\
    \bottomrule
  \end{tabular}}
  \label{tab:ablation_2}
\end{table}

\begin{table}[!h]
  \centering
  \caption{Impact of different Beta distribution parameters $(\alpha, \beta)$ on the feature mixup coefficient $\lambda$.} 
  \scalebox{0.9}{
  \begin{tabular}{llcc}
    \toprule
    \multirow{2}{*}{\textbf{Beta Config}} & \multirow{2}{*}{\textbf{Distribution Shape of $\lambda$}} & \multicolumn{2}{c}{\textbf{CIFAR-FS} \cite{bertinetto2018meta}} \\
    \cmidrule(r){3-4}
    & & 5-way 1-shot & 5-way 5-shot  \\
    \midrule
    $\alpha=0.5, \beta=0.5$ & symmetric, U-shaped & \textbf{41.91 ± 0.72} & \textbf{58.25 ± 0.72} \\
    $\alpha=1, \beta=1$ & symmetric, uniform & 41.20 ± 0.74 & 57.80 ± 0.70 \\
    $\alpha=2, \beta=2$ & symmetric, hump-shaped & 41.06 ± 0.70 & 57.70 ± 0.71 \\
    $\alpha=2, \beta=0.5$ & biased to 1 & 40.90 ± 0.69 & 57.24 ± 0.72 \\
    $\alpha=0.5, \beta=2$ & biased to 0 & 40.96 ± 0.70 & 57.44 ± 0.71 \\
    \bottomrule
  \end{tabular}}
  \label{tab:beta_ablation}
\end{table}

\textbf{Robustness to model pool with varying difficulty level.} To demonstrate that our approach can handle a model pool containing models of varying difficulty, we construct a model pool that includes models trained on datasets with different difficulty levels. Specifically, the model pool includes models trained on VGG-Flower, CIFAR-FS, CUB, and MiniImageNet datasets, ordered from lower to higher difficulty. We compare different model selection criteria, including accuracy and our proposed AMS. The evaluation is conducted on the meta-testing subset of each dataset, respectively.

The results shown in \cref{tab:difficulty} indicate that simply selecting the model with the highest accuracy does not achieve satisfactory results compared with AMS. This is because high accuracy tends to be associated with easier models, which may have a large domain gap with the target distribution and reduce the diversity of selected models. For example, when meta-testing on MiniImageNet, selecting the highest-accuracy models tends to favor easier models trained on datasets such as VGG-Flower or CIFAR-FS, resulting in a large domain gap with the target distribution and leading to suboptimal meta-learning performance.  In contrast, AMS significantly outperforms accuracy-based selection because it is reward-driven and selects the ``correct'' models that lead to higher rewards, regardless of the model’s inherent difficulty, accuracy, or any other static standard.

\vfill

\end{document}